\documentclass{nature}

\usepackage{amssymb}
\usepackage{amsmath}
\usepackage{graphicx}
\usepackage{placeins}
\usepackage{booktabs}
\usepackage{tabularx}
\usepackage{array}
\usepackage{makecell}
\usepackage[margin=0.6in]{geometry}
\usepackage{setspace}
\usepackage{hyperref}
\hypersetup{hidelinks}
\usepackage{float}
\usepackage{caption}

\makeatletter
\let\saved@includegraphics\includegraphics
\AtBeginDocument{\let\includegraphics\saved@includegraphics}

\renewenvironment*{figure}{\@float{figure}}{\endfloat}
\makeatother

\title{From Compound Figures to Medical Multi-image Reasoning: Scaling Multimodal Large Language Models with Biomedical Literature}

\author{Zhen Chen$^{1,\ddag}$,
Yihang Fu$^{1,\ddag}$,
Rong Zhou$^{1}$,
Serina Applebaum$^{1}$,
Min Kyu Kim$^{2}$,
Aidan Gilson$^{3}$,
Morten Lee$^{1}$,
Salahudeen Mirza$^{4}$,
Gabriel Madera$^{1,5}$,
Mauro Giuffre$^{1}$,
Yuanting Pan$^{6}$,
Roy Jiang$^{1,7}$,
Hyunjae Kim$^{1}$,
Hua Xu$^{1}$,
Qingyu Chen$^{1,*}$
}

\begin{document}

\maketitle

\begin{spacing}{1.2}
\begin{affiliations}

\item Department of Biomedical Informatics and Data Science,
Yale School of Medicine, Yale University, USA

\item Scheie Eye Institute, Perelman School of Medicine,
University of Pennsylvania, USA

\item Department of Ophthalmology, Massachusetts Eye and Ear,
Harvard Medical School, USA

\item Yale Child Study Center,
Yale School of Medicine, Yale University, USA

\item School of Medicine, University of Puerto Rico, USA

\item Department of Biomedical Engineering,
Yale School of Engineering \& Applied Science,
Yale University, USA

\item Department of Dermatology,
Yale School of Medicine, Yale University, USA
\end{affiliations}

\vspace{-10mm}

\noindent
$^{\boldsymbol{\ddag}}$These authors contributed equally.\\
*Corresponding author: Qingyu Chen (qingyu.chen@yale.edu)

\end{spacing}

\section*{Abstract}

Multimodal large language models (MLLMs) are increasingly capable of supporting medical imaging tasks, yet most existing approaches focus on single-image settings. Clinical interpretation often requires integrating evidence across multiple related images, including different modalities,
views, or time points, but large-scale medical multi-image data and training strategies designed for such reasoning remain limited. We construct PMC-MI, a large-scale resource comprising 234,956 instruction instances derived from biomedical compound figures, with 10,555 multi-subimage instances further structured for reinforcement learning. Samples of the constructed instructions and policy-refinement data were independently assessed by medical reviewers. We also introduce PMC-MI-Bench, a manually reviewed benchmark separated from PMC-MI at the source-article level. We further propose a three-stage training framework and instantiate it as M$^3$LLM, combining
supervised fine-tuning on multi-image instructions, selection-aware reinforcement learning for question-conditioned visual-evidence selection, and supervised consolidation on a broader instruction mixture. We evaluate M$^3$LLM on PMC-MI-Bench, two public medical benchmarks, longitudinal chest radiographs, and a retrospective dermatology cohort. M$^3$LLM achieved the highest observed performance among representative general-purpose and medical MLLMs across the primary evaluations. On PMC-MI-Bench, it achieved scores of 81.6, 84.6, and 85.2 on single-subimage, multi-subimage, and relative-position tasks, respectively, compared with 75.6, 79.1, and 77.6 for the strongest baseline models and 71.8, 75.6, and 73.9 for the unadapted backbone. It also achieved the highest overall performance among the evaluated models on the provenance-screened OmniMedVQA evaluation set (88.6\%) and MMMU-Med (68.0\%), and the highest observed performance in two clinical evaluation settings after task-specific adaptation. These findings support the use of literature-derived multi-image supervision and targeted post-training for medical multi-image MLLMs. We make the related data and code available to support reproducible evaluation and further research.

\newpage

\section{Introduction}

Multimodal large language models (MLLMs) have shown increasing promise across a range of medical imaging applications, including disease diagnosis, progression assessment, documentation of imaging findings, and medical visual question answering (VQA)~\cite{li2023llavamed,wu2025towards,huatuogpt_vision,healthgpt}. By combining visual understanding with generative language modelling, MLLMs can support diverse medical imaging tasks and generate open-ended responses~\cite{nam2025multimodal,qin2026volmo,liu2025multimodal}. This allows the same model to be applied across different tasks through prompting or instruction tuning, rather than requiring task-specific fine-tuning for each individual task~\cite{cooper2025rethinking,su2026patch,li2024fine}. However, most medical MLLMs have been developed and evaluated primarily in single-image settings, with relatively few approaches designed to address the challenges of multi-image understanding. In clinical practice, healthcare professionals routinely synthesize information across multiple related images, for example, by integrating examinations from different imaging modalities for diagnosis~\cite{brin2025assessing,maleki2025multimodal,chen2021multimodal}
or comparing images acquired across visits to assess disease progression~\cite{carrasco2023prediction,jin2021predicting,sharma2025longitudinal}. Such settings require models not only to interpret individual images, but also to identify relationships among them and synthesize complementary or changing visual evidence.

\begin{figure}[H]
    \centering
    \includegraphics[width=0.85\textwidth]{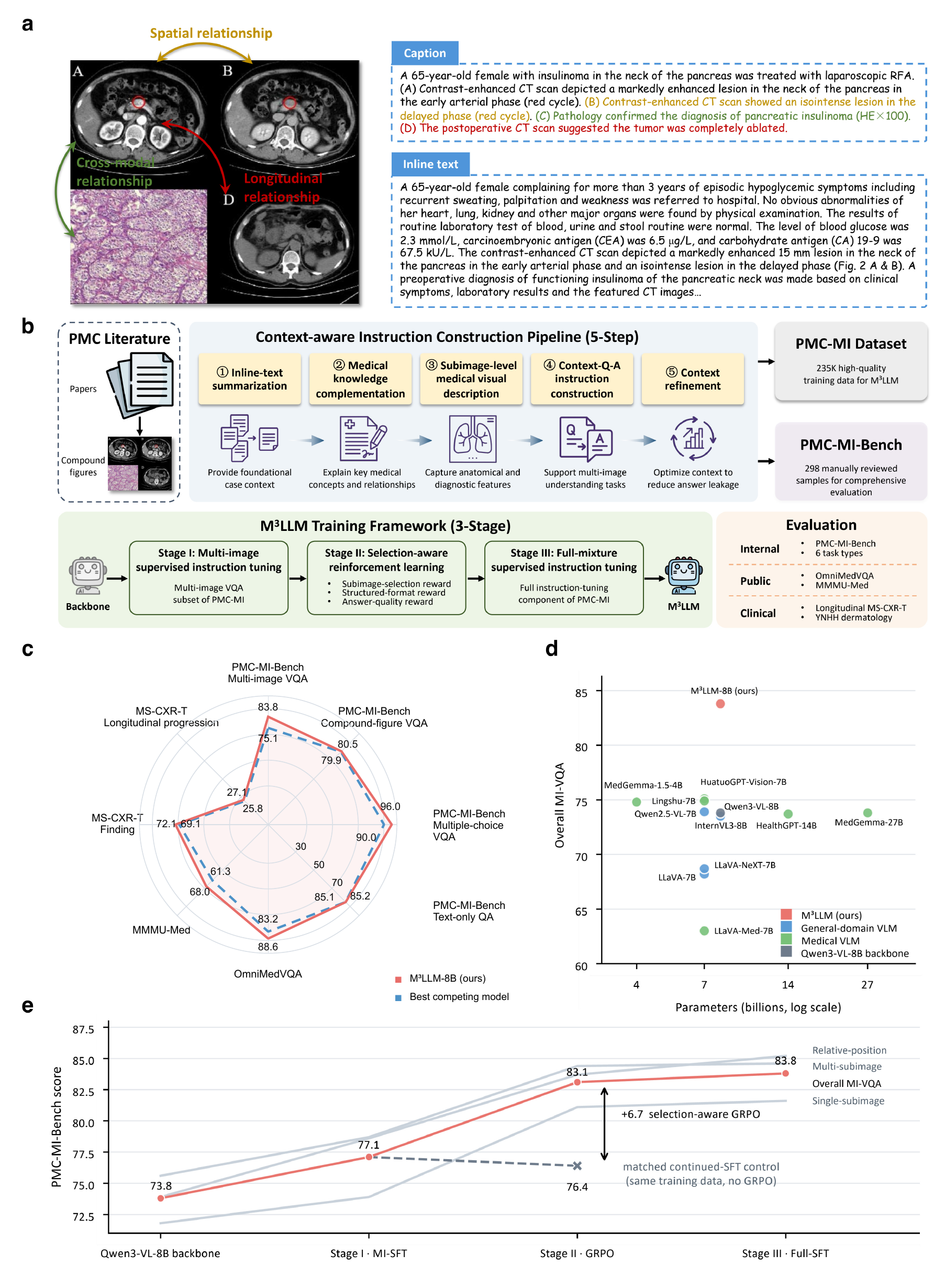}
    \caption{\textbf{Overview of the M$^3$LLM framework.} \textbf{a}, A PubMed Central (PMC)
    compound figure with its caption and inline text, annotated with the
    spatial, cross-modal, and longitudinal relationships across subimages.
    \textbf{b}, The five-step context-aware instruction-construction pipeline that builds PMC-MI and PMC-MI-Bench, the three-stage training framework of M$^3$LLM, and the evaluation settings. \textbf{c}, Comparison without target-task adaptation between M$^3$LLM and the strongest competing model on each of eight tasks. \textbf{d}, Overall multi-image visual question answering (VQA) score on PMC-MI-Bench versus model parameters. \textbf{e}, PMC-MI-Bench score after each training stage, alongside a matched continued supervised fine-tuning (SFT) control trained without reinforcement learning.}
    \label{fig:overview}
\end{figure}

\noindent A major bottleneck is the limited availability of large-scale medical multi-image datasets for model development and evaluation. Despite pioneering initiatives to improve access to medical imaging and multimodal data~\cite{ai2024ai,fedorov2021nci}, clinical multi-image datasets remain difficult to assemble and distribute. Such datasets may require examinations acquired across different devices, views, modalities, or time points to be linked together and paired with clinically meaningful annotations, while privacy, governance, and data-sharing constraints further limit large-scale aggregation and public release~\cite{wu2026real,langlotz2019roadmap,nawaz2026role,mckinney2020international}. Biomedical literature provides a complementary source of multi-image information that can be curated at scale and, where redistribution terms permit, shared to support reproducible research~\cite{huang2025systematic,kim2026medpmc}. In particular, compound figures commonly organize multiple related subimages within a shared biomedical context and accompany them with detailed captions and surrounding article text. As shown in Figure~\ref{fig:overview}a, these figures can capture relationships such as temporal changes, cross-modality correspondence, and anatomical comparisons, providing naturally linked visual and textual information for multi-image understanding~\cite{sala2017unravelling,rougier2014ten,takaya2020new,tae2025current}.

\noindent However, simply providing compound figures to a model is unlikely to be sufficient for effective multi-image understanding. Models need to learn relationships across images, identify the images relevant to a given question, and synthesize evidence across them. Prior work on medical multi-image modelling has primarily addressed specific clinical tasks using domain-specific medical imaging models. For example, models have been developed to jointly analyse paired longitudinal chest radiographs for temporal representation learning and progression assessment~\cite{bannur2023learning,chen2025cocacxr}. Many medical MLLMs, by contrast, have been developed through broad supervised instruction tuning, with relatively limited emphasis on challenges specific to multi-image understanding~\cite{ye2025multimodal,hu2025medical,li2023llavamed,huatuogpt_vision,
xu2025lingshu,healthgpt,sellergren2025medgemma}. More recent work, including MedGemma 1.5-4B, has explored multi-timepoint image interpretation with medical MLLMs, highlighting the feasibility and potential value of extending these models beyond single-image inputs~\cite{sellergren2026medgemma}.

\noindent In this work, we address these challenges in medical multi-image understanding. We first construct PMC-MI, a large-scale resource of 234,956 instruction instances derived from biomedical compound figures using a scalable, context-aware data-construction pipeline. The pipeline integrates article context, medical knowledge, subimage-level visual information, and relationships across images, while remaining adaptable to different language and multimodal models. PMC-MI spans multi-image VQA, compound-figure VQA, text-only QA, and multiple-choice VQA. Of these instances, 10,555 multi-subimage examples retain supporting-subimage identifiers and are allocated to policy refinement, providing supervision for reinforcement learning that explicitly optimizes visual-evidence selection together with answer quality. The remaining instances form the supervised instruction corpus. We further perform independent medical review on samples of the constructed instructions and policy-refinement data. In addition, we construct PMC-MI-Bench as a dedicated, manually reviewed benchmark separated from all PMC-MI training and validation data at the source-article level.

\noindent Second, we propose a three-stage training framework for medical multi-image MLLMs and instantiate it as M$^3$LLM. The first stage uses supervised fine-tuning on multi-image instructions to establish medical multi-image instruction-following capabilities and cross-image information integration. The second stage uses selection-aware reinforcement learning to jointly optimize question-relevant subimage selection, answer quality, and structured-output compliance. The final stage returns to supervised fine-tuning on a broader instruction mixture to retain the acquired multi-image capabilities while maintaining performance across other tasks.

\noindent Third, we conduct a systematic evaluation of both the proposed training framework and the resulting M$^3$LLM. The evaluation spans PMC-MI-Bench, two external public medical benchmarks, and two clinically relevant evaluations using real-world patient data: longitudinal chest radiographs for target-finding identification and progression assessment, and a retrospective cohort of dermatology patients at Yale New Haven Hospital (YNHH) for differential-diagnosis validity. Representative general-purpose and medical MLLMs are evaluated under consistent settings, with additional comparisons against task-specific medical imaging models. We further perform controlled stage-wise comparisons and reward ablations to characterize performance differences across training stages and objective variants. Finally, we make the related data and code available to support reproducible evaluation and further research.

\section{Results}

\subsection{PMC-MI as a training resource for medical multi-image learning}

\begin{figure}[H]
    \centering
    \includegraphics[width=0.95\textwidth]{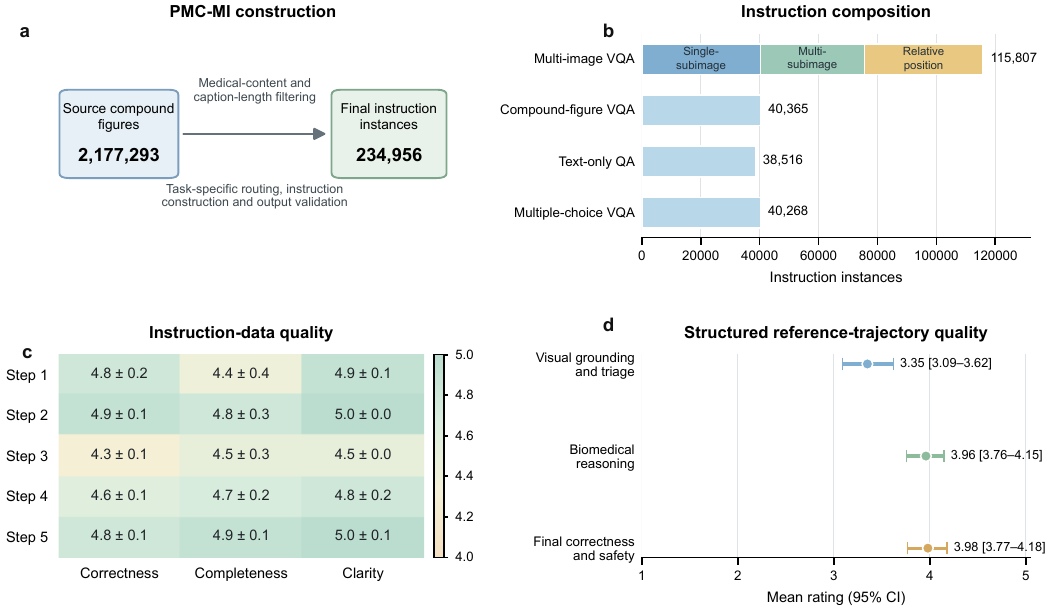}
    \caption{\textbf{Construction and quality assessment of the PMC-MI
    training resource.} \textbf{a}, Source and task-specific filtering of
    compound figures. \textbf{b}, Composition of the four instruction families; multi-image VQA comprises single-subimage, multi-subimage, and relative-position instances. \textbf{c}, Manual assessment of PMC-MI data quality across the five instruction-construction steps (mean $\pm$ standard deviation (SD),
    1--5 scale). \textbf{d}, Manual assessment of structured reference trajectories used for policy refinement (mean rating with 95\% instance-level bootstrap confidence interval (CI), 1--5 scale).}
    \label{fig:pmc_mi}
\end{figure}

\noindent PMC-MI was used as the training resource to develop M$^3$LLM. Its construction and quality-assessment procedures are described in Methods (Section~\ref{sec:pmc_mi_construction}). PMC-MI comprised 234,956 compound-figure-derived instances spanning four instruction families: multi-image visual question answering (VQA), compound-figure VQA, text-only question answering (QA), and multiple-choice VQA. Multi-image VQA constituted the largest component, with 115,807 instances (49.3\%), comprising 40,261 single-subimage VQA, 35,275 multi-subimage VQA, and 40,271 relative-position VQA instances. Compound-figure VQA, text-only QA, and multiple-choice VQA comprised 40,365 (17.2\%), 38,516 (16.4\%), and 40,268 (17.1\%) instances, respectively. The data-processing and PMC-MI composition results are shown in Figure~\ref{fig:pmc_mi}a,b.

\noindent PMC-MI supported both supervised fine-tuning and selection-aware reinforcement learning. Of the 234,956 instances, 10,555 multi-subimage VQA instances were allocated to the dedicated policy-refinement partition, comprising 10,355 instances for Stage~II policy optimization and 200 for validation; the remaining 224,401 instances formed
the supervised instruction corpus. The policy-refinement instances retained the instruction-derived reference subimage identifiers associated with each question and reference answer, providing reference signals for visual-evidence selection during Stage~II. Construction details are provided in Methods (Section~\ref{sec:policy_refinement_data}).

\noindent Next, we assessed the quality of the constructed PMC-MI data through independent medical review. Three medical reviewers participated in the evaluation of 140 sampled outputs spanning the five construction steps, with each output independently rated by two reviewers. Correctness, completeness, and clarity were assessed separately using a prespecified rubric. Across the five steps, mean correctness scores ranged from 4.3 to 4.9, completeness scores from 4.4 to 4.9, and clarity scores from 4.5 to 5.0. Step~3 (subimage-level medical visual description) had the lowest observed correctness and clarity scores, at 4.3 and 4.5, respectively, whereas Step~1 (inline-text summarization) had the lowest completeness score, at 4.4. Step~5 (context refinement to reduce answer leakage) received mean scores of 4.8 for correctness, 4.9 for completeness, and 5.0 for clarity. The manual assessment results are shown in Figure~\ref{fig:pmc_mi}c and Supplementary Table~1.

\noindent Moreover, we assessed the reliability of the medical review ratings among the 78 construction outputs that received three independent ratings. The dimension-specific intraclass correlation coefficients (ICCs) were 0.867 for correctness, 0.751 for completeness, and 0.720 for clarity. Reviewers assigned identical scores in 74.8\% of assessments, and 98.3\% of ratings differed by no more than one rating category. 

\noindent We additionally assessed the quality of the structured reference trajectories used for policy refinement. Five medical reviewers collectively evaluated a simple random sample of 100 trajectories with complete local image assets using the reviewer-assignment scheme described in Methods. The mean ratings were 3.35 (95\% CI, 3.09--3.62) for visual grounding and image triage, 3.96 (95\% CI, 3.76--4.15) for biomedical reasoning, and 3.98 (95\% CI, 3.77--4.18) for final clinical correctness and safety. Ordinal Krippendorff's $\alpha$ was highest for visual grounding and image triage (0.55; 95\% CI, 0.31--0.70), followed by final clinical correctness and safety (0.40; 95\% CI, 0.15--0.59) and biomedical reasoning (0.33; 95\% CI, 0.09--0.53). A secondary low-score-triggered review confirmed material visual-grounding errors in 9 of 100 trajectories, biomedical-reasoning errors in 6 of 100, and final-answer correctness errors in 8 of 100. These non-mutually-exclusive values are screen-confirmed counts within the sampled trajectories. Detailed results and representative low-scoring cases are reported in Supplementary Table~2 and Supplementary Figure~22, respectively.

\subsection{PMC-MI-Bench as a manually reviewed evaluation benchmark}

\noindent In addition to PMC-MI as the training resource, we constructed PMC-MI-Bench as a separate evaluation benchmark. The benchmark curation and review procedures are described in Methods (Section~\ref{sec:pmc_mi_bench_construction}). Of the original 300 examples, 298 were retained; one compound-figure VQA instance and one text-only QA instance were excluded. We then compared source PMCIDs across the retained benchmark and all model-development partitions and removed 117 PMC-MI records whose source articles occurred in the benchmark. Each retained benchmark example had a distinct sample identifier linking its source PMCID and compound-figure image, and no source-article overlap remained. The final PMC-MI-Bench comprised 50 single-subimage VQA, 50 multi-subimage VQA, 50 relative-position VQA, 49 compound-figure VQA, 49 text-only QA, and 50 multiple-choice VQA examples. Benchmark records contained a mean of 4.81 constituent subimages, while their compound-figure captions and linked inline text contained means of 102.2 and 191.1 words, respectively.

\noindent In addition to source-article separation, we assessed potential verbatim memorization using both prefix-conditioned continuation and question-only response analyses (Methods, Section~\ref{sec:memorization_analysis}). In the prefix-conditioned analysis, M$^3$LLM produced no continuation satisfying either \texttt{exact@30} or \texttt{exact@50} when provided with 50, 100, or 150 reference-answer tokens as context (all observed rates, 0\%).

\noindent Verbatim matches were also uncommon in the question-only analysis (Supplementary Table~4). Point estimates were equal-weight means of the four eligible task-specific match rates, with 95\% source-article-clustered bootstrap CIs. M$^3$LLM had an \texttt{exact@30} estimate of 1.01\% (0.00--2.58\%), compared with 0.50\% (0.00--1.67\%) for its Qwen3-VL-8B backbone. The paired difference was 0.51 percentage points (0.00--1.67). At the stricter \texttt{exact@50} threshold, the corresponding estimates were 0.54\% (0.00--1.79\%) and 0.00\% (0.00--0.00\%), with a paired difference of 0.54 percentage points (0.00--1.79).

\subsection{Performance on PMC-MI-Bench}

\begin{figure}[H]
    \centering
    \includegraphics[width=0.9\textwidth]{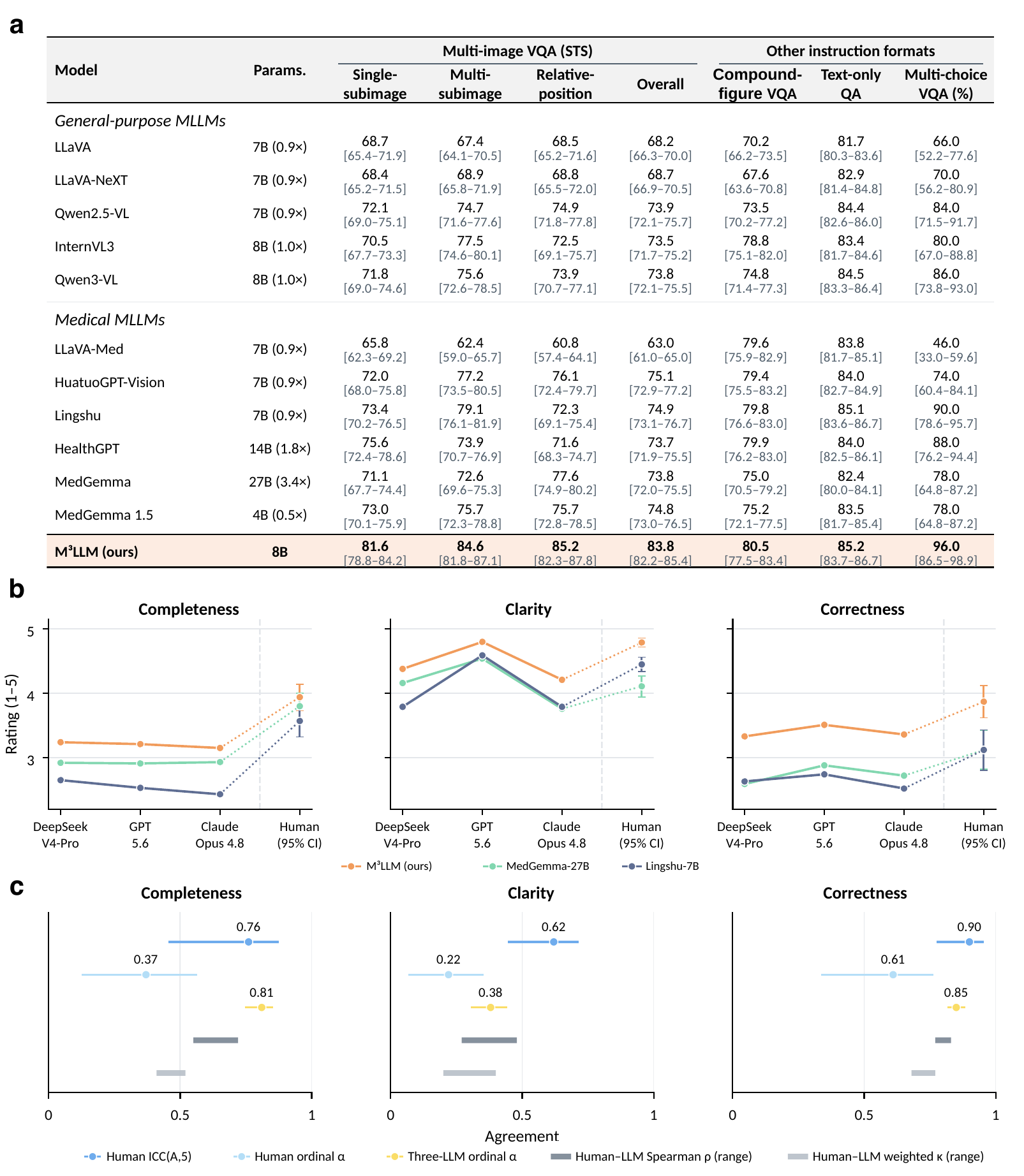}
    \caption{\textbf{Performance on PMC-MI-Bench.} \textbf{a}, Comparison of
    M$^3$LLM with general-purpose and medical multimodal large language models (MLLMs) across the six task categories. Values use Semantic Textual Similarity (STS) for open-ended tasks and accuracy for multiple-choice visual question answering (VQA), with 95\% task-stratified bootstrap confidence intervals (CIs) and Wilson CIs, respectively. Parameter ratios in parentheses are relative to 8 billion parameters (1.0$\times$). \textbf{b}, Mean five-point ratings of open-ended PMC-MI-Bench responses from three blinded large language model (LLM) judges and blinded human reviewers for completeness, clarity, and reference-aligned correctness. The LLM judges evaluated 744 response items, whereas the human reviewers evaluated 210 response items from 70 questions; error bars for the human means denote 95\% bootstrap CIs. \textbf{c}, Reliability and agreement within and between human and LLM evaluators by rating dimension. Coloured points and intervals show the absolute-agreement reliability of the five-reviewer mean (ICC(A,5)), human ordinal Krippendorff's $\alpha$, and three-LLM ordinal Krippendorff's $\alpha$; grey segments show the ranges across LLM judges for human--LLM Spearman $\rho$ and quadratic-weighted Cohen's $\kappa$.}
    \label{fig:pmc_mi_bench}
\end{figure}

\paragraph{Overall performance.}
\noindent We evaluated M$^3$LLM on the 298-question PMC-MI-Bench (Figure~\ref{fig:pmc_mi_bench}a) against five general-purpose MLLMs: LLaVA-7B~\cite{llava}, LLaVA-NeXT-7B~\cite{llava_next}, Qwen2.5-VL-7B~\cite{team2025qwen2_5}, InternVL3-8B~\cite{chen2024internvl}, and Qwen3-VL-8B~\cite{bai2025qwen3}; and six medical MLLMs: LLaVA-Med-7B~\cite{li2023llavamed}, HuatuoGPT-Vision-7B~\cite{huatuogpt_vision}, Lingshu-7B~\cite{xu2025lingshu}, HealthGPT-14B~\cite{healthgpt}, MedGemma-27B~\cite{sellergren2025medgemma}, and MedGemma~1.5-4B~\cite{sellergren2026medgemma}. M$^3$LLM achieved the highest point estimates across all three multi-image tasks, with Semantic Textual Similarity (STS) scores of 81.6 (95\% task-stratified bootstrap CI, 78.8--84.2) on single-subimage VQA, 84.6 (81.8--87.1) on multi-subimage VQA, and 85.2 (82.3--87.8) on relative-position VQA. These point estimates were 9.8, 9.0, and 11.3 points above those of the Qwen3-VL-8B backbone, respectively, and 6.0, 5.5, and 7.6 points above the highest-scoring alternative model in the corresponding task. Across the three multi-image tasks, M$^3$LLM achieved an overall multi-image VQA (MI-VQA) score of 83.8 (95\% task-stratified bootstrap CI, 82.2--85.4), compared with 73.8 for the backbone and 75.1 for the highest-scoring alternative, HuatuoGPT-Vision-7B.

\noindent For the remaining instruction formats, M$^3$LLM yielded the highest point estimate on compound-figure VQA (STS, 80.5; 95\% task-stratified bootstrap CI, 77.5--83.4) and multiple-choice VQA (48/50 correct, 96.0\%; 95\% Wilson CI, 86.5--98.9). On text-only QA, its STS score was 85.2 (95\% task-stratified bootstrap CI, 83.7--86.7), 0.1 points above Lingshu-7B. Relative to the Qwen3-VL-8B backbone, the corresponding point-estimate gains were 5.7 points on compound-figure VQA, 0.7 points on text-only QA, and 10.0 percentage points on multiple-choice VQA. Thus, the higher multi-image scores did not coincide with lower point estimates than the backbone on the other PMC-MI-Bench formats. We additionally report BLEU-4, ROUGE-L, and BERTScore F1 as complementary
reference-based measures in Supplementary Table~5.

\noindent Representative comparisons across relative-position, compound-figure, multi-subimage, text-only, and single-subimage tasks are provided in Supplementary Figs.~23--27.

\paragraph{Human and model-based evaluation of open-ended responses.}
To complement reference-based automatic metrics, we assessed open-ended PMC-MI-Bench responses from M$^3$LLM, MedGemma-27B, and Lingshu-7B using three blinded LLM judges and an independent panel of five medical reviewers. Evaluators received the question, manually verified reference answer, and an anonymized model response. Using a five-point rubric, they rated completeness, clarity, and correctness relative to the reference answer. The correctness criterion captured factual contradictions, unsupported material claims, and clinically consequential errors where applicable.
M$^3$LLM received the highest mean rating in all nine judge-criterion combinations. Averaged across the three judges, M$^3$LLM obtained scores of 3.20 for completeness, 4.46 for clarity, and 3.40 for correctness. The corresponding scores were 2.92, 4.15, and 2.73 for MedGemma-27B, and 2.54, 4.06, and 2.63 for Lingshu-7B. Relative to the strongest baseline for each criterion, the margins were 0.28, 0.31, and 0.67 points, respectively, with the largest separation observed for correctness.

\noindent Independent blinded human evaluation of 210 response items from 70 questions yielded the same top-ranked model across all three criteria (Figure~\ref{fig:pmc_mi_bench}b). M$^3$LLM received mean human ratings of 3.94 (95\% CI, 3.73--4.14) for completeness, 4.79 (95\% CI, 4.72--4.86) for clarity, and 3.87 (95\% CI, 3.62--4.12) for correctness. Thirty response items were evaluated by all five reviewers. On this fully crossed subset, ordinal inter-rater agreement was highest for correctness (Krippendorff's $\alpha=0.61$, 0.34--0.76), compared with completeness ($\alpha=0.37$, 0.13--0.56) and clarity ($\alpha=0.22$, 0.07--0.35). Reliability of the aggregated five-rater mean was higher, with ICC(A,5) values of 0.90, 0.76, and 0.62 for correctness, completeness, and clarity, respectively. The full response-quality results and detailed inter-rater reliability estimates, including 95\% confidence intervals, are reported in Supplementary Tables~6 and~7, respectively.

\noindent Agreement between the human and LLM-based evaluations varied by rating dimension (Supplementary Table~8). Across the three LLM judges, human-LLM Spearman correlations ranged from 0.77 to 0.83 for correctness, from 0.55 to 0.72 for completeness, and from 0.27 to 0.48 for clarity. The corresponding quadratic-weighted $\kappa$ values ranged from 0.68 to 0.77, from 0.41 to 0.52, and from 0.20 to 0.40, respectively. Agreement among the three LLM judges showed the same ordering, with three-way ordinal $\alpha$ values of 0.85 (0.82--0.88) for correctness, 0.81 (0.75--0.85) for completeness, and 0.38 (0.31--0.44) for clarity (Supplementary Table~9). Thus, evaluator
convergence was strongest for correctness and completeness, whereas clarity ratings showed greater variability across both human and model-based evaluators.

\paragraph{Stage-level analysis.}
We next examined how each stage of the M$^3$LLM training procedure affected performance on PMC-MI-Bench. The procedure comprised Stage~I, multi-image supervised fine-tuning (MI-SFT); Stage~II, selection-aware reinforcement learning using Group Relative Policy Optimization (GRPO); and Stage~III, full-mixture supervised fine-tuning (Full-SFT), as described in Methods (Section~\ref{sec:three_stage_training}). Table~\ref{tab:controlled_training_analysis}a summarizes performance after each stage. Stage~I increased the point estimates across all three multi-image tasks relative to the Qwen3-VL-8B backbone, which served as the initialization for M$^3$LLM. The score increased from 71.8 to 73.9 for single-subimage VQA, from 75.6 to 78.7 for multi-subimage VQA, and from 73.9 to 78.6 for relative-position VQA.

\noindent Stage~II produced further gains across all three multi-image tasks. The score increased from 73.9 after MI-SFT to 81.1 after GRPO for single-subimage VQA, from 78.7 to 84.4 for multi-subimage VQA, and from 78.6 to 83.7 for relative-position VQA. To compare GRPO with additional supervised optimization on the same prompt set, we included a continued structured-SFT condition initialized from the same Stage~I checkpoint. The control model achieved a score of 74.8 for single-subimage VQA, 79.8 for multi-subimage VQA, and 74.5 for relative-position VQA. The GRPO condition exceeded the continued structured-SFT condition by 6.3 points for single-subimage VQA, 4.6 points for multi-subimage VQA, and 9.2 points for relative-position VQA.

\noindent Stage~III retained the gains on the multi-image tasks, with a final score of 81.6 for single-subimage VQA, 84.6 for multi-subimage VQA, and 85.2 for relative-position VQA. It also increased the point estimates from 77.2 to 80.5 for compound-figure VQA and from 82.9 to 85.2 for text-only QA, while multiple-choice VQA accuracy increased from 92.0\% to 96.0\%. Thus, the final full-mixture supervised stage retained the Stage~II multi-image gains while increasing the point estimates on the other instruction formats.

\begin{table}[t]
    \centering
    \caption{
    \textbf{Controlled analyses of the three-stage PMC-MI training framework.} \textbf{a}, Checkpoint-level comparisons summarize performance across multi-image initialization, policy refinement, and post-reinforcement-learning instruction consolidation. \textbf{b}, Reward-component ablations compare the full selection-aware Group Relative Policy Optimization (GRPO) objective with variants in which one reward component was disabled.
    }
    \label{tab:controlled_training_analysis}

    \scriptsize
    \setlength{\tabcolsep}{3.3pt}
    \renewcommand{\arraystretch}{1.16}

    \textbf{a,} Checkpoint-level analysis of staged training and consolidation

    \vspace{2pt}

    \begin{tabularx}{\textwidth}{
        @{}
        >{\raggedright\arraybackslash}X
        c c c c c c c
        @{}
    }
        \toprule

        \textbf{Training condition}
        &
        \makecell{\textbf{Single-}\\\textbf{subimage}}
        &
        \makecell{\textbf{Multi-}\\\textbf{subimage}}
        &
        \makecell{\textbf{Relative-}\\\textbf{position VQA}}
        &
        \makecell{\textbf{Overall}\\\textbf{MI-VQA}}
        &
        \makecell{\textbf{Compound-}\\\textbf{figure VQA}}
        &
        \makecell{\textbf{Text-only}\\\textbf{QA}}
        &
        \makecell{\textbf{Multiple-choice}\\\textbf{VQA}} \\

        \midrule

        Qwen3-VL-8B
        & 71.8
        & 75.6
        & 73.9
        & 73.8
        & 74.8
        & 84.5
        & 86.0 \\

        MI-SFT
        & 73.9
        & 78.7
        & 78.6
        & 77.1
        & 75.8
        & 84.1
        & 88.0 \\

        MI-SFT $\rightarrow$ continued structured SFT
        & 74.8
        & 79.8
        & 74.5
        & 76.4
        & 76.5
        & 83.6
        & 88.0 \\

        MI-SFT $\rightarrow$ GRPO
        & 81.1
        & 84.4
        & 83.7
        & 83.1
        & 77.2
        & 82.9
        & 92.0 \\

        MI-SFT $\rightarrow$ GRPO $\rightarrow$ MI-only SFT
        & \textbf{82.2}
        & \textbf{85.0}
        & 84.3
        & \textbf{83.8}
        & 77.9
        & 82.6
        & 90.0 \\

        \textbf{MI-SFT $\rightarrow$ GRPO $\rightarrow$ Full-SFT}
        & 81.6
        & 84.6
        & \textbf{85.2}
        & \textbf{83.8}
        & \textbf{80.5}
        & \textbf{85.2}
        & \textbf{96.0} \\

        \bottomrule
    \end{tabularx}
    \vspace{5mm}

    \textbf{b,} Reward-component ablation after identical consolidation

    \vspace{2pt}

    \begin{tabularx}{\textwidth}{
        @{}
        >{\raggedright\arraybackslash}X
        c c c
        c c c c
        @{}
    }
        \toprule
        \textbf{GRPO reward formulation}
        &
        $R_{\mathrm{sel}}$
        &
        $R_{\mathrm{fmt}}$
        &
        $R_{\mathrm{ans}}$
        &
        \makecell{\textbf{Single-}\\\textbf{subimage}}
        &
        \makecell{\textbf{Multi-}\\\textbf{subimage}}
        &
        \makecell{\textbf{Relative-}\\\textbf{position}}
        &
        \makecell{\textbf{Overall}\\\textbf{MI-VQA}} \\
        \midrule

        Selection reward omitted
        &  & \checkmark & \checkmark
        & 78.4
        & 80.3
        & 81.2
        & 80.0 \\

        Format reward omitted
        & \checkmark &  & \checkmark
        & 80.7
        & 83.5
        & 84.0
        & 82.7 \\

        Answer reward omitted
        & \checkmark & \checkmark & 
        & 67.3
        & 67.8
        & 59.5
        & 64.9 \\

        \textbf{Full reward}
        & \checkmark & \checkmark & \checkmark
        & \textbf{81.6}
        & \textbf{84.6}
        & \textbf{85.2}
        & \textbf{83.8} \\

        \bottomrule
    \end{tabularx}

    \vspace{3pt}

    \begin{minipage}{0.99\textwidth}
        \footnotesize
        \textit{Notes:}
        All Stage~II conditions were initialized from the same MI-SFT checkpoint. Continued structured SFT and GRPO used the same prompt set and structured response schema and were matched for the number of prompt presentations and optimizer updates. GRPO additionally generated 16 on-policy responses per prompt; trajectory generation and total forward-pass computation were therefore not matched between these conditions. In panel~b, all GRPO variants used the same rollout and training configuration and underwent an identical subsequent full-mixture consolidation stage, with the specified reward component disabled in each ablation. MI-SFT denotes multi-image supervised fine-tuning; SFT, supervised fine-tuning.
    \end{minipage}
\end{table}

\paragraph{Reward-component analysis.}
We further compared the full Stage~II reward formulation with variants in which the subimage-selection, structured-format, or answer-quality reward was individually disabled (Table~\ref{tab:controlled_training_analysis}b). All variants underwent the same subsequent Full-SFT stage, providing a controlled comparison of the resulting reward formulations.

\noindent Removing the subimage-selection reward reduced the score from 81.6 to 78.4 for single-subimage VQA, from 84.6 to 80.3 for multi-subimage VQA, and from 85.2 to 81.2 for relative-position VQA. Removing the structured-format reward reduced the score from 81.6 to 80.7 for single-subimage VQA, from 84.6 to 83.5 for multi-subimage VQA, and from 85.2 to 84.0 for relative-position VQA. The largest decreases were observed after removal of the answer-quality reward, with the score decreasing from 81.6 to 67.3 for single-subimage VQA, from 84.6 to 67.8 for multi-subimage VQA, and from 85.2 to 59.5 for relative-position VQA.

\noindent We further examined the sensitivity of Stage~II policy refinement to the relative weighting of the three reward components (Table~\ref{tab:reward_operator_sensitivity}). Among five prespecified operators, the selected coefficients $(\lambda_{\mathrm{ans}},\lambda_{\mathrm{sel}},\lambda_{\mathrm{fmt}})=(0.6,0.3,0.1)$ yielded the highest mean answer reward on the held-out Stage~II validation set (86.8), together with a subimage-selection F1 of 86.9 and a valid-format rate of 98.1\%. This operator was selected using the validation answer reward alone, without reference to PMC-MI-Bench. After an identical subsequent Full-SFT stage, overall MI-VQA varied from
82.9 to 83.8 across the five operators, while BERTScore F1 varied from 84.8 to 85.5. The selected operator also achieved the highest point estimates for both
overall MI-VQA and BERTScore F1 after Full-SFT. It was used for the final
training configuration based on the prespecified validation answer-reward
criterion.

\begin{table*}[h]
\centering
\caption{
\textbf{Sensitivity to the Stage~II reward aggregation operator.} Five candidate weighted linear operators were specified before the
sensitivity analysis. The operator marked with $\dagger$ was selected using the mean answer-quality reward on the held-out Stage~II validation set ($n=200$); PMC-MI-Bench results were not used for operator selection. Validation answer reward and subimage-selection F1 are reported on a 0--100 scale. Valid-format rate is the percentage of responses satisfying the complete prescribed output structure. PMC-MI-Bench results were obtained after an identical subsequent Full-SFT stage. Overall MI-VQA is the unweighted mean of the task-level STS scores for single-subimage, multi-subimage, and relative-position VQA. BERTScore F1 is reported on a 0--100 scale as an additional answer-level diagnostic. Full-SFT denotes full-mixture supervised fine-tuning; MI-VQA, multi-image visual question answering; STS, Semantic Textual Similarity. Bold values denote the highest point estimate in each column.
}
\label{tab:reward_operator_sensitivity}

\small
\setlength{\tabcolsep}{4.5pt}
\renewcommand{\arraystretch}{1.14}

\begin{tabular*}{\textwidth}{
@{\extracolsep{\fill}}
c c c
c c c
c c
@{}
}
\toprule

\multicolumn{3}{c}{\textbf{Operator coefficients}}
&
\multicolumn{3}{c}{\textbf{Stage~II validation ($n=200$)}}
&
\multicolumn{2}{c}{\textbf{PMC-MI-Bench after Full-SFT}}
\\

\cmidrule(lr){1-3}
\cmidrule(lr){4-6}
\cmidrule(l){7-8}

$\lambda_{\mathrm{ans}}$ &
$\lambda_{\mathrm{sel}}$ &
$\lambda_{\mathrm{fmt}}$ &
\makecell{\textbf{Answer}\\\textbf{reward}} &
\makecell{\textbf{Subimage-selection}\\\textbf{F1}} &
\makecell{\textbf{Valid-format}\\\textbf{rate (\%)}} &
\makecell{\textbf{Overall}\\\textbf{MI-VQA}} &
\makecell{\textbf{BERTScore}\\\textbf{F1}} \\

\midrule

0.8 & 0.1 & 0.1
& 86.4 & 74.6 & 97.5 & 83.6 & 84.8 \\

0.7 & 0.2 & 0.1
& 85.7 & 80.8 & 97.8 & 83.2 & 85.1 \\

$0.6^{\dagger}$ & 0.3 & 0.1
& \textbf{86.8} & 86.9 & \textbf{98.1}
& \textbf{83.8} & \textbf{85.5} \\

0.5 & 0.4 & 0.1
& 85.2 & \textbf{87.2} & 98.0 & 83.1 & 85.0 \\

0.6 & 0.2 & 0.2
& 85.1 & 81.4 & 98.0 & 82.9 & 84.9 \\

\bottomrule
\end{tabular*}

\vspace{3pt}

\begin{minipage}{0.985\textwidth}
\footnotesize
\textit{Notes:}
Validation answer reward is the mean $R_{\mathrm{ans}}$, implemented using BERTScore F1. Subimage-selection F1 is the mean set-level F1 between generated and instruction-derived reference subimage sets. A response was counted as format-valid when $R_{\mathrm{fmt}}=1$.
\end{minipage}
\end{table*}

\subsection{Performance on public medical benchmarks}

\begin{figure}[H]
    \centering
    \includegraphics[width=0.9\textwidth]{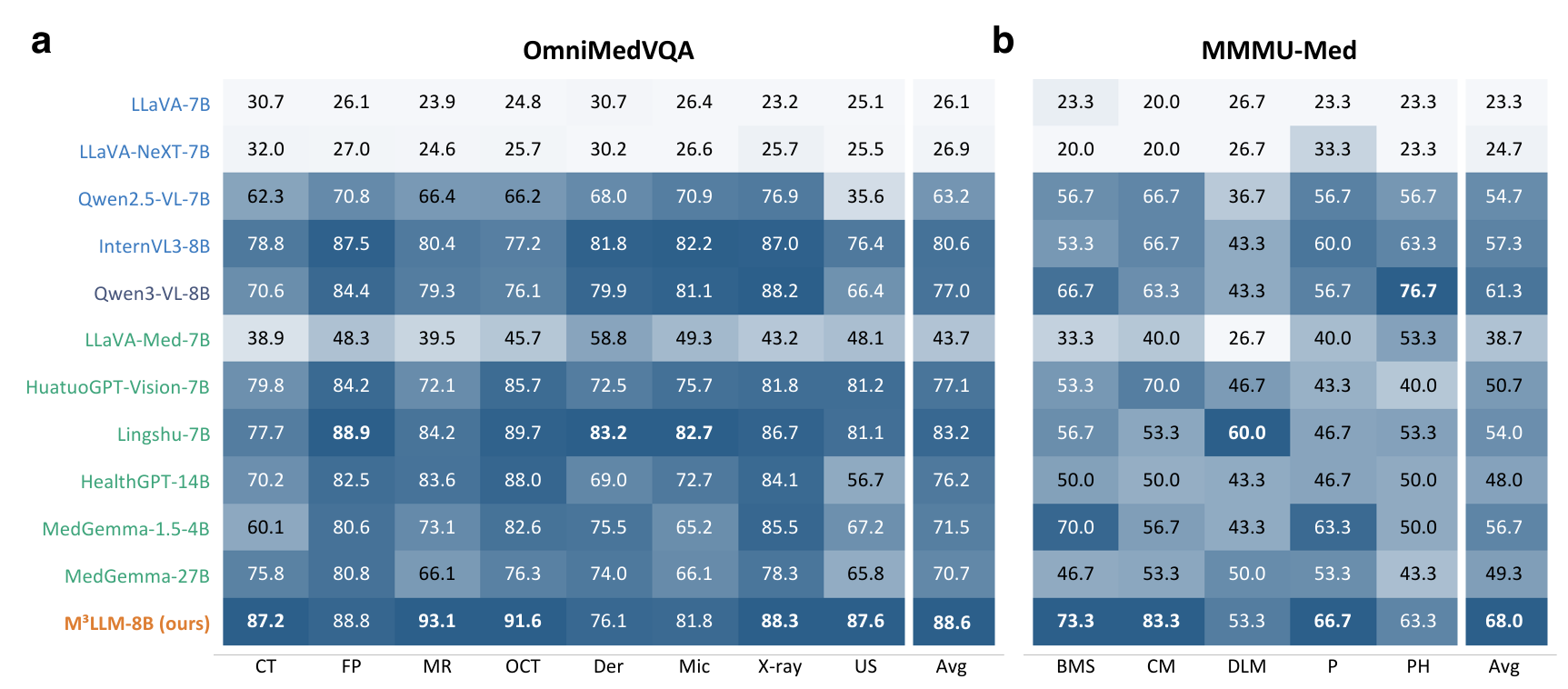}
    \caption{\textbf{Performance on public medical benchmarks.} \textbf{a},
    Zero-shot accuracy on the provenance-screened, publicly accessible OmniMedVQA evaluation set by imaging modality ($n=86{,}938$ question--answer pairs after excluding Covid CT, CoronaHack, Covid19 heywhale, and COVIDx CXR-4). \textbf{b}, Zero-shot accuracy on MMMU-Med by medical discipline ($n=150$ questions; 30 per discipline). Values are percentages. CT, computed tomography; FP, fundus photography; MR, magnetic resonance imaging; OCT, optical coherence tomography; Der, dermoscopy; Mic, microscopy; US, ultrasound; Avg, overall accuracy across all retained question--answer pairs; BMS, basic medical science; CM, clinical medicine; DLM, diagnostics and laboratory medicine; P, pharmacy; PH, public health. Bold indicates the highest value in each column.}
    \label{fig:public_benchmarks}
\end{figure}

\noindent We next evaluated M$^3$LLM on two established public medical benchmarks: the publicly accessible subset of OmniMedVQA~\cite{hu2024omnimedvqa} and the Health and Medicine subset of MMMU (MMMU-Med)~\cite{yue2024mmmu}. Before evaluation, we reviewed source-dataset provenance for the publicly accessible OmniMedVQA release. Four source subsets, Covid CT, CoronaHack, Covid19 heywhale, and COVIDx CXR-4, were excluded because their documentation indicated potential inclusion of images collected from scientific publications or preprints. This source-level exclusion removed 2,058 question--answer items, yielding an evaluation set of 86,938 items from 38 source datasets across eight imaging modalities. MMMU-Med comprised 150 questions from five medical disciplines. Published documentation for the retained OmniMedVQA subsets and MMMU-Med did not identify PMC or PMC-derived resources as data sources.

\noindent All models were evaluated using their released checkpoints under the same zero-shot protocol, without benchmark-specific fine-tuning, parameter updates, or in-context demonstrations. Neither OmniMedVQA nor MMMU-Med was used during PMC-MI post-training or checkpoint selection for M$^3$LLM. Here, zero-shot evaluation denotes the absence of benchmark-specific adaptation in our experiments.

\paragraph{OmniMedVQA results.}
On the revised OmniMedVQA evaluation set, M$^3$LLM achieved the highest overall accuracy of 88.6\% (Figure~\ref{fig:public_benchmarks}a), compared with 83.2\% for Lingshu-7B, the highest-scoring alternative model, and 77.0\% for its Qwen3-VL-8B backbone. Its modality-wise accuracies were 87.2\% for computed tomography (CT), 88.8\% for fundus photography, 93.1\% for magnetic resonance imaging (MRI), 91.6\% for optical coherence tomography (OCT), 76.1\% for dermoscopy, 81.8\% for microscopy, 88.3\% for X-ray, and 87.6\% for ultrasound. M$^3$LLM ranked first in CT, MRI, OCT, X-ray, and ultrasound and was 0.1, 7.1, and 0.9 percentage points below the strongest comparator in fundus photography, dermoscopy, and microscopy, respectively.

\noindent M$^3$LLM exceeded its Qwen3-VL-8B backbone by 11.6 percentage points in overall OmniMedVQA accuracy. These results compare released checkpoints on the same provenance-screened evaluation set using a common zero-shot inference and scoring protocol.

\paragraph{MMMU-Med results.}
On MMMU-Med, M$^3$LLM achieved the highest overall accuracy of 68.0\% (Figure~\ref{fig:public_benchmarks}b). The general-purpose MLLM baselines achieved accuracies ranging from 23.3\% to 61.3\%, whereas the medical MLLM baselines ranged from 38.7\% to 56.7\%. M$^3$LLM exceeded its Qwen3-VL-8B backbone by 6.7 percentage points and ranked first in three of the five medical disciplines.

\subsection{Validation on longitudinal chest radiographs}

\noindent Beyond standard benchmarks, we further evaluated M$^3$LLM in two clinical validation settings using patient-level data. The first evaluation used an MS-CXR-T-derived cohort of longitudinal chest radiographs and included two tasks, both of which supplied the same prior and current radiograph pair as input. For target-finding identification, the model received the radiograph pair together with a queried finding and predicted, for each of five radiographic findings in turn, whether that finding was the annotated target. For progression classification, the model received the same radiograph pair together with the name of the annotated target finding and predicted whether that finding was improving, stable, or worsening. We evaluated models both without target-task adaptation and after task-specific fine-tuning. The comparison included the same general-purpose and medical MLLM baselines used above, as well as specialized chest-radiograph models. For task-specific adaptation, all models were evaluated using the same patient-level training and test split to enable direct comparison.

\begin{figure}[H]
    \centering
    \includegraphics[width=0.9\textwidth]{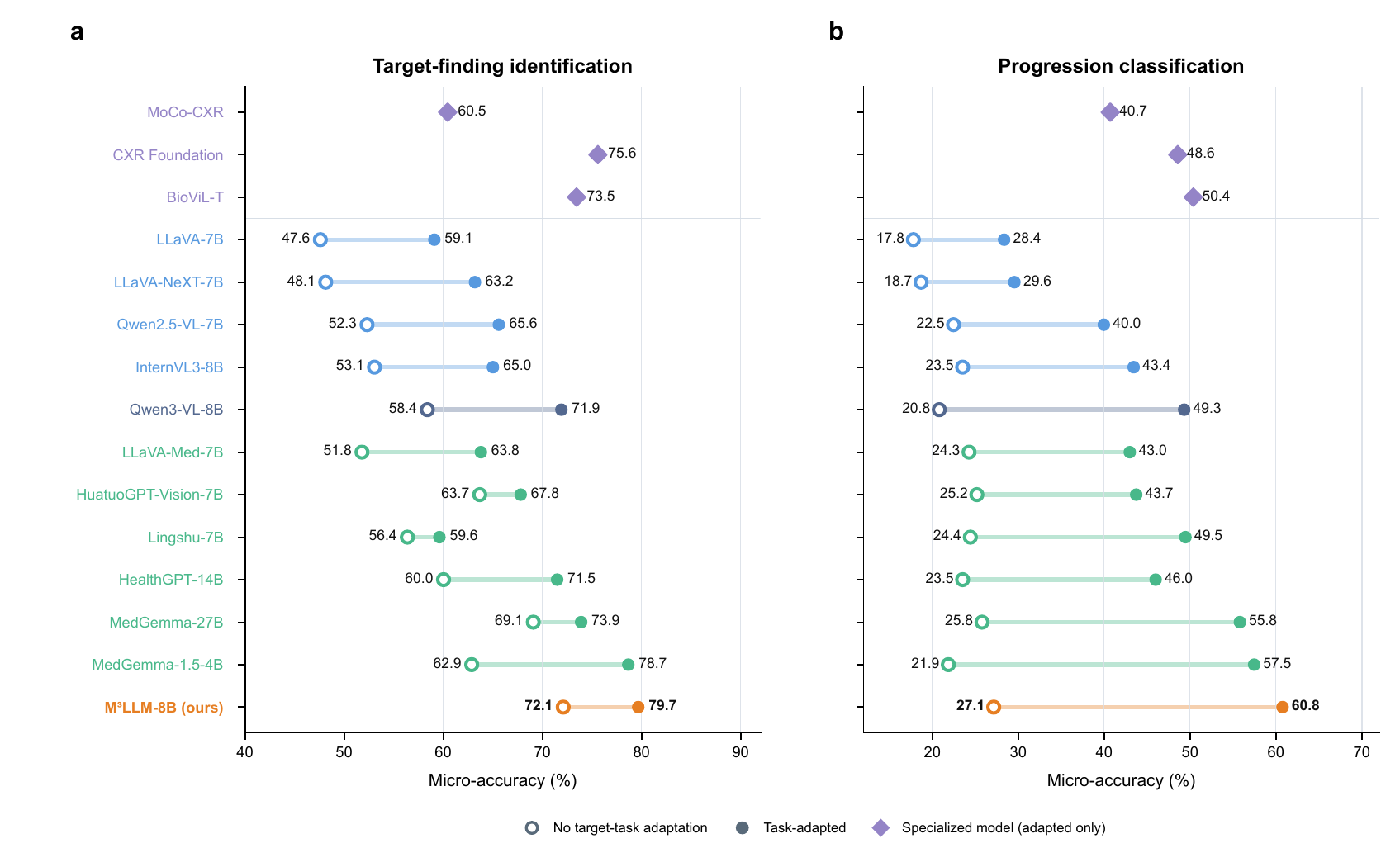}
    \caption{\textbf{Evaluation on the MS-CXR-T-derived longitudinal chest
    radiograph benchmark.} \textbf{a}, Micro-accuracy for target-finding
    identification, without target-task adaptation (open circles) and after
    target-task adaptation (filled circles); specialized chest-radiograph
    models are evaluated after adaptation only (diamonds). \textbf{b}, Micro-accuracy for progression classification under the same settings. Points denote micro-accuracy, and connecting segments link the results before and after target-task adaptation for each generative MLLM.}
    \label{fig:cxr}
\end{figure}

\noindent All uncertainty intervals in this analysis were 95\% patient-clustered bootstrap CIs.

\paragraph{Target-finding identification.}
Figure~\ref{fig:cxr}a summarizes the results for target-finding identification. Without target-task adaptation, M$^3$LLM achieved a micro-accuracy of 72.10\% (69.84--74.36),
compared with 58.40\% for its Qwen3-VL-8B backbone. Among the other generative MLLMs, MedGemma-27B achieved the next-highest micro-accuracy of 69.08\%, while MedGemma~1.5-4B achieved 62.87\%. After target-task adaptation, M$^3$LLM achieved a micro-accuracy of 79.67\% (78.26--81.00), compared with 71.92\% for the Qwen3-VL-8B backbone. MedGemma~1.5-4B achieved the next-highest micro-accuracy among the generative MLLMs at 78.67\%. Both M$^3$LLM and MedGemma~1.5-4B also exceeded the specialized chest-radiograph models evaluated for this task, including MoCo-CXR (60.45\%), CXR Foundation (75.60\%), and BioViL-T (73.45\%).

\noindent Class-balanced evaluation yielded balanced accuracies of 67.57\% without target-task adaptation and 79.03\% after adaptation for M$^3$LLM. The corresponding macro-F1 values were 63.70\% and 73.42\%, respectively (Supplementary Table~10).

\paragraph{Progression classification.}
\noindent The held-out progression set showed unequal class frequencies,
comprising 112 improving, 303 stable, and 248 worsening examples
(Supplementary Table~3).
We therefore considered balanced accuracy and macro-F1 alongside
micro-accuracy when assessing progression-classification performance.
Figure~\ref{fig:cxr}b summarizes the progression-classification results. Without target-task adaptation, M$^3$LLM achieved a micro-accuracy of 27.15\% (22.16--32.14), compared with 20.81\% for its Qwen3-VL-8B backbone. Among the other generative MLLMs, MedGemma-27B achieved the next-highest micro-accuracy of 25.79\%, followed by HuatuoGPT-Vision-7B at 25.19\%.

\noindent After target-task adaptation, M$^3$LLM achieved the highest observed micro-accuracy of 60.78\% (57.02--64.43), compared with 49.32\% for the Qwen3-VL-8B backbone. MedGemma~1.5-4B achieved the next-highest micro-accuracy among the other generative MLLMs at 57.47\%, followed by MedGemma-27B at 55.81\%. These point estimates were also higher than those of the specialized chest-radiograph models, including MoCo-CXR (40.72\%), CXR Foundation (48.57\%), and BioViL-T (50.38\%). Complete class-balanced results and class-level confusion counts are reported in Supplementary
Tables~11 and~12, respectively.

\noindent For M$^3$LLM, balanced accuracy increased from 25.91\% without target-task adaptation to 58.25\% after adaptation, and macro-F1 increased from 25.47\% to 59.10\%.

\subsection{Validation on a retrospective dermatology cohort}
\noindent Having evaluated M$^3$LLM in a fixed-pair longitudinal imaging setting, we further examined a distinct clinical setting with variable numbers of images per patient. We evaluated M$^3$LLM on a retrospective cohort of dermatology cases from Yale New Haven Hospital (YNHH), in which each case contained one or more real-world clinical photographs. Each model received the case photographs and a task-specific textual query, without additional clinical metadata. In the differential-diagnosis validity task, the model was given the clinical photographs together with a candidate diagnosis and predicted whether the candidate was annotated as a valid differential diagnosis for that case. The 595 cases were randomly divided at the patient level into an 80\% training partition ($n=473$) and a 20\% held-out test partition ($n=122$). Detailed cohort construction, task definition, and annotation procedures are described in Methods (Section~\ref{sec:ynhh_validation_methods}). Models were evaluated both without target-task adaptation and after YNHH-specific low-rank adaptation (LoRA).

\begin{figure}[H]
    \centering
    \includegraphics[width=\textwidth]{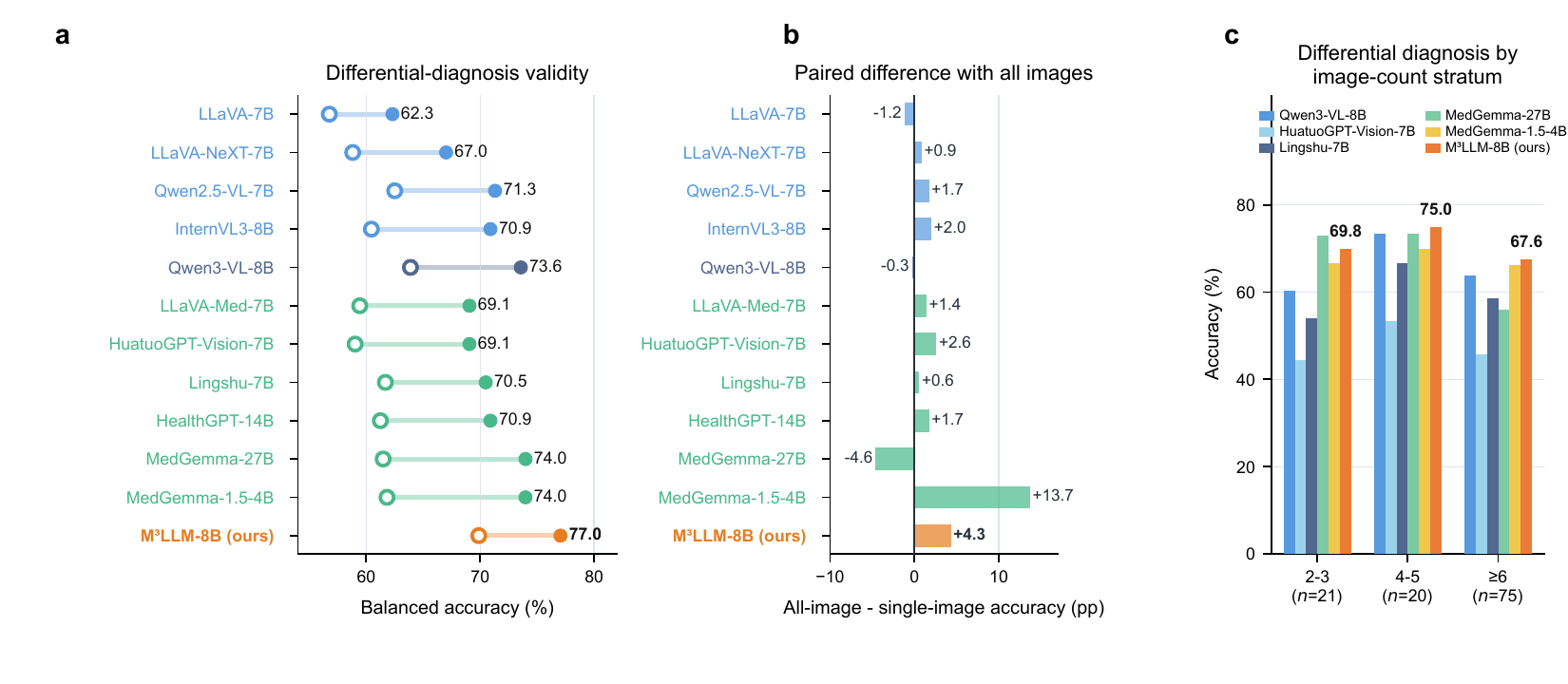}
    \caption{\textbf{Clinical validation on the retrospective Yale New Haven Hospital (YNHH) dermatology cohort.}
    \textbf{a}, Balanced accuracy for differential-diagnosis validity without target-task adaptation (open circles) and after YNHH-specific low-rank adaptation (LoRA; filled circles) in 122 held-out cases comprising 366 case--diagnosis queries. Horizontal whiskers denote 95\% case-clustered bootstrap confidence intervals, and connecting segments link the two conditions for each model.
    \textbf{b}, Paired multi-image minus single-image accuracy without target-task adaptation among 116 cases containing at least two photographs, comprising 348 queries.
    \textbf{c}, Differential-diagnosis validity accuracy without target-task adaptation stratified by the number of available photographs. The three strata contained 21, 20, and 75 cases with 2--3, 4--5, and $\geq 6$ available photographs, respectively; six representative models are shown for readability. Model input was capped at eight photographs, with approximately uniform sequence-wide subsampling when more than eight were available. The dashed line denotes the 66.7\% all-negative reference accuracy resulting from one positive and two negative queries per case.}
    \label{fig:ynhh}
\end{figure}

\paragraph{Differential-diagnosis validity.}
Figure~\ref{fig:ynhh}a summarizes the results for differential-diagnosis validity. Without target-task adaptation, M$^3$LLM achieved a balanced accuracy of 69.88\% and a macro-F1 of 67.70\%, the highest observed values among the evaluated models. The Qwen3-VL-8B backbone achieved a balanced accuracy of 63.88\% and a macro-F1 of 61.13\%. Among the remaining baselines, Qwen2.5-VL-7B achieved the highest balanced accuracy at 62.50\%, whereas MedGemma~1.5-4B achieved the highest macro-F1 at 61.18\%. After YNHH-specific LoRA adaptation, M$^3$LLM achieved a balanced accuracy of 77.05\% and a macro-F1 of 77.32\%. The Qwen3-VL-8B backbone achieved 73.57\% balanced accuracy and 72.97\% macro-F1. MedGemma-27B and MedGemma~1.5-4B shared the next-highest balanced accuracy of 73.98\%, while MedGemma~1.5-4B achieved the next-highest macro-F1 of 74.64\%. Complete results are provided in Supplementary Table~13.

\paragraph{Paired comparison of single-image and multi-image inputs.}
Dermatology cases frequently contain multiple clinical photographs, providing a natural setting to examine whether additional images contribute to model performance. We therefore conducted a paired analysis without target-task adaptation restricted to cases with at least two clinical photographs. Models received either the first photograph in the prespecified case order or up to eight photographs from the same case. For cases with more than eight photographs, eight were sampled at approximately uniform positions across the ordered sequence. Figure~\ref{fig:ynhh}b summarizes the results. Multi-image input yielded higher differential-diagnosis validity accuracy for nine of the twelve evaluated models, with positive differences of up to 13.70 percentage points. M$^3$LLM achieved the highest multi-image accuracy. Its differential-diagnosis validity accuracy was 69.25\% with multi-image input, compared with 64.94\% using a single image, a difference of 4.31 percentage points. Model-specific results are reported in Supplementary Table~14.

\paragraph{Analysis across image-count strata.}
We further examined performance without target-task adaptation across patient cases containing 2--3, 4--5, or at least 6 available clinical photographs (Figure~\ref{fig:ynhh}c). Model inputs followed the same eight-image rule, including approximately uniform subsampling when more than eight photographs were available. This analysis describes performance across naturally occurring case groups rather than within-case increments in image number. M$^3$LLM achieved differential-diagnosis validity accuracies of 69.84\%, 75.00\%, and 67.56\% across the three strata, respectively, with the highest observed values in the latter two strata. Complete results are provided in Supplementary Table~15.

\section{Discussion}

Effectively interpreting and synthesizing information across multiple
medical images is important in many clinical settings, including the
integration of examinations across imaging modalities~\cite{brin2025assessing,maleki2025multimodal,chen2021multimodal}
and the comparison of imaging examinations acquired across patient
visits to assess disease progression~\cite{carrasco2023prediction,jin2021predicting,sharma2025longitudinal}.
Although MLLMs are rapidly showing promise across a range of medical
applications, multi-image data resources and modelling approaches remain
comparatively less developed. In this study, we investigated both the
data and methodological challenges involved in developing MLLMs for
multi-image understanding and reasoning. Specifically, we investigated
whether compound figures from biomedical literature
could provide a scalable source of multi-image supervision, developed
corresponding benchmarks, proposed training approaches to improve the
integration of information across multiple medical images while preserving
broader model capabilities, and systematically evaluated the resulting
model across complementary benchmark and clinical settings.

\noindent A primary finding of this study is that medical compound figures can be systematically transformed into a scalable source of supervision for medical multi-image learning. Starting from biomedical literature and leveraging community-curated resources~\cite{kim2026medpmc}, we developed a data-construction pipeline that preserves relationships among subimages and uses their surrounding biomedical context to construct complementary supervision for both instruction tuning and policy refinement. This process yielded PMC-MI, comprising 234,956 instruction instances across multi-image,
compound-figure, text-only, and multiple-choice tasks. Among these,
10,555 multi-subimage instances retaining instruction-derived
supporting-subimage information were allocated to policy refinement to
support optimization of visual evidence selection and answer generation.
In parallel, we constructed PMC-MI-Bench as a separate, manually reviewed evaluation benchmark with source-article-level separation from PMC-MI to mitigate potential training--evaluation leakage.

\noindent Manual evaluations provided complementary evidence for the quality of the resulting supervision. Across the instruction-construction steps, mean correctness, completeness, and clarity scores ranged from 4.3--4.9, 4.4--4.9, and 4.5--5.0, respectively. Within the policy-refinement data, biomedical reasoning and final clinical correctness and safety received mean ratings of 3.96 and 3.98, whereas visual grounding and image triage received a lower mean rating of 3.35. This finding highlights the identification and selection of relevant visual evidence as a comparatively challenging component of multi-image reasoning and is consistent with prior studies reporting difficulties in identifying and prioritizing relevant medical evidence~\cite{keer2026med,kim2025rethinking}. Together, these results support the feasibility of transforming biomedical compound figures into scalable and structured supervision for medical multi-image learning.

\noindent A second main finding is that the proposed three-stage training strategy improved medical multi-image understanding while maintaining performance across broader tasks. Existing medical MLLMs have been developed largely through supervised instruction tuning, with more recent studies also exploring reinforcement-learning-based post-training~\cite{ye2025multimodal,hu2025medical,li2023llavamed,huatuogpt_vision,
xu2025lingshu,healthgpt,sellergren2025medgemma,sellergren2026medgemma}.
However, these approaches generally do not explicitly address challenges
specific to multi-image understanding. The proposed framework first adapts the model to medical multi-image instructions, subsequently uses
selection-aware reinforcement learning to optimize visual evidence
selection together with answer quality, and finally returns to supervised fine-tuning on mixed
multi-image and non-multi-image instructions to maintain broader
capabilities.

\noindent Controlled stage-wise analyses characterized the performance changes observed across the training procedure. Starting from the same Qwen3-VL-8B backbone, multi-image SFT increased the overall PMC-MI-Bench multi-image VQA score from 73.8 to 77.1, followed by a further increase to 83.1 after selection-aware reinforcement learning. To contextualize the GRPO results against additional supervised optimization, we compared this model with a continued structured-SFT condition initialized from the same Stage~I checkpoint and trained on the same policy-refinement prompts. The continued-SFT control achieved an overall multi-image VQA score of 76.4, compared with 83.1 after selection-aware reinforcement learning. The final full-mixture SFT stage retained these multi-image gains, yielding an overall score of 83.8, while also improving performance on compound-figure VQA, text-only QA, and multiple-choice VQA. The reward-ablation results were consistent with contributions from the individual training signals; after the same full-mixture consolidation stage, removing the subimage-selection reward reduced the overall multi-image VQA score from 83.8 to 80.0. These findings support a training strategy in which explicit optimization of visual evidence selection complements answer-level supervision, while subsequent full-mixture fine-tuning helps balance multi-image specialization with performance across broader task formats. These comparisons characterize training-stage effects on a fixed PMC-MI corpus; data scale, curation quality, and instruction content are represented jointly in that corpus.

\noindent A third main finding was that performance differences relative to the unadapted backbone were also observed across public benchmarks and real-world patient data. M$^3$LLM achieved the highest overall accuracy among the evaluated models on the provenance-screened OmniMedVQA set (88.6\%, compared with 77.0\% for Qwen3-VL-8B) and on MMMU-Med (68.0\%, compared with 61.3\%). Similar patterns were observed in the two clinical evaluations, which differed substantially from the PMC-derived compound-figure setting. In longitudinal chest-radiograph analysis, M$^3$LLM achieved progression-classification accuracies of 27.1\% without target-task adaptation and 60.8\% after adaptation, the highest observed values among the evaluated models under the corresponding settings. In the retrospective dermatology cohort, it achieved the highest observed balanced accuracy for differential-diagnosis validity under both evaluation settings.

\noindent Task-specific adaptation produced substantial gains in both clinical evaluations. For longitudinal progression assessment, M$^3$LLM accuracy increased from 27.1\% without adaptation to 60.8\% after adaptation. For dermatology differential-diagnosis validity, balanced accuracy increased from 69.9\% to 77.1\%. These findings support literature-derived multi-image training as a scalable initialization that can be refined using task- or domain-specific clinical data.

\noindent In the clinical evaluation, we also compared M$^3$LLM and other representative MLLMs with task-adapted domain-specific representation models widely used for chest-radiograph analysis~\cite{sowrirajan2021moco,xu2023elixr,bannur2023learning}. CXR Foundation and BioViL-T achieved progression-classification accuracies of 48.6\% and 50.4\%, respectively, after task-specific adaptation. M$^3$LLM was evaluated directly in a generative setting and reached 60.8\% after adaptation. Related comparisons between generative LLMs or MLLMs and task-specific encoder models have highlighted trade-offs in labelled-data requirements, computational demands, and downstream task characteristics~\cite{hu2024improving,cooper2025rethinking}. Future systematic evaluations across clinical specialties, applications, and computational budgets will be valuable for characterizing these trade-offs and identifying the settings in which each modelling paradigm is most appropriate.

\noindent Another important observation relates to how open-ended responses from medical MLLMs should be evaluated. Unlike multiple-choice tasks, for which accuracy can be directly calculated, the optimal evaluation of free-text MLLM responses remains an open challenge~\cite{chen2025evaluating,bedi2025testing,chen2025benchmarking}. In this study, to provide a more comprehensive evaluation, we combined reference-based automatic metrics, LLM-as-judge evaluation, and manual medical review, all of which have been widely adopted in prior studies~\cite{bedi2025testing,shool2025systematic}. The results suggest that these approaches provide complementary information. Automatic metrics provided scalable measures of similarity to the reference answers, whereas human and LLM-based ratings assessed response completeness, clarity, and correctness relative to the manually verified references. Agreement between human reviewers and LLM judges was strongest for correctness, with Spearman correlations ranging from 0.77 to 0.83, compared with 0.55--0.72 for completeness and 0.27--0.48 for clarity. To reduce potential bias in the manual evaluation, each response was independently assessed by at least two medical reviewers, and model identities were concealed. Nevertheless, expert manual evaluation remains resource intensive. Further work is needed to develop more scalable and standardized approaches for evaluating medical MLLMs, including expert evaluation that considers both the underlying medical images and the clinical significance of model errors.

\subsection*{Limitations}
\noindent Several limitations should be considered when interpreting the findings of
this study. First, PMC-MI was constructed from biomedical literature and
therefore reflects the characteristics of published medical evidence.
Although the scale and heterogeneous composition of PMC provide broad
coverage across medical domains, compound figures are curated for scientific
communication and cannot fully represent the distribution or complexity of
routine clinical imaging workflows. We partially addressed this limitation
by evaluating the resulting model beyond PMC-MI-Bench, including a provenance-screened
public benchmark, an examination-derived public benchmark, and two clinical datasets with substantially
different image sources and task formulations. Nevertheless, future work
should further expand multi-image supervision using additional clinical and
non-literature data sources and evaluate performance across broader
institutions, patient populations, specialties, and imaging settings.
Rare conditions, unusual presentations, and other long-tail subgroups may be sparsely represented in published compound figures; aggregate benchmark results therefore do not establish comparable performance in these subgroups.

\noindent Second, the construction of large-scale multi-image supervision necessarily
involves automated processing. To support scalability while maintaining data
quality, our construction pipeline was designed to accommodate different
language and multimodal models and incorporated independent manual review of
samples from both the instruction-construction and policy-refinement data. The results
showed generally high ratings across the evaluated dimensions. However,
visual grounding and image triage received comparatively lower ratings,
consistent with the broader challenge of reliably identifying and
prioritizing relevant medical evidence~\cite{keer2026med,kim2025rethinking}. Further improvements in automated
quality control and expert verification could strengthen the construction of
multi-image supervision and, in turn, support further improvements in model
development. 

\noindent The longitudinal chest-radiograph evaluation used the released MS-CXR-T annotations and therefore inherits variation in their source-label quality. These annotations were not independently re-adjudicated in the present study.

\noindent Third, throughout the study, we implemented source-article separation between PMC-MI and PMC-MI-Bench, record-level linkage through source PMCIDs and sample identifiers, source-dataset provenance screening for OmniMedVQA, and a complementary verbatim-regeneration analysis. The complete pretraining and post-training corpora of many evaluated models are not publicly available, which leaves uncertainty regarding prior model exposure to public evaluation data. This remains a general challenge in the evaluation of contemporary LLMs and MLLMs~\cite{ong2024ethical,chen2025benchmarking}. Future evaluations using newly collected datasets and broader cross-dataset similarity analyses would provide additional evidence of generalizability. We have also made the related data and code
available to facilitate independent evaluation with emerging models and
datasets and to support further development by the research community.

\noindent Fourth, the clinical evaluations were retrospective and were designed to
assess transfer to real-world medical imaging data rather than prospective
clinical utility. Although the longitudinal chest-radiograph and
retrospective dermatology evaluations extended the study beyond
literature-derived data and included both fixed-pair and variable-cardinality
multi-image settings, they covered a limited number of clinical domains and
institutions. Moreover, the improvements observed after task-specific
adaptation indicate that additional clinical supervision may remain
important for particular applications. These results should therefore not
be interpreted as establishing readiness for clinical deployment or
improvements in patient outcomes. Future studies should extend these retrospective evaluations to pilot
prospective validation and more systematic assessment in real-world clinical
workflows~\cite{brodeur2026prospective,wan2024outpatient,goh2024large}.

\section{Methods}\label{sec:methods}
\subsection{PMC-MI construction and quality control}
\label{sec:pmc_mi_construction}

\paragraph{Source data.} PMC-MI was constructed from biomedical compound figures and their associated text. PubMed Central (PMC) provides a large-scale source of medical image--text data, and several multimodal resources have previously been curated from PMC~\cite{zhang2025multimodal,lozano2025biomedica,kim2026medpmc}.
For PMC-MI, we used the initial release of MedPMC~\cite{kim2026medpmc}, which contains approximately 11 million curated medical image--text pairs extracted from 6.1 million articles in the PMC Open Access collection through June 2024. We selected MedPMC as the source resource for two main reasons. First, it preserved the source-article provenance required for article-level overlap and licence screening. Source records were matched by PMCID to official PMC metadata and retained when the source article carried a CC BY, CC BY-NC, CC BY-NC-SA, CC0, or CC BY-SA licence. Each released record includes its source PMCID, compound-figure identifier, article-level licence, attribution and reuse fields, and an evidence status; PMID and citation metadata are supplied where available in the accompanying sample-level licence evidence keyed by PMCID. Second, MedPMC provided compound-figure records, separated subimages, bounding boxes, and aligned subcaptions as fixed upstream inputs. MedPMC generated the subimage boxes using a YOLOv10 detector trained on ImageCLEF and reported a figure-separation mAP of 89.8; it separately reports validation of compound-figure identification and subimage--caption alignment~\cite{kim2026medpmc}. This allowed the present study to focus on constructing instruction data for medical multi-image understanding.

\paragraph{Data processing.}
The MedPMC snapshot used to construct PMC-MI contained 2,177,293 compound figures and defined the fixed source pool for this study. Later MedPMC releases expanded the source resource; PMC-MI was constructed exclusively from this snapshot. Compound figures can contain heterogeneous subimages, including medical images, flow charts, diagrams, and other graphical content. To enrich the corpus for predominantly medical visual content, we applied the medical-image classifier released with MedPMC to each subimage. This classifier is a fine-tuned Vision Transformer with a reported F1 score of 96.5\%, a metric distinct from the figure-separation mAP reported above~\cite{kim2026medpmc}. We retained compound figures in which at least 75\% of the constituent subimages were classified as medical images.

\noindent Eligible records also required sufficient textual support for instruction construction. Compound-figure captions were required to contain more than 50 words, and each available aligned subimage caption was required to contain at least 10 words. We then assessed each record against the input and structural requirements of the individual instruction formats. These requirements included at least two eligible subimages for multi-subimage VQA and the spatial metadata needed to derive relative-position labels. Records were routed only to formats for which the required images, captions, aligned subimage information, and task-specific metadata were available.

\noindent Generated outputs were parsed according to the task-specific schemas defined in Step~4. We retained outputs containing all required fields, including a question and reference answer, auxiliary context where applicable, and any required answer options or reference subimage identifiers. Records with generation failures, missing required fields, parsing failures, or invalid output schemas were removed before final assembly.

\noindent Following eligibility and article-licence screening, instruction construction, parsing, quality control, and source-article separation from PMC-MI-Bench, PMC-MI contained 234,956 instruction instances with unique sample identifiers linked to their source compound figures. Of these, 10,555 eligible multi-subimage VQA instances were assigned to the policy-refinement partition, comprising 10,355 training and 200 validation instances. The remaining 224,401 instances formed the supervised instruction corpus used in Stages~I and III. The construction and quality assessment of both partitions are described below.

\subsubsection{Instruction construction}
Instruction-following data were constructed from the eligible compound figures and associated text retained after the preceding processing steps. The supervised instruction partition provided the training targets for Stages~I and III, whereas the policy-refinement partition was used for Stage~II. The instruction-construction pipeline was model-agnostic at the workflow level, such that its processing steps
were not tied to a specific language-model or multimodal-model architecture. Complete prompt specifications and representative transformations are provided in Supplementary Methods~S1.1, Supplementary Figs.~1--10, and Supplementary Figs.~11--15. PMC-MI is publicly available at \href{https://huggingface.co/datasets/Yale-BIDS-Chen/M3LLM-data-v1.0.0}
{https://huggingface.co/datasets/Yale-BIDS-Chen/M3LLM-data-v1.0.0}.

\paragraph{Instruction formats.}
To provide complementary supervision across different input and task formats, four instruction families were constructed: multi-image visual question answering (VQA), compound-figure VQA, text-only question answering (QA), and multiple-choice VQA.

\noindent Multi-image VQA contained three task formats. In \emph{single-subimage VQA}, the compound figure together with its individual subimages was provided as input, while the answer was primarily supported by one relevant subimage. This format was designed to require question-conditioned localization of relevant visual evidence within a multi-image input. In \emph{multi-subimage VQA}, complementary evidence from at least two subimages was required to answer the question. In \emph{relative-position VQA}, the spatial relationship between two specified subimages in the parent compound figure was queried, including above, below, left, and right relationships. Reference spatial relationships were derived from the subimage bounding boxes provided by MedPMC; bounding-box coordinates were not required to be predicted.

\noindent Compound-figure VQA presented the compound figure as a single visual input without separately supplying its individual subimages. No visual input was provided for text-only QA. For multiple-choice VQA, either a compound-figure or multi-image input was used, depending on the constructed instruction, and an answer was selected from a predefined set of options.

\paragraph{Five-step instruction-construction pipeline.}
These instruction formats were constructed using a five-step pipeline in which source-associated textual information, medical background knowledge, and subimage-level visual descriptions were progressively assembled before task-specific instructions were constructed and their auxiliary context was refined. The five steps are detailed below.

\paragraph{Step 1: Inline-text summarization.}
Biomedical articles may refer to the same compound figure in multiple passages throughout an article. The figure-associated passages linked to each source record were first summarized. Qwen2.5-32B~\cite{team2025qwen2_5} was used in this step. Information relevant to
interpretation of the figure, including clinical presentation, pathological findings, diagnostic procedures, interventions, and outcomes when available, was prioritized by the prompt, while material unrelated to the figure was omitted. The resulting summary was used as a compact figure-level textual context for subsequent steps.

\paragraph{Step 2: Domain-specific medical knowledge complementation.}
Qwen2.5-32B~\cite{team2025qwen2_5} was used to construct concise medical background for each source record. The model first identified medically relevant keywords grounded in the original figure caption and inline-text summary and then generated up to five sentences of explanatory knowledge centred on these keywords. The generated explanation was permitted to incorporate directly relevant clinical or diagnostic background not explicitly stated in the source record, but was required to remain aligned with the extracted concepts and source-specific clinical context. Depending on the record, this background covered diseases, symptoms, pathological findings, diagnostic procedures, imaging characteristics, or treatments. It was subsequently combined with the inline-text summary and original figure caption for visual description and instruction construction.

\paragraph{Step 3: Subimage-level medical visual description.}
Each subimage was next processed separately for medical visual description. HuatuoGPT-Vision-34B~\cite{huatuogpt_vision} was used in this step. One subimage together with the available textual context was provided as input, and a description of visible content relevant to interpretation of the source record was produced. Anatomical structures, pathological findings, imaging characteristics, and other visible features were described when applicable. By processing subimages individually, the association between each subimage and its corresponding visual description was preserved before information was integrated across the compound figure.

\paragraph{Step 4: Context--question--answer instruction construction.}
Context--question--answer instructions were then constructed from the textual and visual information assembled in Steps~1--3. Qwen2.5-32B~\cite{team2025qwen2_5} was used in this step. Separate task-specific prompts were used for the instruction formats defined above, with the model role, admissible evidence, question structure, output schema, and expected answer format specified in each prompt.

\noindent For multi-image instructions, the identifiers of the subimage or subimages used to construct the question and reference answer were additionally retained. These instruction-derived subimage identifiers preserved the provenance of the visual evidence associated with each question and were subsequently used to construct the policy-refinement data for reinforcement learning, as described below.

\paragraph{Step 5: Context refinement to reduce answer leakage.}
Finally, a quality-control step was performed to reduce the possibility that the answer could be directly revealed or strongly implied by the auxiliary textual context. The context, question, and reference answer were jointly inspected, and answer-revealing information in the auxiliary context was removed or rewritten while information needed to interpret the question was retained. Only the auxiliary context was revised; the visual input, question, and reference answer were retained unchanged. Qwen2.5-32B~\cite{team2025qwen2_5} was used in this step. 

\paragraph{Instruction quality assessment.}
To assess the quality of the instruction-construction pipeline, 140 outputs spanning the five construction steps were randomly sampled. Thirty outputs were sampled from each of Steps~1--4, with five outputs per task format, and 20 were sampled from Step~5; multiple-choice and relative-position VQA were not included in Step~5 because these formats did not contain an auxiliary context for refinement. Three medical reviewers (S.A., G.M., and M.G.) participated in the evaluation. Each output received two independent ratings for correctness, completeness, and clarity using criteria adapted from prior medical AI studies~\cite{chen2025benchmarking,bedi2025testing}; 78 outputs received an additional third rating for the inter-rater reliability analysis. Inter-rater reliability was quantified separately for correctness, completeness, and clarity using intraclass correlation coefficients (ICCs). The detailed sampling, rating, and reliability procedures are provided in Supplementary Methods~S1.3.

\subsubsection{Policy-refinement data construction}\label{sec:policy_refinement_data}
PMC-MI was partitioned into 224,401 supervised instruction instances and 10,555 multi-subimage VQA instances dedicated to policy refinement.
The policy-refinement instances provided reference signals for question-conditioned visual-evidence selection and answer quality during Stage~II. We allocated 10,355 instances to policy optimization and 200 to validation. Additional implementation and quality-assessment details are provided in Supplementary Methods~S1.2 and~S1.3. The policy-refinement JSONL files are publicly available in the versioned \href{https://huggingface.co/datasets/Yale-BIDS-Chen/M3LLM-data-v1.0.0}{Yale-BIDS-Chen/M3LLM-data-v1.0.0} repository.

\paragraph{Instance selection and reformatting.}
Candidate instances were derived exclusively from the multi-subimage VQA instructions constructed in Step~4 above. 
For each instance, the question, auxiliary context when applicable, candidate subimages, reference answer, and instruction-derived reference subimage identifiers were preserved. Each instance was then reformatted to require selection of relevant visual evidence from a set of candidate images. The subimages associated with the question and reference answer were presented together with additional subimages from the same compound figure that served as distractors. The candidate subimages were shuffled, and image identifiers were reassigned such that Fig-0 denoted the complete compound figure and Fig-1 through Fig-$N$ denoted the individual subimages. In total, 10,555 policy-refinement instances were derived from 10,555 compound figures. 

\paragraph{Structured reference trajectories.}

A structured reference trajectory was constructed for each policy-refinement instance using GPT-5.5. The model was provided with the question, candidate images, optional auxiliary context, the instruction-derived reference subimage identifiers, and the Step~4 reference answer. The trajectory contained three fields:
\newline
\centerline{\texttt{<Thought>...</Thought><Selection>...</Selection><Answer>...</Answer>}}
\newline
The \texttt{<Thought>} field followed three ordered components:
\emph{image triage}, \emph{visual-feature extraction}, and
\emph{medical synthesis}. Image triage identified the subimages relevant to the question, visual-feature extraction described question-relevant findings in each selected subimage, and medical synthesis integrated these findings to support the final response.

\noindent The \texttt{<Selection>} field encoded the instruction-derived reference subimage identifiers retained from Step~4. The \texttt{<Answer>} field contained a GPT-5.5-refined reference answer that integrated the Step~4 reference answer with the supplied multimodal evidence and auxiliary context. The original Step~4 reference answer was retained as provenance metadata. During Stage~II, the instruction-derived subimage identifiers served as the reference target for the subimage-selection reward, whereas the refined reference answer served as the reference target for the answer-quality reward. The complete structured reference trajectory was also used as the supervised target for the matched continued-SFT control. The trajectory-construction and Stage~II rollout prompts are provided in Supplementary Figs.~16 and~17.

\paragraph{Policy-refinement data assessment.}
We evaluated the structured reference trajectories for visual grounding and image triage, biomedical reasoning, and final clinical correctness and safety. A simple random sample of 100 trajectories with complete local image assets was assessed by five medical reviewers (M.L., M.K.K., S.M., A.G., and S.A.). Thirty trajectories were independently evaluated by all five reviewers, and each of the remaining 70 trajectories was evaluated by two reviewers.

\noindent Reviewers applied a prespecified five-point rubric covering
\emph{visual grounding and image triage}, \emph{biomedical reasoning}, and \emph{final clinical correctness and safety}. These dimensions assessed the visual evidence supporting each trajectory, the coherence of the medical synthesis, and the clinical quality of the refined reference answer. The complete rubric, sampling procedure, and reviewer-assignment protocol are provided in Supplementary
Methods~S1.3, Supplementary Figure~18, and Supplementary Table~2.

\noindent We further conducted a secondary error-typing analysis within the same 100-trajectory sample. Trajectories with a mean reviewer score of 2 or lower in a given dimension were re-reviewed by a medical reviewer using prespecified, non-mutually-exclusive error categories. A screen-confirmed material error required both the low-score screen and confirmation of a substantive error during this secondary review. The resulting counts characterize confirmed errors among the low-score-screened trajectories rather than exhaustive error rates for the sample or corpus.

\subsection{PMC-MI-Bench construction and quality control}
\label{sec:pmc_mi_bench_construction}
\noindent In addition to PMC-MI, which provides large-scale data for model training, PMC-MI-Bench was constructed as a manually reviewed benchmark for model evaluation. The complete curation and provenance-audit procedures are provided in Supplementary Methods~S1.4. First, every candidate was independently screened by three medical reviewers (S.A., G.M., and M.G.) for source consistency, answer correctness, question clarity, and task-format validity. Candidates were retained only when all three reviewers found no issues; candidates requiring correction were excluded rather than edited or adjudicated into the benchmark. One compound-figure VQA candidate and one text-only QA candidate were excluded, yielding a 298-example benchmark. Second, source PMCIDs were compared across the retained benchmark and all supervised and policy-refinement training and validation partitions. This final audit removed 117 model-development records whose source articles occurred in the retained benchmark, comprising 112 supervised instances and five policy-optimization instances. Each retained benchmark record had a distinct sample identifier linking its source PMCID and compound-figure image. No source-article overlap remained between the final 298-example benchmark and PMC-MI. The benchmark contained 50 single-subimage VQA, 50 multi-subimage VQA, 50 relative-position VQA, 49 compound-figure VQA, 49 text-only QA, and 50 multiple-choice VQA examples.

\subsection{\texorpdfstring{M$^3$LLM}{M3LLM} training framework}\label{sec:three_stage_training}

\paragraph{Backbone and multi-image input formulation.}
M$^3$LLM was built on Qwen3-VL-8B~\cite{bai2025qwen3} without modifying its underlying multimodal architecture. No additional positional-encoding scheme or cross-image attention module was introduced. Multi-image examples were provided through the backbone's native multi-image interface, such that all supplied images and the accompanying textual context were processed within a single multimodal sequence. Each image was preprocessed independently by the native vision-processing pipeline, and the resulting image-derived representations were incorporated at the corresponding image positions in the multimodal input. Explicit image identifiers were included in the textual sequence to preserve the correspondence between visual inputs and image-specific references: Fig-0 denoted the complete compound figure, whereas Fig-1 through Fig-$N$ denoted its constituent subimages in their stored record order. The same convention was used throughout Stage~I supervised fine-tuning, Stage~II policy refinement, and Stage~III supervised fine-tuning, allowing questions, generated selections, and answers to refer consistently to individual visual inputs. Each multi-image instance contained at most four images, comprising the compound figure and up to three constituent subimages. Detailed image preprocessing, multimodal-context limits, and stage-specific input settings are reported in Supplementary Methods~S1.2.

\noindent The three-stage M$^3$LLM training framework progressively established multi-image instruction-following capability, refined evidence-grounded response generation, and consolidated these capabilities within a broader instruction mixture. Stage~I performed supervised fine-tuning on multi-image data to provide the initial multi-image adaptation. Stage~II applied GRPO-based policy refinement to optimize visual evidence selection, answer quality, and structured-output compliance. Stage~III performed full-mixture supervised fine-tuning using both multi-image and complementary instruction formats, thereby reintroducing broader task supervision after policy refinement. This final stage was designed to retain the multi-image capabilities developed during Stages~I and II while maintaining performance across the broader instruction formats. The three stages are described below, with complete implementation details provided in Supplementary Methods~S1.2.

\paragraph{Stage I: multi-image supervised fine-tuning.}
Stage~I was designed to establish multi-image instruction-following capability. Supervised fine-tuning was performed using the multi-image VQA instances of PMC-MI described above. For each training instance, the visual inputs, question, auxiliary context when applicable, and reference answer were provided. The trainable model parameters were optimized using the standard autoregressive language-modelling objective, following established supervised fine-tuning practice~\cite{caffagni2024revolution,liang2026comprehensive}.

\paragraph{Stage II: selection-aware reinforcement learning.}
Stage~II was designed to optimize question-conditioned visual-evidence selection, answer quality, and structured response generation. The final Stage~I checkpoint initialized the policy. Training used 10,355 policy-refinement instances, with the remaining 200 instances reserved for validation. Policy optimization used Group Relative Policy Optimization (GRPO)~\cite{shao2024deepseekmath}.

\noindent For each training prompt, the current policy generated multiple candidate responses in the prescribed format: \texttt{<Thought>...</Thought><Selection>...</Selection><Answer>...</Answer>}. Each prompt contained the question, candidate images, and optional auxiliary context. Within each response, the \texttt{<Thought>} field followed the sequence of image triage, subimage-specific visual interpretation, and medical synthesis. The \texttt{<Selection>} field recorded the subimage identifiers selected by the current policy, and the \texttt{<Answer>} field contained the final response generated by the current policy.

\noindent Each policy-generated response received three complementary reward signals. The instruction-derived reference subimage identifiers provided the target for the subimage-selection reward, the refined reference answer provided the target for the answer-quality reward, and a structured-format reward assessed compliance with the prescribed response schema.

\noindent (1) \textbf{Subimage-selection reward.} The subimage-selection reward, $R_{\mathrm{sel}}$, assessed whether the model identified the visual evidence relevant to the question. It was calculated as the set-level F1 score between the subimage identifiers selected by the model and the instruction-derived reference subimage identifiers:
\begin{equation}
R_{\mathrm{sel}}
=
\frac{
2\left|\mathcal{S}\cap\mathcal{S}^{*}\right|
}{
\left|\mathcal{S}\right|+\left|\mathcal{S}^{*}\right|
},
\label{eq:selection_reward}
\end{equation}
where $\mathcal{S}$ denotes the subimage set selected by the current policy and $\mathcal{S}^{*}$ denotes the instruction-derived reference subimage set retained during Step~4.

\noindent (2) \textbf{Structured-format reward.} The structured-format reward, $R_{\mathrm{fmt}}$, assessed whether the generated response followed the prescribed \texttt{<Thought>}, \texttt{<Selection>}, and \texttt{<Answer>} structure. Responses received higher rewards when the required fields were present, well-formed, and appeared in the prescribed order, with partial credit assigned to partially compliant responses. This reward assessed structural compliance rather than the semantic correctness of the intermediate reasoning. 

\noindent (3) \textbf{Answer-quality reward.}
The answer-quality reward, $R_{\mathrm{ans}}$, assessed semantic agreement between the final answer generated by the current policy and the refined reference answer. The content of the generated \texttt{<Answer>} field was compared with the corresponding refined reference answer using
BERTScore F1~\cite{zhang2019bertscore}:
\begin{equation}
R_{\mathrm{ans}}
=
\operatorname{BERTScore}_{F_1}(a,a^{*}),
\label{eq:answer_reward}
\end{equation}
where $a$ denotes the answer generated by the current policy and $a^{*}$ denotes the refined reference answer.

\noindent The three reward signals were scaled to the range $[0,1]$ and combined using a weighted linear function:
\begin{equation}
R =
\lambda_{\mathrm{ans}}R_{\mathrm{ans}}
+
\lambda_{\mathrm{sel}}R_{\mathrm{sel}}
+
\lambda_{\mathrm{fmt}}R_{\mathrm{fmt}}.
\label{eq:combined_reward}
\end{equation}
The final coefficients were $(\lambda_{\mathrm{ans}},\lambda_{\mathrm{sel}},\lambda_{\mathrm{fmt}})=(0.6,0.3,0.1)$. They were selected using the held-out Stage~II validation set as described below; PMC-MI-Bench was not used for reward-operator selection.

\paragraph{Stage III: full-mixture supervised fine-tuning.}
Stage~III was initialized from the policy-refined model obtained after
Stage~II and returned to supervised fine-tuning. Consistent with Stage~I, the model was optimized using the standard autoregressive language-modelling objective. Unlike Stage~I, however, Stage~III used the full supervised instruction component of PMC-MI rather than only the multi-image subset. This stage was designed to retain the multi-image capabilities acquired in the first two stages while maintaining performance across the broader task formats represented in PMC-MI.

\subsection{Training controls and reward ablations}
\label{sec:controlled_training_analysis}

\noindent Two sets of controlled comparisons were used to characterize the training framework. The first compared continued structured SFT with GRPO after initialization from the same Stage~I checkpoint, whereas the second compared the full GRPO reward formulation with one-component-removed variants.

\paragraph{Checkpoint-level comparison of staged training.}
The checkpoint-level analysis included the pretrained Qwen3-VL-8B model, the Stage~I checkpoint $\pi_{\mathrm{MI\text{-}SFT}}$, and four subsequent training conditions. For the Stage~II comparison, continued structured SFT and GRPO were initialized from $\pi_{\mathrm{MI\text{-}SFT}}$ and received the same prompt set using the same structured response schema. Continued structured SFT optimized the likelihood of one fixed structured reference trajectory per prompt (Section~\ref{sec:policy_refinement_data}), comprising the instruction-derived subimage selection, structured intermediate reasoning, and the refined reference answer. GRPO instead generated 16 on-policy responses per prompt and optimized the reward-based objective defined above.

\noindent The two conditions were matched for the number of prompt presentations, effective global batch size, number of optimizer updates, checkpointing schedule, and checkpoint-selection rule. They differed in optimization objective, trajectory generation, on-policy sampling, and total forward-pass computation. For both conditions, the comparison checkpoint was selected using the best score on the held-out 200-example Stage~II validation set. PMC-MI-Bench was not used for checkpoint selection.

\noindent To evaluate Stage~III, the Stage~II checkpoint was subsequently trained under two supervised data configurations: multi-image-only supervised fine-tuning (MI-only SFT), which used the supervised multi-image instruction set alone, and full-mixture supervised fine-tuning (Full-SFT), which used the complete supervised instruction mixture. The MI-only condition therefore tested whether the Stage~III gains on non-multi-image formats required broadening the mixture or followed from additional supervised training of any kind. Instructions allocated to Stage~II were excluded from both configurations. The two conditions used the same supervised training configuration and differed only in the instruction data. For each condition, the checkpoint obtained after the prespecified training schedule was used for evaluation.

\paragraph{Reward-component ablations.}
We further evaluated the composite Stage~II reward by independently disabling the subimage-selection reward ($R_{\mathrm{sel}}$), structured-format reward ($R_{\mathrm{fmt}}$), or answer-quality reward ($R_{\mathrm{ans}}$). Each ablation was initialized from the same $\pi_{\mathrm{MI\text{-}SFT}}$ checkpoint and retained the same training prompts, response schema, rollout group size, generation configuration, effective global batch size, and number of optimizer updates as the full-reward condition. Only the corresponding reward component was disabled; the remaining reward and optimization settings were unchanged.

\noindent After Stage~II, the three ablated variants and the full-reward model underwent an identical Stage~III pass using the same supervised instruction mixture and training configuration, so that the comparison isolated the reward formulation. All checkpoint and reward-ablation variants were evaluated on the same PMC-MI-Bench examples using an identical inference and scoring protocol. Overall MI-VQA was calculated as the unweighted mean across the single-subimage, multi-subimage, and relative-position VQA categories.

\paragraph{Sensitivity to the reward aggregation operator.}
The reward aggregation operator was selected from five prespecified coefficient combinations using only the randomly sampled 200-instance Stage~II validation set. Mean answer-quality reward was the primary selection criterion, whereas subimage-selection F1 and valid-format rate were examined as secondary diagnostics. PMC-MI-Bench was not used for operator or checkpoint selection. The complete candidate set and selection protocol are provided in Supplementary Methods~S1.2.

\subsection{Evaluation models and common protocols}
\noindent M$^3$LLM was evaluated across complementary settings, including PMC-MI-Bench, two additional public benchmarks, and two clinical cohorts. The datasets, baseline models, evaluation settings, and metrics used for each analysis are described below.

\subsubsection{Evaluation on PMC-MI-Bench}

\paragraph{Baseline models.}
Representative general-purpose and medical MLLMs that have been commonly used in prior studies were included as baselines. The general-purpose baselines included LLaVA-7B~\cite{llava}, LLaVA-NeXT-7B~\cite{llava_next}, Qwen2.5-VL-7B~\cite{team2025qwen2_5}, InternVL3-8B~\cite{chen2024internvl}, and Qwen3-VL-8B~\cite{bai2025qwen3}. Notably, Qwen3-VL-8B served as the backbone of M$^3$LLM and therefore provided a direct comparison for assessing the effect of the proposed training framework. The medical baselines included LLaVA-Med-7B~\cite{li2023llavamed}, HuatuoGPT-Vision-7B~\cite{huatuogpt_vision}, Lingshu-7B~\cite{xu2025lingshu}, HealthGPT-14B~\cite{healthgpt}, MedGemma-27B~\cite{sellergren2025medgemma}, and MedGemma~1.5-4B~\cite{sellergren2026medgemma}.

\paragraph{Evaluation setting.}
All models were evaluated without task-specific adaptation (i.e., zero-shot evaluation); no PMC-MI-Bench examples were used for fine-tuning. The same inference configuration was applied across models. Inference used greedy decoding with a temperature of 0, a repetition penalty of 1.0, and a maximum of 2,048 newly generated tokens. The same generation settings were applied to all evaluated models, while each model retained its native image-preprocessing procedure. The complete evaluation prompts and additional inference details are provided in Supplementary Methods~S1.5.1 and Supplementary Figure~19.

\paragraph{Evaluation measures.}
PMC-MI-Bench consists of six task categories, five of which require free-text responses and one of which uses a multiple-choice format. For the five free-text tasks, generated responses were compared with the reference answers using four commonly used automatic reference-based measures: Semantic Textual Similarity (STS)~\cite{reimers-2019-sentence-bert}, BLEU~\cite{papineni2002bleu}, ROUGE-L~\cite{lin2004rouge}, and BERTScore~\cite{zhang2019bertscore}. For multiple-choice VQA, performance was assessed using accuracy based on
the predicted and reference answer options. The definitions and implementation details of these measures are provided in Supplementary Methods~S1.5.1.

\noindent In addition to automatic reference-based measures, we further adopted an LLM-as-judge evaluation as a complementary assessment of response quality, following its increasing use in recent LLM evaluation studies~\cite{kim2024prometheus,zheng2023judging,croxford2025evaluating}. Three independent LLM judges were used: DeepSeek-V4-Pro~\cite{xu2026deepseek}, GPT-5.6~\cite{openai2026gpt56}, and Claude Opus 4.8~\cite{anthropic2026claudeopus48}. Each judge was provided with the question, the reference answer, and an anonymized candidate response, and independently rated the response for completeness, clarity, and correctness. Detailed judge configurations, evaluation rubrics, and prompts are provided in Supplementary Methods~S1.5.2 and Supplementary Figure~20.

\noindent While reference-based and LLM-based evaluations enable scalable assessment, we further performed manual review to provide an independent human evaluation of response quality. Seventy questions were randomly sampled from the five open-ended task categories. For each question, responses from M$^3$LLM, MedGemma-27B, and Lingshu-7B were evaluated, yielding 210 response items. Five medical reviewers (M.L., M.K.K., S.M., A.G., and S.A.) independently applied the same completeness, clarity, and correctness criteria. To reduce potential evaluation bias, the responses were anonymized so that the reviewers were unaware of the models that generated them. Thirty response items were independently reviewed by all five medical reviewers to assess inter-rater agreement, whereas each of the remaining 180 response items was independently reviewed by two medical reviewers. Additional details are provided in Supplementary Methods~S1.5.2.

\subsubsection{Evaluation on public medical benchmarks}

In addition to PMC-MI-Bench, we evaluated all models on two public medical benchmarks, OmniMedVQA~\cite{hu2024omnimedvqa} and MMMU-Med~\cite{yue2024mmmu}. Both benchmarks have been used in recent evaluations of medical MLLMs~\cite{xu2025lingshu,gu2026evaluating,xia2026mmedagent,chen2024towards}. Before inference and scoring, we reviewed the provenance of their constituent source datasets, as detailed below.

\noindent The complete OmniMedVQA benchmark contains 118,010 images and 127,995 question--answer items derived from 73 medical datasets, spanning 12 imaging modalities and more than 20 anatomical regions. The fully open-access pool comprised 88,996 question--answer items from 42 source datasets. We reviewed the provenance of these source datasets and excluded four subsets---Covid CT (199 items), CoronaHack (684), Covid19 heywhale (690), and COVIDx CXR-4 (485)---because their documented sources indicated potential inclusion of images derived from scientific publications or preprints. The retained subsets were primarily derived from institutional collections or established challenge datasets rather than directly from scientific publications or preprints. The exclusion was applied by source-dataset label before inference and scoring, leaving 86,938 items from 38 source datasets across eight modalities: computed tomography, magnetic resonance imaging, X-ray, ultrasound, dermoscopy, fundus photography, optical coherence tomography, and microscopy.

\noindent For MMMU-Med, we used the \texttt{MMMU-Medical-val} task implemented in MedEvalKit, corresponding to the validation split of the official MMMU release. This subset contains 150 questions from the Health and Medicine discipline, with 30 questions from each of five subjects: Basic Medical Science, Clinical Medicine, Diagnostics and Laboratory Medicine, Pharmacy, and Public Health. The questions were originally collected from college-level examinations, quizzes, and textbooks and require domain-specific multimodal perception and reasoning. All 150 validation questions were retained for evaluation.

\noindent We applied the PMC-MI-Bench zero-shot evaluation protocol to both public benchmarks. Within each benchmark, all applicable baseline models were evaluated on the same benchmark instances using identical prompts, inference settings, and answer-extraction procedures. Both benchmarks comprise multiple-choice questions, and performance was measured by accuracy. Benchmark descriptions and detailed evaluation configurations are provided in Supplementary Methods~S1.5.3.

\subsubsection{Evaluation on longitudinal chest radiographs}

Beyond the public benchmarks, we further evaluated the models using real-world clinical datasets. We first used MS-CXR-T~\cite{bannur2023mscxr}, a temporal chest-radiograph dataset derived from MIMIC-CXR~\cite{johnson2019mimic}. MS-CXR-T contains 1,326 temporally ordered chest-radiograph pairs annotated for the progression of five radiographic findings: consolidation, edema, pleural effusion, pneumonia, and pneumothorax. Each example consists of a prior radiograph, a subsequent radiograph from the same patient, an
annotated target finding, and a progression label of \emph{improving}, \emph{stable}, or \emph{worsening}. The image pairs represent temporally ordered examinations and were not assumed to correspond specifically to pre-treatment and post-treatment imaging. This setting reflects a common clinical scenario in which findings are compared and synthesized across longitudinal examinations. We used the progression labels released with MS-CXR-T without re-deriving or re-adjudicating them; the source annotation and label-quality procedures are detailed in Supplementary Methods~S1.5.5.

\noindent Two evaluation tasks were constructed from MS-CXR-T. In \emph{progression classification}, the model received the prior and subsequent radiographs together with the target finding and predicted whether the finding was improving, stable, or worsening. In \emph{target-finding identification}, the model received the same image pair together with a queried finding and generated a binary \emph{Yes}/\emph{No} response. Five queries were constructed for each image pair, one for each of the five findings. The finding associated with the original MS-CXR-T annotation was treated as the positive target, whereas the remaining four findings were treated as non-target labels. Because these non-target labels were derived from the dataset annotations rather than independent verification of clinical absence, this task evaluated identification of the annotated target finding rather than general finding-presence classification.

\paragraph{Baseline models.}
We evaluated the general-purpose and medical MLLM baselines described above. We additionally included three domain-specific chest-radiograph representation models. MoCo-CXR is a ResNet-18 image encoder pretrained on unlabelled chest radiographs using momentum contrastive learning~\cite{sowrirajan2021moco}. CXR Foundation (ELIXR-C) is a language-aligned chest-radiograph encoder that produces image embeddings learned from radiographs and their associated reports~\cite{xu2023elixr}. BioViL-T is a temporal vision--language model with a multi-image encoder designed to jointly represent prior and current chest radiographs~\cite{bannur2023learning}.

\paragraph{Evaluation setting.}
The generative MLLMs were evaluated both without and with task-specific adaptation. The 1,326 image-pair examples were divided at the patient level into training and held-out test partitions, such that all image pairs from the same patient were assigned to the same partition. The resulting partitions each contained 663 examples. The held-out test set therefore contained 663 progression-classification examples and 3,315 target-finding-identification queries.

\noindent Without task-specific adaptation, the MLLMs were evaluated directly on the held-out test set. With task-specific adaptation, each MLLM was fine-tuned using only the training partition and subsequently evaluated on the same held-out test set. The same adaptation protocol, prompts, inference settings, and evaluation examples were applied across the MLLMs. The domain-specific chest-radiograph models were evaluated only with task-specific adaptation because they do not support generative instruction following.
The task-specific heads used for the three specialized baselines were developed for the present evaluation, with optimization settings informed by the corresponding original studies. Additional details and the evaluation prompt are provided in Supplementary Methods~S1.5.5 and Supplementary Figure~21.

\paragraph{Evaluation measures.}
Performance was evaluated separately for target-finding identification and progression classification using accuracy, balanced accuracy, and macro-averaged F1 score (macro-F1). Accuracy was defined as the proportion of predictions assigned the correct class. Balanced accuracy was the unweighted mean of class-specific recall values, and macro-F1 was the unweighted mean of class-specific F1 scores. Metrics were calculated over all 3,315 binary target-finding decisions or all 663 three-class progression decisions in the held-out test partition, as applicable.

\subsubsection{Evaluation on a retrospective dermatology cohort}
\label{sec:ynhh_validation_methods}
\noindent As a second evaluation using a real-world clinical dataset, we used a retrospective cohort of de-identified dermatology patients from Yale New Haven Hospital (YNHH). The dataset and annotation procedures are described in detail in~\cite{jiang2026multimodal}. Each patient record consists of one or more clinical photographs collected during routine care. The use of these data for the present evaluation was conducted under an exemption granted by the Yale institutional review board (protocol no.~2000041054).

\noindent We evaluated \emph{differential-diagnosis validity}, which assessed, given the patient's
clinical images, whether a queried disease constituted a clinically valid differential diagnosis for the presented case. The task included 595 cases and a vocabulary of 11 canonical diagnostic categories. For each case, the expert-annotated primary diagnosis was used to construct one positive query, and two diagnoses sampled from the remaining categories were used to construct candidate negative queries. Each candidate negative diagnosis was subsequently reviewed by medical experts and retained only when it was confirmed not to constitute a valid differential diagnosis for the corresponding case.

\paragraph{Baseline models and evaluation setting.}
The same general-purpose and medical MLLM baselines described above were used for comparison.
All photographs were retained during cohort construction. Model training and inference accepted up to eight photographs per case; when more than eight were available, eight were sampled at approximately uniform positions across the ordered sequence. The models were evaluated both without and with task-specific adaptation. Training and held-out test partitions were divided at the patient level. Without task-specific adaptation, the models were evaluated directly on the held-out test set. With task-specific adaptation, each model was fine-tuned using only the corresponding training partition and subsequently evaluated on the same held-out test set.

\paragraph{Evaluation measures.}
The task was evaluated as binary classification. Performance was assessed using
accuracy, balanced accuracy, and macro-averaged F1 score (macro-F1).
Accuracy was defined as the proportion of predictions assigned the
correct class. Balanced accuracy was the unweighted mean of
positive-class recall and negative-class recall, corresponding to
sensitivity and specificity, respectively. Macro-F1 was the unweighted
mean of the F1 scores calculated separately for the positive and negative classes. The positive and negative classes corresponded to
\emph{Yes} and \emph{No}, respectively.

\noindent Metrics were calculated over 366 \textit{case--diagnosis} queries from 122 held-out cases. Where uncertainty intervals were estimated, resampling was performed at the case level; the three diagnosis queries derived from the same case were retained as a single cluster. Additional details on cohort construction, task definition, data splitting, adaptation, and evaluation are provided in Supplementary Methods~S1.5.6.

\subsection{Potential data leakage and memorization analysis}
\label{sec:memorization_analysis}

\noindent Mitigating potential training--evaluation leakage was considered throughout the study. First, PMC-MI and PMC-MI-Bench were separated by comparing source PMCIDs across all supervised and policy-refinement training and validation partitions. This audit removed 117 model-development records whose source articles occurred in the retained benchmark. Each retained benchmark record also had a distinct sample identifier linking its source PMCID and compound-figure image, and no source-article overlap remained between the final benchmark and the model-development partitions. The provenance-based audit was designed to establish source-article-level separation and did not attempt exhaustive image-content or near-duplicate matching across different PMC articles. Second, we reviewed source-dataset provenance for the public benchmarks OmniMedVQA and MMMU-Med. For OmniMedVQA, we excluded four source subsets with potential publication- or preprint-derived content before inference and scoring. The retained OmniMedVQA subsets and MMMU-Med were not identified in their published documentation as being derived from PMC or explicitly PMC-derived resources.
Third, we evaluated M$^3$LLM in two clinical settings not derived from PMC-MI: the MS-CXR-T-derived longitudinal chest-radiograph evaluation and the retrospective YNHH dermatology cohort. These evaluations provided additional assessment on clinically distinct data distributions outside the literature-derived training resource.

\noindent Nevertheless, the complete pretraining and post-training corpora of
many evaluated backbones are not publicly available, and potential prior data
exposure therefore cannot be completely excluded. This remains a broader
challenge in contemporary LLM evaluation~\cite{ong2024ethical,chen2025benchmarking}.
To examine whether the performance gains of M$^3$LLM on PMC-MI-Bench were
associated with increased verbatim regeneration, we compared M$^3$LLM with its
Qwen3-VL-8B backbone, which had not undergone the proposed training on PMC-MI.
Both models were evaluated on PMC-MI-Bench without task-specific adaptation.

\noindent Following established approaches for quantifying memorization in LLMs~\cite{carlini2022quantifying,li2026memorization}, we predefined two verbatim-match thresholds, \texttt{exact@30} and \texttt{exact@50}. For a generated response and its corresponding reference text, \texttt{exact@}$K$ was considered positive if the generated response contained an identical contiguous sequence of at least $K$ tokens from the reference text, where $K\in\{30,50\}$. Generated and reference texts were processed using the same tokenizer. Four open-ended PMC-MI-Bench task categories were included: single-subimage VQA, multi-subimage VQA, compound-figure VQA, and text-only QA. Relative-position VQA was excluded because its reference answers were shorter than the 30-token minimum required for this analysis, whereas multiple-choice VQA was excluded because it did not require free-text generation. We evaluated M$^3$LLM and Qwen3-VL-8B using two complementary analyses.

\paragraph{Prefix-conditioned continuation.}
For each eligible example, each model received the benchmark question followed by the first $\ell$ tokens of the reference answer, where $\ell\in\{50,100,150\}$, and was prompted to continue the answer. The generated continuation was compared only with the remaining reference-answer suffix. A continuation satisfied \texttt{exact@}$K$ if it contained an identical contiguous sequence of at least $K$ tokens from the suffix. Only reference answers containing sufficient tokens to provide both the specified prefix and at least $K$ remaining tokens were eligible. Prefixes, continuations, and reference suffixes were processed using the same tokenizer.

\paragraph{Question-only response.}
We additionally analysed the complete responses generated by M$^3$LLM and Qwen3-VL-8B during the primary PMC-MI-Bench evaluation, without providing any reference-answer prefix or introducing additional prompting or truncation. Empty reasoning tags were removed before tokenization. Each generated response was compared with the complete reference answer using the predefined \texttt{exact@30} and \texttt{exact@50} criteria. Reference answers shorter than the corresponding threshold were excluded, yielding 198 eligible examples for \texttt{exact@30} and 195 for \texttt{exact@50}, derived from 168 source articles. Match rates were calculated separately for the four eligible task categories and averaged with equal weight.

\noindent Differences between M$^3$LLM and Qwen3-VL-8B were evaluated using 10,000 paired cluster-bootstrap replicates at the source-article level, with source articles sampled with replacement (seed 42)~\cite{carletti2025multimodal,pelt2024building}. The same task-macro statistic and paired between-model difference were recalculated in each replicate. Two-sided 95\% confidence intervals were calculated from the resulting bootstrap distributions.

\section{Data availability}

The PMC-MI data records and PMC-MI-Bench are available in the versioned Hugging Face dataset repository
\href{https://huggingface.co/datasets/Yale-BIDS-Chen/M3LLM-data-v1.0.0}{Yale-BIDS-Chen/M3LLM-data-v1.0.0}. The released PMC-MI corpus contains 234,956 instruction instances: 224,401 supervised instruction records, 10,355 policy-optimization records, and 200 policy-validation records. Each released record includes a PMCID, compound-figure identifier, article-level licence, attribution and reuse fields, and an evidence status; PMID and citation metadata are supplied where available in the sample-level licence evidence keyed by PMCID. Licence eligibility was assessed at the source-article level. The release records this scope in the \texttt{figure\_rights\_audit\_status} field; figure-specific reuse remains governed by any source-specific notices. Training records reference source-image filenames in the MedPMC source-image repository, while the images required for PMC-MI-Bench are packaged with the benchmark. PMC-derived images and text remain governed by their recorded source licences, and project-generated annotations are released under the terms specified in the repository data-licence statement.

\noindent The MS-CXR-T-derived evaluation data originate from MIMIC-CXR and are not redistributed by this study. Access to the underlying radiographs requires credentialed access through PhysioNet and compliance with the applicable data-use agreement. The YNHH dermatology images cannot be made
publicly available because of patient-privacy and institutional
data-governance requirements. OmniMedVQA and MMMU-Med are publicly available through their respective original repositories, as described in the cited studies.

\section{Code availability}

The source code and configurations for data construction, model training, inference, and the reported benchmark and clinical evaluations are publicly available under the versioned GitHub tag
\href{https://github.com/Yale-BIDS-Chen-Lab/M3LLM-v2}
{Yale-BIDS-Chen-Lab/M3LLM-v2 v1.0.2}.
The repository includes the five-step instruction-construction scripts and prompts, configurations for the three-stage M$^3$LLM training procedure, and the reported benchmark analyses. It also provides the exact response-quality rubric and an open-weight judge runner configured for \path{deepseek-ai/DeepSeek-V4-Flash}. The M$^3$LLM model checkpoint will be released at
\href{https://huggingface.co/Yale-BIDS-Chen/M3LLM-v2}
{huggingface.co/Yale-BIDS-Chen/M3LLM-v2} before final publication.

\section{Acknowledgements}
This work was supported by the National Library of Medicine of the National Institutes of Health under award number R01LM014604.

\section{Author contributions}
Z.C., Y.F., and Q.C. conceived and designed the study. Z.C. and Y.F. contributed to data construction, model development, implementation, and experimental evaluation. S.A., G.M., and M.G. conducted the instruction quality assessment and PMC-MI-Bench construction and quality control. M.L., M.K.K., S.M., A.G., and S.A. conducted the human evaluation of PMC-MI-Bench responses and the quality assessment of the policy-refinement data. R.Z., Y.P., R.J., H.K., and H.X. contributed to data analysis, experimental evaluation, and interpretation of the results. All authors reviewed and revised the manuscript and approved the final version.  

\section{Competing interests}
The authors declare no competing interests.

\newpage 

\section{References}
\begin{spacing}{0.9}
\bibliographystyle{naturemag}
\bibliography{ref}
\end{spacing}

\end{document}

\hypersetup{pageanchor=false}

\begin{titlepage}
\centering
\vspace*{1.7cm}
{\LARGE\bfseries From Compound Figures to Medical Multi-image Reasoning:\\[4pt]
Scaling Multimodal Large Language Models with Biomedical Literature\par}
\vspace{1.0cm}
{\Large Supplementary Information\par}
\vspace{1.1cm}
{\large
Zhen Chen, Yihang Fu, Rong Zhou, Serina Applebaum, Min Kyu Kim,\\
Aidan Gilson, Morten Lee, Salahudeen Mirza, Gabriel Madera,\\
Mauro Giuffre, Yuanting Pan, Roy Jiang, Hyunjae Kim, Hua Xu,\\
and Qingyu Chen\par}
\vfill
\end{titlepage}

\hypersetup{pageanchor=true}
\pagenumbering{roman}
\tableofcontents
\clearpage
\pagenumbering{arabic}

\makeatletter
\setlength{\@fptop}{0pt}
\setlength{\@fpsep}{14pt}
\setlength{\@fpbot}{0pt plus 1fil}
\makeatother
\renewcommand{\topfraction}{0.95}
\renewcommand{\bottomfraction}{0.95}
\renewcommand{\textfraction}{0.06}
\renewcommand{\floatpagefraction}{0.80}
\setlength{\textfloatsep}{16pt plus 3pt minus 3pt}
\setlength{\floatsep}{14pt plus 3pt minus 3pt}

\setcounter{figure}{0}
\renewcommand{\thefigure}{\arabic{figure}}
\renewcommand{\figurename}{Supplementary Figure}
\setcounter{table}{0}
\renewcommand{\thetable}{\arabic{table}}
\renewcommand{\tablename}{Supplementary Table}
\setcounter{section}{0}
\renewcommand{\thesection}{S\arabic{section}}
\renewcommand{\thesubsection}{\thesection.\arabic{subsection}}
\renewcommand{\thesubsubsection}{\thesubsection.\arabic{subsubsection}}
\setcounter{secnumdepth}{3}
\setcounter{tocdepth}{2}

\lstdefinestyle{suppprompt}{
    language={},
    basicstyle=\ttfamily\footnotesize,
    keywordstyle=\color{black},
    stringstyle=\color{black},
    commentstyle=\color{black},
    numbers=none,
    frame=single,
    framerule=0.25pt,
    rulecolor=\color{gray!45},
    backgroundcolor=\color{white},
    breaklines=true,
    breakatwhitespace=true,
    columns=fullflexible,
    keepspaces=true,
    showstringspaces=false,
    xleftmargin=0.5em,
    xrightmargin=0.5em,
    aboveskip=0.4em,
    belowskip=0.4em
}

\section{Supplementary Methods}
\label{sec:supplementary_methods}
\noindent The Supplementary Methods provide implementation and quality-control details that support the workflow summarized in the main Methods. They are organized in the same order as the study: instruction construction, three-stage training, manual data-quality assessment, benchmark curation, and evaluation protocols.

\noindent PMC-MI contained 234,956 instruction instances with unique sample identifiers linked to their source compound figures. Benchmark candidates were assembled separately and underwent the source-provenance and record-identifier checks described in Supplementary Methods~\ref{sec:supp_pmc_mi_bench_curation}.

\subsection{Instruction-construction prompts and representative examples}
\label{sec:supp_instruction_prompts}
\noindent Complete prompt specifications for the five-step instruction-construction pipeline are presented as searchable, copyable text in Supplementary Figs.~\ref{fig:prompt-step1}--\ref{fig:prompt-step5}. Each specification defines the model role, supplied fields, construction task, constraints, and output schema. Synthetic examples illustrate the expected input--output mapping. Source-derived transformations from the construction pipeline are provided separately in Supplementary Figs.~\ref{fig:stage1}--\ref{fig:stage5}.

\subsubsection{Prompt templates}
\noindent Step~4 used separate templates for the six instruction formats so that the target visual input, admissible construction evidence, question structure, and response format were defined for each task.

\begin{figure*}[!htbp]
    \centering
    \begin{minipage}{0.96\linewidth}
\begin{lstlisting}[style=suppprompt]
ROLE
You are a biomedical assistant specializing in medical imaging literature.

INPUT
Inline text: <article passages linked to the compound figure>

TASK
Summarize the medically relevant observations, analyses, and conclusions that
are directly associated with the compound figure.

REQUIREMENTS
1. Preserve the meaning and level of certainty of the source text.
2. Include only information supported by the supplied passages.
3. Remove repetition, publication-process details, and unrelated content.
4. Use clear, formal medical language and no more than 100 words.

OUTPUT
Inline summary: <summary>

SYNTHETIC FORMATTING EXAMPLE
Input: Axial and coronal chest CT views show a well-circumscribed pulmonary
nodule in the left upper lobe. No ground-glass component is described, and the
surrounding lung parenchyma is reported as unremarkable. Follow-up imaging was
recommended to assess interval stability.
Output: Inline summary: Axial and coronal chest CT views show a
well-circumscribed left upper-lobe pulmonary nodule without a reported
ground-glass component. The surrounding lung parenchyma is unremarkable, and
follow-up imaging was recommended to assess stability.
\end{lstlisting}
    \end{minipage}
\caption{\textbf{Step 1 prompt specification: inline-text summarization.} The synthetic example preserves the observations and level of certainty in the source passages.}
\label{fig:prompt-step1}
\end{figure*}

\begin{figure*}[!htbp]
\centering
\begin{minipage}{0.96\linewidth}
\begin{lstlisting}[style=suppprompt]
ROLE
You are a biomedical assistant specializing in medical imaging literature.

INPUT
Compound-figure caption: <caption>
Inline summary: <inline_summary>

TASK
Extract medically important keywords and provide concise background knowledge
that helps interpret the supplied material.

REQUIREMENTS
1. Select keywords supported by the caption or inline summary.
2. Provide established clinical or scientific background rather than a case-
   specific diagnosis.
3. Preserve uncertainty and avoid causal, prognostic, or management claims that
   are not supported by the input.
4. Limit the medical-knowledge note to five sentences.

OUTPUT
Keywords: <comma-separated keywords>
Medical knowledge: <up to five sentences>

SYNTHETIC FORMATTING EXAMPLE
Input caption: Axial and coronal chest CT views of a solitary pulmonary nodule.
Input inline summary: The nodule is well circumscribed and located in the left
upper lobe; follow-up imaging was recommended to assess stability.
Output keywords: pulmonary nodule, chest CT, nodule morphology, interval growth
Output medical knowledge: Pulmonary-nodule assessment integrates size,
attenuation, morphology, interval growth, and patient-level risk factors.
Smooth margins can lower suspicion but do not establish benignity. Comparison
with prior imaging and guideline-based follow-up may help determine stability.
\end{lstlisting}
\end{minipage}
\caption{\textbf{Step 2 prompt specification: domain-specific medical knowledge complementation.} The synthetic example provides established background relevant to the source record while preserving diagnostic uncertainty.}
\label{fig:prompt-step2}
\end{figure*}

\begin{figure*}[!htbp]
\centering
\begin{minipage}{0.96\linewidth}
\begin{lstlisting}[style=suppprompt]
ROLE
You are a biomedical multimodal assistant specializing in medical imaging.

INPUT
Subimage: <subimage>
Subimage caption: <subimage_caption>
Compound-figure caption: <compound_figure_caption>
Inline summary: <inline_summary>
Medical knowledge: <medical_knowledge>

TASK
Describe the medically relevant visual features of the supplied subimage.

REQUIREMENTS
1. Ground every visual statement in the supplied subimage.
2. Use the supplied text only to identify structures, conditions, channels, or
   experimental context that cannot be established from pixels alone.
3. Keep the subimage and compound-figure context distinct.
4. Do not invent measurements, absent findings, diagnoses, or causal claims.
5. Use formal medical language and no more than five sentences.

OUTPUT
Visual description: <description>

SYNTHETIC FORMATTING EXAMPLE
Input subimage: <axial chest CT subimage>
Input subimage caption: Axial CT view of a left upper-lobe pulmonary nodule.
Input compound-figure caption: Axial and coronal CT views of the same pulmonary
nodule.
Input inline summary: The nodule is well circumscribed; follow-up imaging was
recommended to assess interval stability.
Input medical knowledge: Nodule morphology and interval growth contribute to
pulmonary-nodule assessment.
Output visual description: The axial CT subimage contains a discrete,
well-circumscribed pulmonary nodule with a smooth margin. The caption identifies
its location as the left upper lobe.
\end{lstlisting}
\end{minipage}
\caption{\textbf{Step 3 prompt specification: subimage-level medical visual description.} The synthetic example distinguishes visible morphology from anatomical information supplied by the caption.}
\label{fig:prompt-step3}
\end{figure*}

\begin{figure*}[!htbp]
\centering
\begin{minipage}{0.96\linewidth}
\begin{lstlisting}[style=suppprompt]
ROLE
You are a biomedical assistant constructing a multi-subimage VQA instruction.

CONSTRUCTION INPUTS
Compound-figure caption: <compound_figure_caption>
Inline summary: <inline_summary>
Medical knowledge: <medical_knowledge>
Target visual-input type: compound figure plus constituent subimages
Selected subimages: <subimage_ids>
For each selected subimage:
  identifier: <Fig-X>
  caption: <subimage_caption>
  visual description: <visual_description>

TASK
Construct a context-question-answer instruction that requires integration of
evidence from all selected subimages.

REQUIREMENTS
1. Construct the answer from the evidence associated with the selected
   subimages.
2. Keep each finding associated with the correct subimage identifier.
3. Make the question answerable only by integrating the selected evidence.
4. Keep the context useful but avoid directly revealing the answer.
5. Do not introduce unsupported findings, diagnoses, or causal claims.

OUTPUT
Context: <auxiliary context>
Question: <multi-subimage question>
Answer: <reference answer>

SYNTHETIC FORMATTING EXAMPLE
Target visual-input type: compound figure plus constituent subimages.
Compound-figure caption: Axial and coronal CT views of the same left upper-lobe
pulmonary nodule.
Inline summary: The nodule is well circumscribed; follow-up imaging was
recommended to assess interval stability.
Medical knowledge: Multiplanar CT views provide complementary information about
lesion morphology and anatomical location.
Selected subimages: Fig-1 and Fig-2.
Fig-1 caption: Axial CT view of a pulmonary nodule.
Fig-1 visual description: A discrete pulmonary nodule has a smooth,
well-circumscribed margin.
Fig-2 caption: Coronal CT view of the same left upper-lobe pulmonary nodule.
Fig-2 visual description: The nodule is shown in the left upper lobe on the
coronal view.
Output context: Orthogonal CT planes provide complementary information about
lesion morphology and anatomical location.
Output question: How do Fig-1 and Fig-2 jointly characterize the pulmonary
nodule?
Output answer: Fig-1 shows a well-circumscribed nodule on the axial view, while
Fig-2 confirms its left upper-lobe location on the coronal view. Together, the
two views establish the nodule's morphology and location.
\end{lstlisting}
\end{minipage}
\caption{\textbf{Step 4 prompt specification for multi-subimage visual question answering (VQA) instruction construction.} The synthetic example integrates complementary evidence from two selected subimages.}
\label{fig:prompt-step4-multi-subimage}
\end{figure*}

\begin{figure*}[!htbp]
\centering
\begin{minipage}{0.96\linewidth}
\begin{lstlisting}[style=suppprompt]
ROLE
You are a biomedical assistant constructing a single-subimage VQA instruction.

CONSTRUCTION INPUTS
Compound-figure caption: <compound_figure_caption>
Inline summary: <inline_summary>
Medical knowledge: <medical_knowledge>
Target visual-input type: compound figure plus constituent subimages
Selected subimage identifier: <Fig-X>
Selected subimage caption: <subimage_caption>
Selected subimage visual description: <visual_description>

TASK
Construct a context-question-answer instruction for which one specified
subimage provides the visual evidence.

REQUIREMENTS
1. Base the visual content of the question and answer on the selected subimage.
2. Refer to the selected subimage by its identifier.
3. Do not rely on unselected subimages or the compound figure as a whole.
4. Keep the context useful but avoid directly revealing the answer.
5. Do not add unsupported measurements, absent findings, or diagnoses.

OUTPUT
Context: <auxiliary context>
Question: <single-subimage question>
Answer: <reference answer>

SYNTHETIC FORMATTING EXAMPLE
Target visual-input type: compound figure plus constituent subimages.
Compound-figure caption: Axial and coronal CT views of the same left upper-lobe
pulmonary nodule.
Inline summary: The nodule is well circumscribed; follow-up imaging was
recommended to assess interval stability.
Medical knowledge: Pulmonary-nodule assessment incorporates morphology,
attenuation, interval growth, and patient-level risk factors.
Selected subimage identifier: Fig-1.
Selected subimage caption: Axial CT view of a left upper-lobe pulmonary nodule.
Selected subimage visual description: A discrete pulmonary nodule has a smooth,
well-circumscribed margin.
Output context: CT morphology contributes to pulmonary-nodule assessment.
Output question: What margin characteristic is visible for the pulmonary
nodule in Fig-1?
Output answer: The nodule in Fig-1 has a smooth, well-circumscribed margin.
\end{lstlisting}
\end{minipage}
\caption{\textbf{Step 4 prompt specification for single-subimage visual question answering (VQA) instruction construction.} The synthetic example associates the question and answer with one selected subimage.}
\label{fig:prompt-step4-single-subimage}
\end{figure*}

\begin{figure*}[!htbp]
\centering
\begin{minipage}{0.96\linewidth}
\begin{lstlisting}[style=suppprompt]
ROLE
You are a biomedical assistant constructing a relative-position VQA
instruction from the layout of two subimages in a compound figure.

CONSTRUCTION INPUTS
First subimage: <Fig-X>
First bounding-box centre: <x1, y1>
Second subimage: <Fig-Y>
Second bounding-box centre: <x2, y2>
Precomputed relative position: <relative_position>
Coordinate convention: image origin at the upper left; x increases to the
right and y increases downward

TASK
Rephrase the precomputed layout relationship as a concise question-answer pair.

REQUIREMENTS
1. Ask only about literal physical placement within the compound figure.
2. Preserve the supplied subimage identifiers and relative-position label.
3. Do not interpret the biomedical content of either subimage.
4. Return the requested spatial relation directly, without explanatory reasoning.

OUTPUT
Question: <relative-position question>
Answer: <relative-position label>

SYNTHETIC FORMATTING EXAMPLE
Input: Fig-1 centre = (100, 150); Fig-2 centre = (250, 100);
precomputed relative position = left and below.
Output question: What is the position of Fig-1 relative to Fig-2 within the
compound figure?
Output answer: Fig-1 is positioned to the left of and below Fig-2.
\end{lstlisting}
\end{minipage}
\caption{\textbf{Step 4 prompt specification for relative-position visual question answering (VQA) instruction construction.} The synthetic example expresses a precomputed relationship between bounding-box centres using the image-coordinate convention stated in the prompt.}
\label{fig:prompt-step4-relative-position}
\end{figure*}

\begin{figure*}[!htbp]
\centering
\begin{minipage}{0.96\linewidth}
\begin{lstlisting}[style=suppprompt]
ROLE
You are a biomedical assistant constructing a compound-figure VQA instruction.

CONSTRUCTION INPUTS
Target visual-input type: compound figure (Fig-0)
Compound-figure caption: <compound_figure_caption>
Inline summary: <inline_summary>
Medical knowledge: <medical_knowledge>
Constituent-subimage descriptions: <visual_descriptions>

TASK
Construct a context-question-answer instruction for which the downstream model
receives the complete compound figure as one visual input.

REQUIREMENTS
1. Ask about evidence represented by the compound figure as a whole.
2. Base the answer on the supplied caption, summary, and visual descriptions,
   and ensure that it can be supported by the target compound figure.
3. Use subimage-specific details only when they are needed to interpret the
   complete figure.
4. Keep the context useful but avoid directly revealing the answer.
5. Do not introduce unsupported findings or conclusions.

OUTPUT
Context: <auxiliary context>
Question: <compound-figure question>
Answer: <reference answer>

SYNTHETIC FORMATTING EXAMPLE
Target visual-input type: compound figure (Fig-0).
Input caption: Axial and coronal CT views of the same left upper-lobe pulmonary
nodule.
Input inline summary: The nodule is well circumscribed; follow-up imaging was
recommended to assess interval stability.
Input medical knowledge: Multiplanar CT can characterize lesion morphology and
anatomical location across complementary planes.
Input visual descriptions: The axial view shows a smooth, well-circumscribed
nodule; the coronal view shows its left upper-lobe location.
Output context: Multiplanar CT depicts a lesion from complementary anatomical
perspectives.
Output question: What information does the complete compound figure provide
about the pulmonary nodule?
Output answer: The compound figure shows the nodule in axial and coronal planes,
jointly depicting its morphology and left upper-lobe location.
\end{lstlisting}
\end{minipage}
\caption{\textbf{Step 4 prompt specification for compound-figure visual question answering (VQA) instruction construction.} The construction evidence is represented as structured text, and the resulting instruction assigns Fig-0 as the downstream visual input.}
\label{fig:prompt-step4-single-image}
\end{figure*}

\begin{figure*}[!htbp]
\centering
\begin{minipage}{0.96\linewidth}
\begin{lstlisting}[style=suppprompt]
ROLE
You are a biomedical assistant constructing a text-only QA instruction. The
downstream model will not receive an image.

INPUT
Compound-figure caption: <compound_figure_caption>
Inline summary: <inline_summary>
Medical knowledge: <medical_knowledge>

TASK
Construct a self-contained context-question-answer instruction from the supplied
textual information.

REQUIREMENTS
1. The question and answer must be understandable without visual access.
2. Do not refer to an image, figure, panel, subimage, or visual location.
3. Use the caption and inline summary to identify the topic; use the medical-
   knowledge note to formulate a related, self-contained question.
4. Keep the context useful but avoid directly revealing the answer.
5. Preserve uncertainty and avoid unsupported patient-specific conclusions.

OUTPUT
Context: <text-only auxiliary context>
Question: <self-contained medical question>
Answer: <reference answer>

SYNTHETIC FORMATTING EXAMPLE
Input caption: Axial and coronal chest CT views of a solitary pulmonary nodule.
Input inline summary: Follow-up imaging was recommended to assess interval
stability.
Input medical knowledge: Pulmonary-nodule assessment integrates size,
attenuation, morphology, interval growth, and patient-level risk factors.
Output context: Pulmonary nodules are assessed using imaging characteristics,
interval change, and patient-level risk factors.
Output question: Which factors are commonly considered when estimating the
risk of malignancy in a pulmonary nodule?
Output answer: Assessment commonly considers nodule size, attenuation,
morphology, interval growth, and patient-level factors such as age, smoking
history, and prior malignancy.
\end{lstlisting}
\end{minipage}
\caption{\textbf{Step 4 prompt specification for text-only question answering (QA) instruction construction.} The synthetic example yields a self-contained question and answer that do not require visual access.}
\label{fig:prompt-step4-text-only}
\end{figure*}

\begin{figure*}[!htbp]
\centering
\begin{minipage}{0.96\linewidth}
\begin{lstlisting}[style=suppprompt]
ROLE
You are a biomedical assistant constructing a four-option multiple-choice VQA
instruction.

CONSTRUCTION INPUTS
Target visual-input type: <compound figure or multi-image input>
Target visual identifiers: <Fig-0 or selected subimage identifiers>
Compound-figure caption: <compound_figure_caption>
Inline summary: <inline_summary>
Relevant visual descriptions: <visual_descriptions>
Medical knowledge: <medical_knowledge>
Required correct-option position: <A, B, C, or D>

TASK
Construct one clinically or scientifically relevant question with one supported
answer and three plausible distractors.

REQUIREMENTS
1. Make the question answerable from the supplied visual evidence and context.
2. Provide exactly four mutually exclusive options labelled A-D.
3. Include one unambiguously correct option and three plausible but incorrect
   distractors drawn from the same conceptual domain.
4. Avoid answer cues based on wording, length, or option position.
5. Place the correct option at the required position supplied by the dataset-
   construction pipeline.

OUTPUT
Question: <question>
Options: A. <option>; B. <option>; C. <option>; D. <option>
Correct answer: <A, B, C, or D>

SYNTHETIC FORMATTING EXAMPLE
Target visual-input type: multi-image input.
Target visual identifiers: Fig-1.
Compound-figure caption: Axial and coronal CT views of a solitary pulmonary
nodule.
Inline summary: The nodule is well circumscribed; follow-up imaging was
recommended to assess interval stability.
Relevant visual description: Fig-1 shows a pulmonary nodule with a smooth,
well-circumscribed margin.
Medical knowledge: Margin morphology is one feature considered in
pulmonary-nodule assessment.
Required correct-option position: B.
Question: Which visible feature in Fig-1 describes the pulmonary nodule margin?
Options: A. Spiculated; B. Smooth and well circumscribed; C. Lobulated;
D. Indistinct
Correct answer: B
\end{lstlisting}
\end{minipage}
\caption{\textbf{Step 4 prompt specification for multiple-choice visual question answering (VQA) instruction construction.} The synthetic example uses four alternatives from the same conceptual domain and a correct-option position assigned by the construction pipeline.}
\label{fig:prompt-step4-multiple-choice}
\end{figure*}

\begin{figure*}[!htbp]
\centering
\begin{minipage}{0.96\linewidth}
\begin{lstlisting}[style=suppprompt]
ROLE
You are a biomedical assistant refining auxiliary context for an instruction.

INPUT
Compound-figure caption: <compound_figure_caption>
Inline summary: <inline_summary>
Medical knowledge: <medical_knowledge>
Visual descriptions: <visual_descriptions>
Question: <question>
Reference answer: <reference_answer>
Draft context: <draft_context>

TASK
Rewrite the draft context so that it remains relevant to the question but does
not reveal the reference answer.

REQUIREMENTS
1. Remove phrases that state, paraphrase, or uniquely imply the answer.
2. Retain only background needed to understand the task.
3. Preserve medically relevant uncertainty and source-supported terminology.
4. Do not add new findings, diagnoses, or causal claims.
5. Return only the refined context.

OUTPUT
Refined context: <context>

SYNTHETIC FORMATTING EXAMPLE
Compound-figure caption: Axial and coronal CT views of a solitary pulmonary
nodule.
Inline summary: The nodule is well circumscribed; follow-up imaging was
recommended to assess interval stability.
Medical knowledge: Nodule morphology, interval growth, and patient-level risk
factors contribute to assessment.
Visual description: Fig-1 shows a pulmonary nodule with a smooth,
well-circumscribed margin.
Question: What margin characteristic is visible for the nodule in Fig-1?
Reference answer: The nodule has a smooth, well-circumscribed margin.
Draft context: Fig-1 shows a smooth, well-circumscribed left upper-lobe nodule.
Output refined context: CT margin morphology is one component of pulmonary-
nodule assessment and should be interpreted with size, attenuation, interval
change, and patient-level risk factors.
\end{lstlisting}
\end{minipage}
\caption{\textbf{Step 5 prompt specification: context refinement to reduce answer leakage.} The synthetic example retains relevant background while removing the answer-bearing description from the auxiliary context.}
\label{fig:prompt-step5}
\end{figure*}

\FloatBarrier
\subsubsection{Illustrative pipeline transformations}
\noindent Source-derived examples are shown below to illustrate the five construction steps, including transformation of article-linked text into compact context, medical background, subimage-level visual descriptions, task-specific instructions, and leakage-reduced auxiliary context.

\begin{figure*}[!htbp]
\centering
   \includegraphics [width=0.95\textwidth]{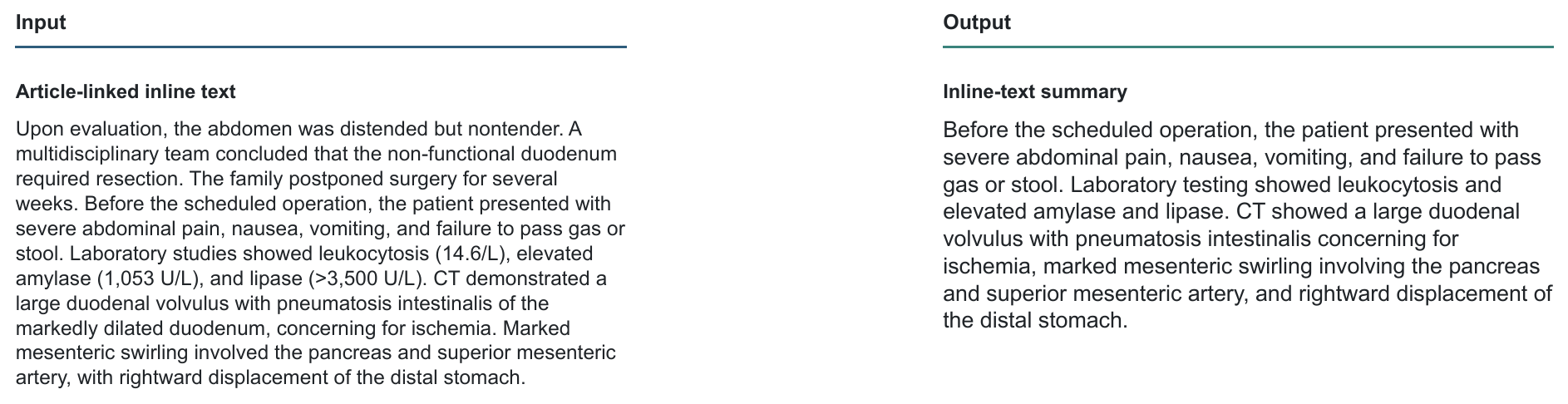}
\caption{\textbf{Illustrative Step 1 transformation.} Source-associated inline text and the resulting summary, with the reported level of certainty retained.}
\label{fig:stage1}
\end{figure*}

\begin{figure*}[!htbp]
\centering
   \includegraphics [width=0.95\textwidth]{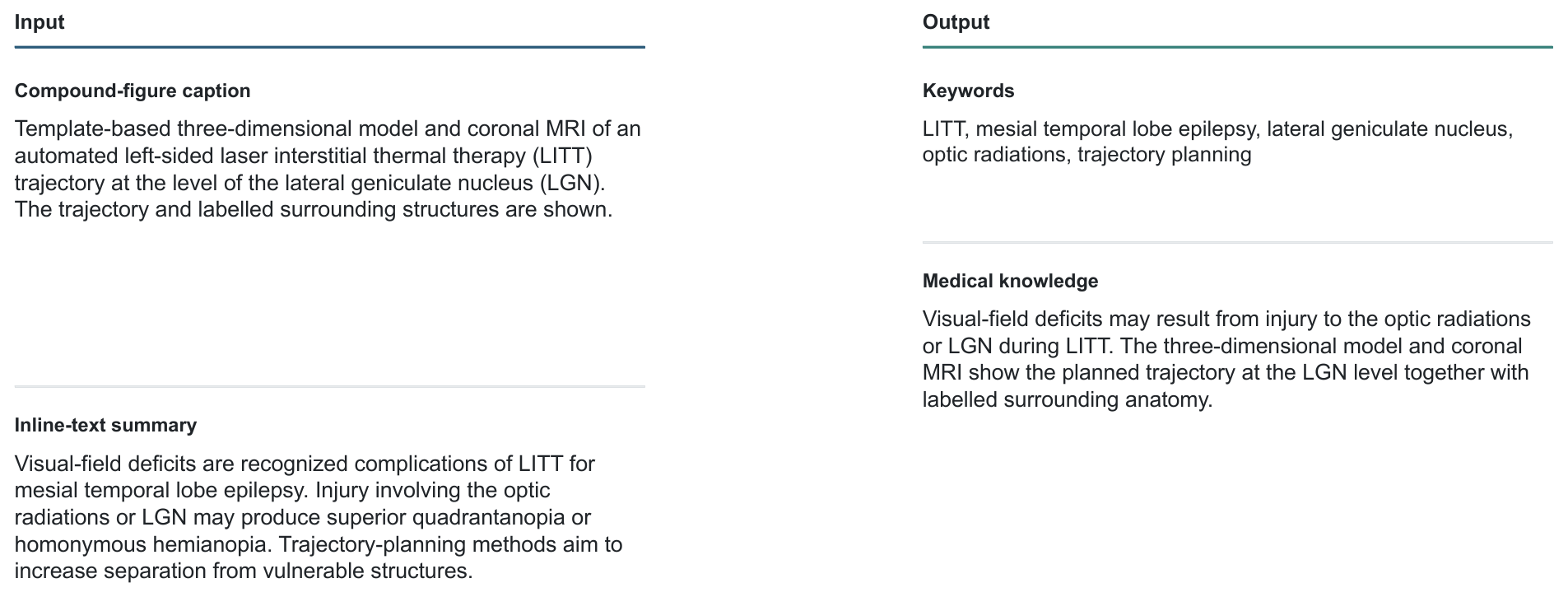}
\caption{\textbf{Illustrative Step 2 transformation.} The compound-figure caption and inline-text summary are provided as inputs; the output contains the extracted keywords and complementary medical knowledge relevant to LITT trajectory planning.}
\label{fig:stage2}
\end{figure*}

\begin{figure*}[!htbp]
\centering
   \includegraphics [width=0.95\textwidth]{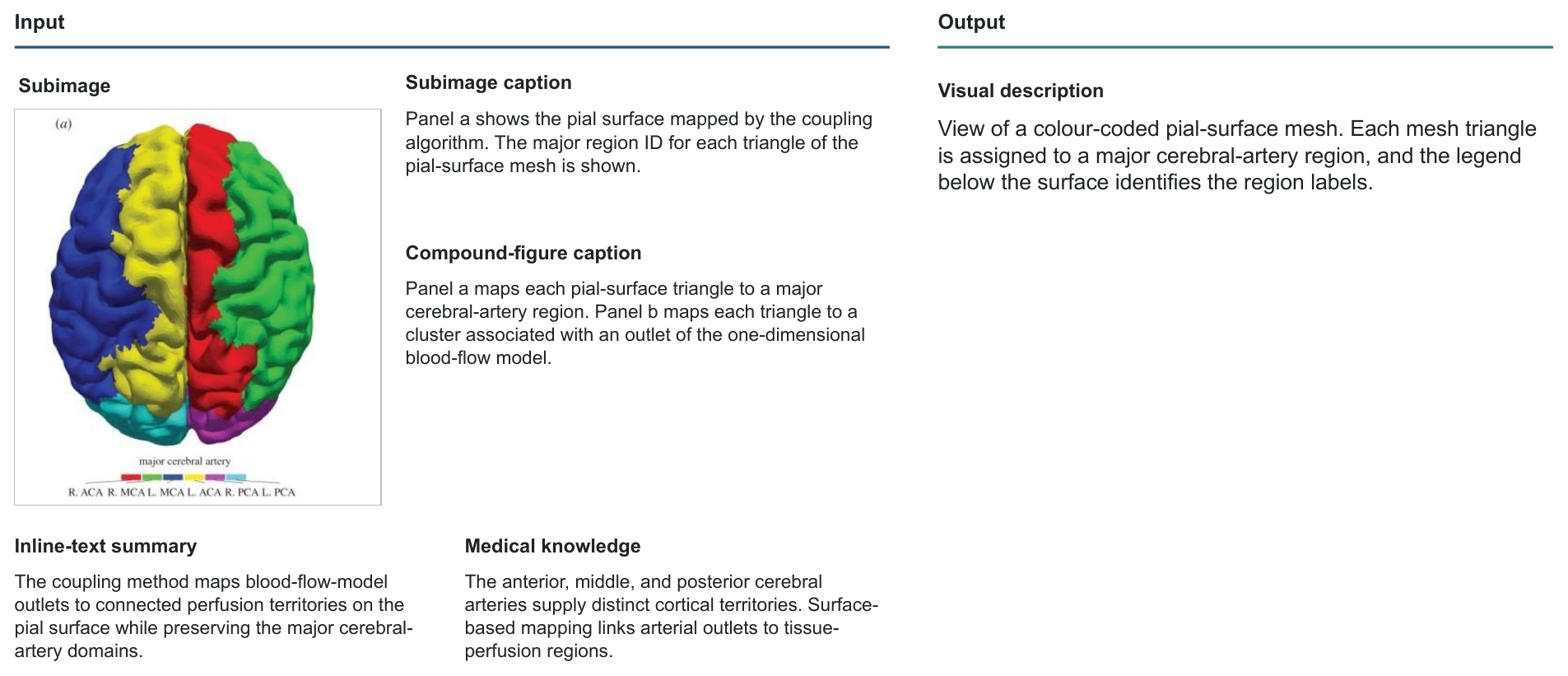}
\caption{\textbf{Illustrative Step 3 transformation.} The subimage, subimage caption, compound-figure caption, inline-text summary, and medical-knowledge note are provided as inputs. The output visual description is limited to visible and caption-supported content.}
\label{fig:stage3}
\end{figure*}

\begin{figure*}[!htbp]
\centering
   \includegraphics [width=1\textwidth]{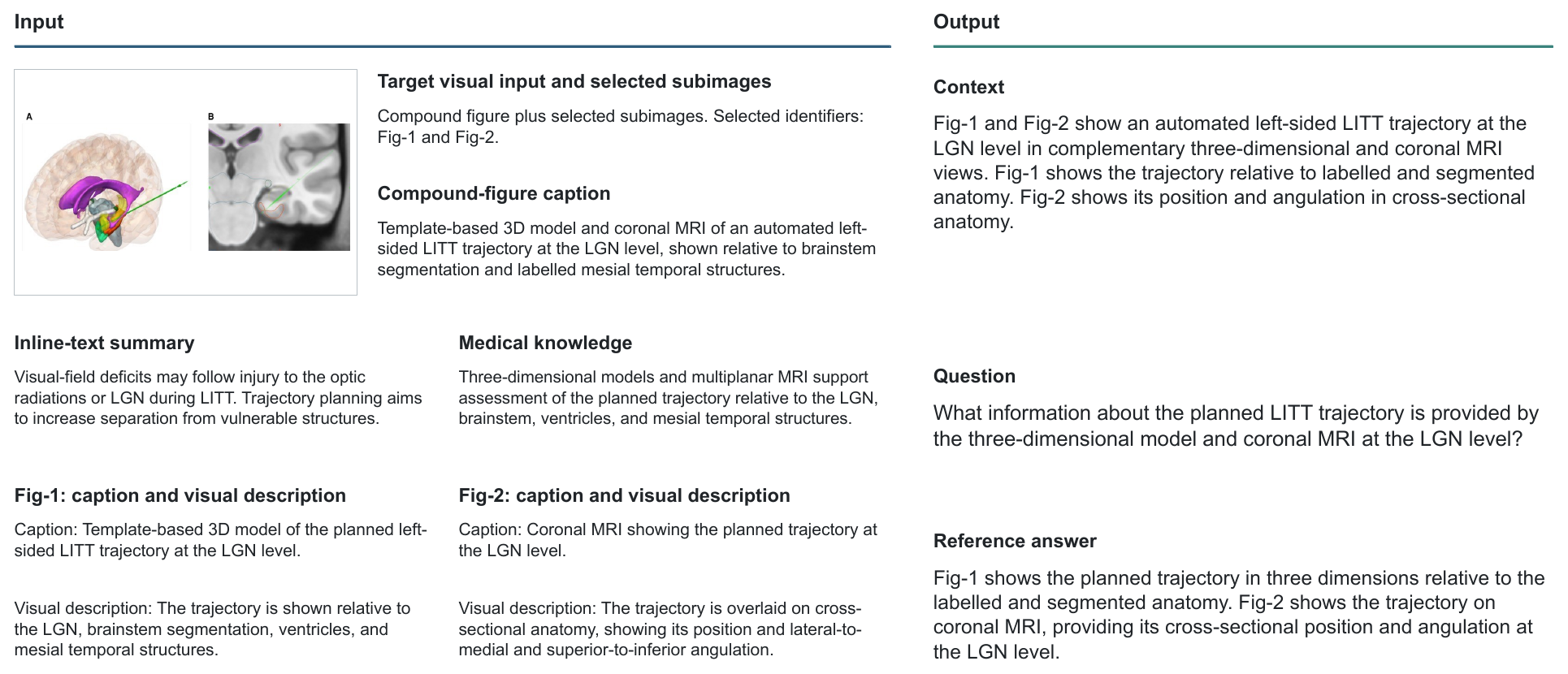}
\caption{\textbf{Illustrative Step 4 transformation.} The target visual-input type, selected subimages, compound-figure caption, inline-text summary, medical knowledge, subimage captions, and subimage-level visual descriptions are assembled into a context--question--answer instruction.}
\label{fig:stage4}
\end{figure*}

\begin{figure*}[!htbp]
\centering
   \includegraphics [width=0.95\textwidth]{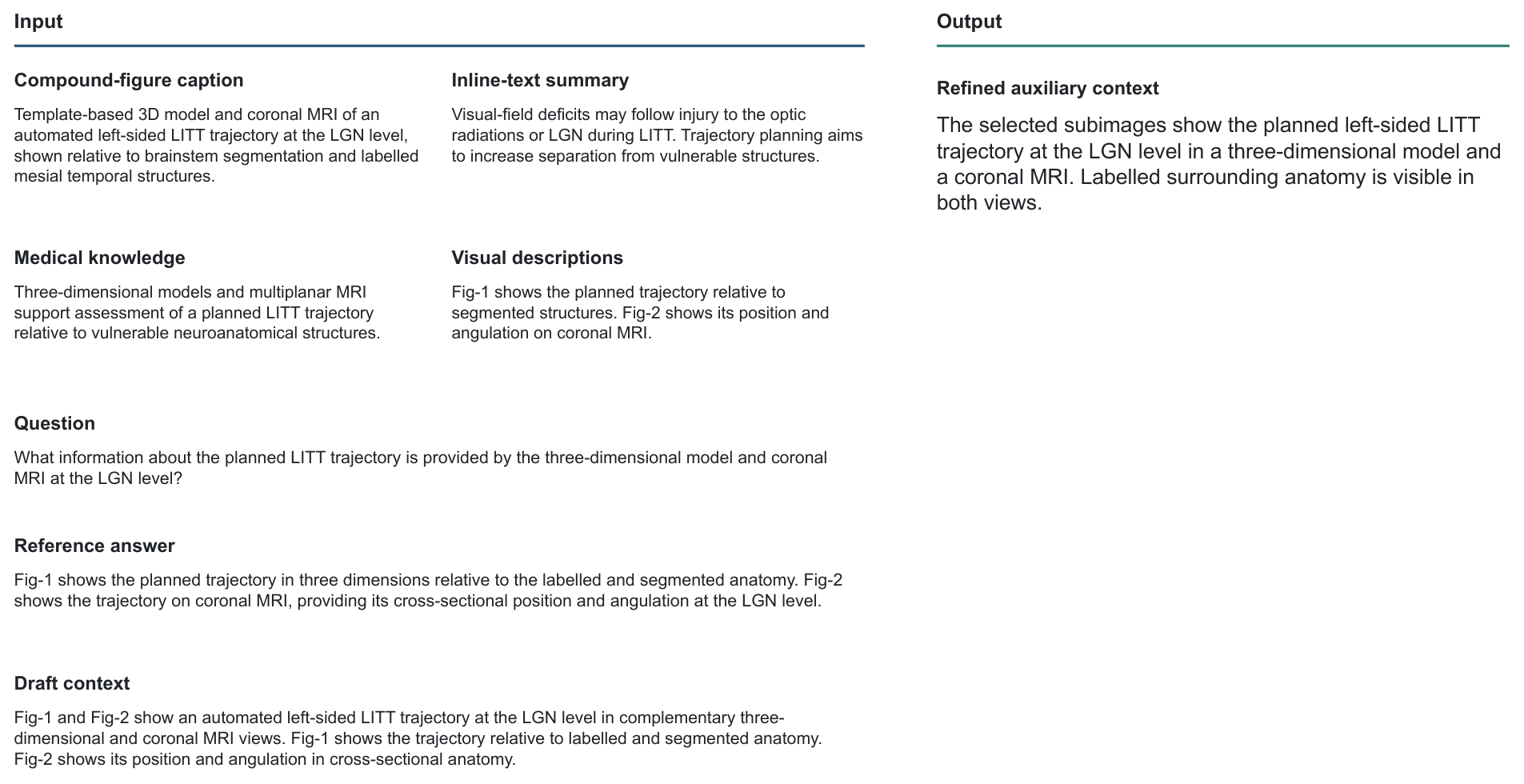}
\caption{\textbf{Illustrative Step 5 transformation.} The complete refinement input includes the compound-figure caption, inline-text summary, medical knowledge, visual descriptions, question, reference answer, and draft context. The refined auxiliary context removes answer-bearing content while retaining relevant background.}
\label{fig:stage5}
\end{figure*}

\FloatBarrier

\subsection{\texorpdfstring{M$^3$LLM}{M3LLM} training implementation}
\label{sec:supp_training_implementation}
\noindent This section reports the data allocation, input formatting, software environment, optimization settings, and checkpoint policy for the three-stage training framework. Stages~I and III used the same full-parameter supervised fine-tuning (SFT) configuration, whereas Stage~II used selection-aware Group Relative Policy Optimization (GRPO) on a dedicated policy-refinement partition.

\paragraph{Policy-refinement data and structured reference trajectories.}
The policy-refinement component contained 10,555 multi-subimage VQA instances derived from distinct compound figures. A random subset of 200 instances was reserved for Stage~II validation, leaving 10,355 instances for optimization. Each record contained the question, optional auxiliary context, candidate images, the Step~4 instruction answer, and instruction-derived reference subimage identifiers.

\noindent GPT-5.5 constructed one structured reference trajectory for each instance from the question, candidate images, optional auxiliary context, instruction-derived reference subimage identifiers, and Step~4 reference answer. The \textit{Thought} field followed an ordered protocol comprising image triage, subimage-specific visual-feature extraction, and medical synthesis. The \textit{Selection} field preserved the instruction-derived reference identifiers, and the \textit{Answer} field contained a GPT-5.5-refined reference answer that integrated the Step~4 answer with the supplied multimodal evidence and auxiliary context. During Stage~II, these two fields supplied the targets for $R_{\mathrm{sel}}$ and $R_{\mathrm{ans}}$, respectively. The complete structured reference trajectory served as the supervised target for the matched continued-SFT control.

\noindent Structured reference trajectories were constructed on 10 May 2026 through the Azure OpenAI application programming interface (API) using the \texttt{gpt5.5} deployment and API version \path{2024-12-01-preview}. No decoding parameters were explicitly specified, and generation followed the API defaults. The GPT-5.5 trajectory-construction and Stage~II rollout prompts are reproduced as searchable text in Supplementary Figs.~\ref{fig:prompt-structured-reference-trajectory}
and~\ref{fig:prompt-stage2-grpo}.

\begin{figure*}[!htbp]
    \centering
    \begin{minipage}{0.96\linewidth}
\begin{lstlisting}[style=suppprompt]
ROLE
You are a biomedical multimodal reasoning assistant. Construct one evidence-
grounded structured reference trajectory for a medical multi-image question.

INPUT
Question: <question>
Candidate images:
  Fig-0: <compound_figure>
  Fig-1 ... Fig-N: <subimages>
Auxiliary context: <context, if available>
Reference subimage identifiers: <reference_subimage_ids>
Step 4 reference answer: <step4_reference_answer>

TASK
Use the reference subimage identifiers as an evidence constraint. Inspect the
supplied images and context, construct the ordered reasoning trajectory, and
refine the Step 4 reference answer into a concise final answer.

REASONING SEQUENCE
1. Image triage: identify the question-relevant images and explain why the
   remaining images are not required.
2. Visual-feature extraction: describe the relevant visible findings separately
   for each selected image and retain its identifier.
3. Medical synthesis: integrate the selected visual evidence and auxiliary
   context into a coherent conclusion that supports the final answer.

ANSWER REFINEMENT
Preserve the supported content of the Step 4 reference answer while improving
its clarity and integration with the supplied evidence. Resolve ambiguity only
when the images or auxiliary context provide support. Do not introduce new
findings, diagnoses, or causal claims.

OUTPUT
<Thought>
Image triage: ...
Visual-feature extraction:
Fig-X: ...
Fig-Y: ...
Medical synthesis: ...
</Thought>
<Selection>Fig-X, Fig-Y</Selection>
<Answer>...</Answer>
\end{lstlisting}
    \end{minipage}
    \caption{\textbf{GPT-5.5 prompt for constructing structured reference trajectories.} GPT-5.5 received the Step~4 reference answer together with the multimodal evidence and reference subimage identifiers, and returned a structured trajectory containing a refined reference answer.}
    \label{fig:prompt-structured-reference-trajectory}
\end{figure*}

\begin{figure*}[!htbp]
    \centering
    \begin{minipage}{0.96\linewidth}
\begin{lstlisting}[style=suppprompt]
ROLE
You are a biomedical multimodal assistant. Select the images relevant to the
question, interpret their visual evidence, and answer the question.

INPUT
Question: <question>
Candidate images:
  Fig-0: <compound_figure>
  Fig-1 ... Fig-N: <subimages>
Auxiliary context: <context, if available>

REASONING SEQUENCE
1. Image triage: inspect all candidate images, identify those required to answer
   the question, and explain why the remaining images are not selected.
2. Visual-feature extraction: describe the relevant findings separately for
   each selected image and cite its identifier.
3. Medical synthesis: integrate the selected evidence into a coherent medical
   interpretation before producing the final answer.

CONSTRAINTS
Use only valid candidate-image identifiers. Keep each image-specific claim
associated with the correct identifier. Return each field once, in the order
shown below, with no text outside the three fields.

OUTPUT
<Thought>
Image triage: ...
Visual-feature extraction:
Fig-X: ...
Medical synthesis: ...
</Thought>
<Selection>Fig-X, Fig-Y</Selection>
<Answer>...</Answer>
\end{lstlisting}
    \end{minipage}
    \caption{\textbf{Stage II rollout prompt for selection-aware Group Relative Policy Optimization (GRPO).}
    The policy received the question, candidate images, and optional auxiliary context. Reference subimage identifiers and the GPT-5.5-refined reference answer were withheld during rollout and used only to calculate the selection and answer rewards after generation.}
    \label{fig:prompt-stage2-grpo}
\end{figure*}

\paragraph{Shared SFT software and input processing.}
Stages~I and III were implemented in LLaMA-Factory (version \texttt{0.9.5.dev0}; git revision \texttt{v0.9.4-48-}\allowbreak\texttt{g184304b5}). The software environment used DeepSpeed ZeRO-3~\texttt{0.18.4}, PyTorch~\texttt{2.10.0}, and Transformers~\texttt{5.0.0}. Unspecified training arguments inherited the Hugging Face defaults: fused AdamW with $\beta_1=0.9$, $\beta_2=0.999$, and $\epsilon=10^{-8}$; zero weight decay; and global $\ell_2$ gradient clipping at 1.0. Sequence packing was disabled. Images were independently resized with preserved aspect ratio to a pixel area in $[32^2,512^2]$ before Qwen3-VL preprocessing. Each instance contained at most four images: Fig-0 was the compound figure, and Fig-1 through Fig-$N$ were up to three constituent subimages. The combined text and visual-token context was limited to 2,048 tokens; sequences exceeding this budget were truncated by the LLaMA-Factory source--target allocation procedure rather than discarded.
\paragraph{Stage I implementation.}
Stage~I was initialized from the pretrained Qwen3-VL-8B checkpoint. The 10,555 instances used for policy refinement were excluded from the multi-image instruction-tuning data in this stage, leaving 105,252 single-subimage, multi-subimage, and relative-position VQA instances for supervised instruction tuning. Full-parameter training was performed using LLaMA-Factory with DeepSpeed ZeRO-3, with the visual and language components jointly updated. Stage~I was trained for one epoch on two NVIDIA H200 GPUs using fused AdamW, a learning rate of $2\times10^{-6}$, no weight decay, cosine learning-rate decay, and a warm-up ratio of 0.03. We used bfloat16 precision, a per-device batch size of 1, and eight gradient-accumulation steps, yielding an effective global batch size of 16. The training seed was 42.

\noindent Each image was independently resized with its aspect ratio preserved to a maximum pixel area of 262,144 pixels, equivalent to $512\times512$ pixels for a square image. No aggregate pixel budget was imposed across the images in an instance. The combined multimodal prompt and response, including visual tokens, was limited to 2,048 tokens, and sequence packing was disabled. Instances exceeding this limit were retained rather than discarded: the expanded sequence was truncated to 2,048 tokens, preferentially preserving the prompt and removing overflow from the end of the response. Sequence truncation did not reduce the image list, and all images in the instance were supplied to the vision encoder. Stage~I used no validation partition or validation-based checkpoint selection; the checkpoint from the end of training was used to initialize Stage~II.

\paragraph{Stage II implementation.}
Stage~II was initialized from the final Stage~I checkpoint. It used 10,355 policy-optimization instances and the randomly sampled 200-instance validation set described above. The complete multimodal model was optimized; LoRA was not used, the vision encoder was not frozen, and gradient checkpointing was enabled. Policy refinement was implemented with EasyR1 (git \texttt{v0.3.2-41-gdd71bbd}) and \texttt{verl} 0.3.3.dev0 in the \path{hiyouga/verl:ngc-th2.8.0-cu12.9-vllm0.11.0} container, using PyTorch 2.8 and vLLM 0.11.0. Transformers was constrained to versions $\geq4.54$ and $<5.0$. Training ran for 15 epochs on four NVIDIA H200 GPUs.

\noindent The actor was optimized using AdamW with a learning rate of $1\times10^{-6}$, weight decay of 0.01, no learning rate warm-up, and gradient clipping at 1.0. Four prompts were processed per global batch, and 16 on-policy responses were sampled per prompt. The policy-update microbatch size was 1, and the experience microbatch size was 2. Rollouts used a temperature of 1.0, top-$p$ of 1.0, tensor parallelism of 4, a maximum of 15,000 batched tokens, and a vLLM GPU-memory utilization target of 0.45. The maximum prompt and response lengths were 10,000 and 2,048 tokens, respectively; prompts exceeding the maximum were filtered before rollout generation. Stage~II images were processed using minimum and maximum per-image pixel areas of 262,144 and 4,194,304 pixels, respectively. The Stage~II random seed was 1.

\noindent For prompt $x_i$, the policy generated a group of $G=16$ responses
$\{y_{i,j}\}_{j=1}^{G}$. If $r_{i,j}$ denotes the composite reward, the group-relative advantage was
\begin{equation}
\hat{A}_{i,j}=\frac{r_{i,j}-\bar r_i}{s_i+10^{-6}},
\end{equation}
where $\bar r_i$ and $s_i$ are the group mean and the unbiased group standard deviation, respectively. The GRPO surrogate used one policy-update epoch, asymmetric ratio clipping with lower and upper coefficients 0.2 and 0.3, and dual clipping with coefficient 3. A frozen copy of the final Stage~I model served as the reference policy. A low-variance token-level KL penalty with coefficient $\beta_{\mathrm{KL}}=0.01$ constrained deviation from this reference.

\paragraph{Stage II reward-operator selection.}
\label{sec:supp_reward_operator_selection}
Five reward aggregation operators were specified before evaluation:
$(\lambda_{\mathrm{ans}},\lambda_{\mathrm{sel}},
\lambda_{\mathrm{fmt}})$
$=(0.8,0.1,0.1)$, $(0.7,0.2,0.1)$, $(0.6,0.3,0.1)$,
$(0.5,0.4,0.1)$ and $(0.6,0.2,0.2)$.
Each condition used the same final Stage~I checkpoint, policy-refinement training and validation data, rollout-group size, and optimization schedule. Mean answer-quality reward on the held-out Stage~II validation set was the prespecified primary selection criterion; subimage-selection F1 and valid-format rate were examined as secondary diagnostics. PMC-MI-Bench was not used for operator selection. The selected operator was
$(\lambda_{\mathrm{ans}},\lambda_{\mathrm{sel}},
\lambda_{\mathrm{fmt}})=(0.6,0.3,0.1)$.

\paragraph{Stage III implementation.}
Stage~III was initialized from the final Stage~II checkpoint and trained on the 224,401 supervised PMC-MI instructions remaining after exclusion of the 10,555 policy-refinement instances. Its software, full-parameter optimization, optimizer, learning-rate schedule, epoch count, precision, batch size, gradient accumulation, image processing, sequence truncation, hardware, and random seed were identical to those used in Stage~I. Stage~III had no validation set and no validation-based checkpoint selection. The checkpoint at the end of the prespecified one-epoch schedule was designated M$^3$LLM.

\FloatBarrier

\subsection{Manual quality assessment of constructed data}
\label{sec:supp_data_quality}
\noindent Two complementary manual audits were performed: one assessed outputs from the five-step instruction-construction pipeline, and the other assessed the structured reference trajectories used for policy refinement. The sampling and assignment procedures are detailed below; the resulting quality estimates are reported in Supplementary Tables~\ref{tab:training_sample_assessment} and~\ref{tab:policy_refinement_data_quality}.

\paragraph{Manual quality assessment of PMC-MI instruction data.}
A total of 140 construction outputs were randomly sampled. Thirty outputs were drawn from each of Steps~1--4, comprising five examples from each of the six task formats; Step~5 contributed 20 outputs because multiple-choice and relative-position VQA did not contain an auxiliary context requiring leakage refinement. Three medical reviewers (S.A., G.M., and M.G.) participated. Every output received two independent ratings for correctness, completeness, and clarity on a five-point scale, and 78 outputs received a third independent rating for the reliability analysis. Reviewers assessed the generated output against the source material available at the corresponding construction step.

\begin{table}[H]
\centering
\caption{\textbf{Manual assessment of PMC-MI instruction quality across the five construction steps.} Values are mean $\pm$ standard deviation ratings on a five-point scale. Each of the 140 outputs received two independent ratings, and 78 outputs received a third rating for the reliability analysis.}
\label{tab:training_sample_assessment}
\small
\setlength{\tabcolsep}{12pt}
\renewcommand{\arraystretch}{1.12}
\begin{tabular}{@{}lcccc@{}}
\toprule
\textbf{Step} & \textbf{Correctness} & \textbf{Completeness} & \textbf{Clarity} & \textbf{Average} \\
\midrule
Step 1 & $4.8 \pm 0.2$ & $4.4 \pm 0.4$ & $4.9 \pm 0.1$ & $4.7 \pm 0.2$ \\
Step 2 & $4.9 \pm 0.1$ & $4.8 \pm 0.3$ & $5.0 \pm 0.0$ & $4.9 \pm 0.1$ \\
Step 3 & $4.3 \pm 0.1$ & $4.5 \pm 0.3$ & $4.5 \pm 0.0$ & $4.4 \pm 0.1$ \\
Step 4 & $4.6 \pm 0.1$ & $4.7 \pm 0.2$ & $4.8 \pm 0.2$ & $4.7 \pm 0.1$ \\
Step 5 & $4.8 \pm 0.1$ & $4.9 \pm 0.1$ & $5.0 \pm 0.1$ & $4.9 \pm 0.1$ \\
\bottomrule
\end{tabular}
\end{table}

\noindent Among the 78 outputs with three independent ratings, the intraclass correlation coefficients were 0.867 for correctness, 0.751 for completeness, and 0.720 for clarity. Reviewers assigned identical scores in 74.8\% of assessments, and 98.3\% of ratings differed by no more than one scale point.

\paragraph{Manual quality assessment of policy-refinement data.}
\label{sec:supp_policy_refinement_quality}

From the policy-refinement trajectories with complete local image assets, we drew a simple random sample of 100 multi-subimage trajectories (seed 20260615). Sampling was not stratified by source shard or question type. Thirty trajectories were assigned to all five reviewers (M.L., M.K.K., S.M., A.G., and S.A.); the remaining 70 were assigned to exactly two reviewers, with seven trajectories allocated to each of the ten possible reviewer pairs. Each reviewer therefore assessed 58 trajectories.

\noindent Reviewers were blinded to the identity of the model used to construct the trajectories, the API provider, and the generation settings. They were shown the question, all candidate images, the instruction-derived reference selection, the structured reasoning trajectory, the refined reference answer and, when available, the associated caption or auxiliary context. The reviewer prompt instructed annotators to evaluate each structured reference trajectory independently using a five-point ordinal rubric across three dimensions: \emph{visual grounding and image triage}, \emph{biomedical reasoning}, and \emph{final clinical correctness and safety}. Dimension-specific descriptors anchored scores from 1 (\emph{unacceptable}) to 5 (\emph{fully satisfactory}), as summarized in Supplementary Figure~\ref{fig:policy_refinement_quality_rubric}.
Higher scores indicated stronger visual grounding, more coherent and evidence-based reasoning, and greater factual and clinical reliability.
For each trajectory and dimension, available reviewer scores were averaged before calculating the mean quality rating. Inter-reviewer agreement across the incomplete reviewer-assignment design was summarized using ordinal Krippendorff's $\alpha$. Ninety-five percent confidence intervals for the mean ratings and agreement coefficients were obtained from 10,000 instance-level bootstrap resamples.

\noindent We also conducted a secondary error-typing analysis within the same 100-trajectory sample. A trajectory entered this analysis when its mean reviewer score was 2 or lower in at least one dimension. One medical reviewer then re-examined each triggered dimension using prespecified, non-mutually-exclusive categories for visual-grounding, biomedical-reasoning, and final-answer errors. A material error was counted only when the low-score screen and the secondary review both indicated a substantive error. These screen-confirmed counts characterize the sampled trajectories and were not treated as exhaustive corpus-wide error rates.

\begin{figure}[H]
    \centering
    \scriptsize
    \setlength{\tabcolsep}{5pt}
    \renewcommand{\arraystretch}{1.10}
    \begin{tabularx}{0.98\textwidth}{@{}c>{\raggedright\arraybackslash}X@{}}
    \toprule
    \multicolumn{2}{@{}l}{\textbf{Visual grounding and image triage}: selection of question-relevant images and fidelity of visual claims} \\
    \midrule
    5 & \textbf{Fully grounded.} Selects all and only the relevant images; visual findings are precise and fully supported. \\
    4 & \textbf{Mostly grounded.} Selection and visual descriptions are correct, with only minor omissions or imprecision. \\
    3 & \textbf{Acceptable grounding.} The main evidence is supported, but minor omissions, unnecessary images, or imprecision are present. \\
    2 & \textbf{Substantial error.} Image selection or visual description contains important errors that limit reliability. \\
    1 & \textbf{Severe error.} Selection is largely incorrect, or major hallucinated or contradictory visual findings are present. \\
    \midrule
    \multicolumn{2}{@{}l}{\textbf{Biomedical reasoning}: logical and medically appropriate integration of visual evidence} \\
    \midrule
    5 & \textbf{Rigorous.} Clinically sound, logically structured, evidence-based, and directly relevant to the question. \\
    4 & \textbf{Sound.} Overall correct, with only minor gaps or slightly imprecise connections. \\
    3 & \textbf{Acceptable.} Core logic is reasonable, but links between evidence and conclusion are incomplete, generic, or imprecise. \\
    2 & \textbf{Substantial flaw.} Important logical errors or weakly supported conclusions are present. \\
    1 & \textbf{Severe flaw.} Reasoning is unsupported, incoherent, or clinically misleading. \\
    \midrule
    \multicolumn{2}{@{}l}{\textbf{Final clinical correctness and safety}: factual reliability, professionalism, and clinical safety of the final answer} \\
    \midrule
    5 & \textbf{Clinically sound.} Correct, complete, professional, and free of safety concerns. \\
    4 & \textbf{Mostly correct.} Minor imprecision is present but does not affect the key conclusion. \\
    3 & \textbf{Acceptable.} Generally correct, with minor factual noise, incomplete wording, or mild overstatement. \\
    2 & \textbf{Substantial error.} Contains important factual or clinical errors that limit its suitability as training data. \\
    1 & \textbf{Severe error.} Incorrect, unsafe, misleading, or otherwise unsuitable as training data. \\
    \bottomrule
    \end{tabularx}
    \caption{\textbf{Medical-reviewer rating rubric for policy-refinement trajectories.} Each trajectory was independently rated on a five-point ordinal scale across visual grounding and image triage, biomedical reasoning, and final clinical correctness and safety. Safety concerns included unsupported diagnostic or treatment claims, misleading clinical guidance, and other errors with potential for harm.}
    \label{fig:policy_refinement_quality_rubric}
\end{figure}

\begin{table}[H]
\centering
\caption{
\textbf{Manual assessment of policy-refinement trajectory quality.}
Five medical reviewers evaluated a simple random sample of 100
policy-refinement trajectories using a five-point scale. Thirty trajectories
were evaluated by all five reviewers, and each of the remaining 70
trajectories was evaluated by two reviewers. Values are mean ratings with
95\% confidence intervals obtained from 10,000 instance-level bootstrap
resamples. Ordinal Krippendorff's $\alpha$ summarizes inter-reviewer agreement across the available ratings.
}
\label{tab:policy_refinement_data_quality}

\small
\setlength{\tabcolsep}{12pt}
\renewcommand{\arraystretch}{1.15}

\begin{tabular}{lcc}
\toprule
\textbf{Evaluation dimension}
&
\textbf{Mean rating (95\% CI)}
&
\makecell{\textbf{Ordinal $\alpha$}\\\textbf{(95\% CI)}} \\
\midrule

Visual grounding and image triage
& 3.35 [3.09--3.62]
& 0.55 [0.31--0.70] \\

Biomedical reasoning
& 3.96 [3.76--4.15]
& 0.33 [0.09--0.53] \\

Final clinical correctness and safety
& 3.98 [3.77--4.18]
& 0.40 [0.15--0.59] \\

\bottomrule
\end{tabular}
\end{table}

\FloatBarrier

\subsection{PMC-MI-Bench curation and source-article-level decontamination}
\label{sec:supp_pmc_mi_bench_curation}
\noindent Benchmark candidates were sampled separately from the training resource and organized into six task categories: single-subimage VQA, multi-subimage VQA, relative-position VQA, compound-figure VQA, text-only QA, and multiple-choice VQA. Three medical reviewers (S.A., G.M., and M.G.) independently assessed source consistency, question clarity, reference-answer correctness, and task-format validity. A candidate was retained only when all three reviewers identified no issues; candidates requiring correction were excluded rather than edited or adjudicated into the benchmark.

\noindent One compound-figure VQA candidate and one text-only QA candidate were excluded during benchmark curation, yielding 298 retained examples: 50 single-subimage VQA, 50 multi-subimage VQA, 50 relative-position VQA, 49 compound-figure VQA, 49 text-only QA, and 50 multiple-choice VQA examples. We then audited provenance across the retained benchmark, all supervised instruction data, and Stage~II training and validation data. Source PMCIDs were compared across partitions, and every retained benchmark record had a distinct sample identifier linking its source PMCID and compound-figure image. This final audit removed 117 model-development records whose source articles occurred in the retained benchmark, comprising 112 supervised instances and five Stage~II policy-optimization instances. No source-article overlap remained between the final benchmark and the supervised or policy-refinement training and validation partitions. Image-content hashing, perceptual-image matching, and caption-similarity matching were not used in this audit. The benchmark was not used for checkpoint selection.

\FloatBarrier

\subsection{Evaluation setup}
\label{sec:supp_evaluation_setup}
\noindent The following subsections document task-specific prompts, decoding and parsing, reference-based metrics, evaluator protocols, target-task adaptation and uncertainty estimation. Common settings are stated once and task-specific deviations are reported with the corresponding benchmark.

\subsubsection{Evaluation on PMC-MI-Bench}
\label{sec:supp_pmc_bench_inference}

\paragraph{Inference and task-specific prompts.}
All models were evaluated using fixed checkpoints without
PMC-MI-Bench-specific adaptation or in-context demonstrations. Each model generated one response per benchmark example. Inference used greedy decoding with temperature 0, top-$p$ 1.0, a repetition penalty of 1.0, and a maximum of 2,048 newly generated tokens. The evaluation pipeline accepted up to eight input images, and no PMC-MI-Bench example was excluded on the basis of image count. Each model retained its native image-preprocessing procedure.

\noindent A task-specific system prompt was used for each of the six PMC-MI-Bench categories to specify the relevant evidence source and required response format. Within each category, the same system prompt was applied to all evaluated models. The prompts are reproduced as searchable text in Supplementary Figure~\ref{fig:supp_pmc_bench_prompts}.

\begin{figure}[H]
    \centering
    \begin{minipage}{0.96\linewidth}
\begin{lstlisting}[style=suppprompt]
TEXT-ONLY QA
You are a medical expert. Answer the following text-only medical question
accurately and concisely. Base your response on the information provided, use
complete sentences, and do not overstate the available evidence.

MULTIPLE-CHOICE VQA
You are a medical expert. Select the best answer to the following medical
multiple-choice question. Respond with the option letter only.

SINGLE-SUBIMAGE VQA
You are a medical expert. Answer the following question using the specified
subimage within the provided multi-image input. Base your response on the
relevant visual evidence and use concise, complete sentences.

RELATIVE-POSITION VQA
You are a medical expert. Determine the relative spatial position of the two
specified subimages from their locations within the provided compound figure.
Respond with the requested spatial relation only.

COMPOUND-FIGURE VQA
You are a medical expert. Answer the following question using the provided
compound medical figure. Base your response on the relevant visual evidence and
use concise, complete sentences.

MULTI-SUBIMAGE VQA
You are a medical expert. Answer the following question by integrating evidence
across the specified subimages. Base your response on the relevant visual
evidence and use concise, complete sentences.
\end{lstlisting}
    \end{minipage}
    \caption{
    \textbf{Task-specific system prompts used for PMC-MI-Bench evaluation.} The same system prompt was applied to all evaluated models within each task category; task inputs were appended by the evaluation pipeline.
    }
    \label{fig:supp_pmc_bench_prompts}
\end{figure}

\paragraph{Reference-based evaluation of open-ended responses.}
Semantic Textual Similarity (STS) was calculated using
\texttt{sentence-transformers/all-MiniLM-L6-v2}. Predictions and reference answers were encoded independently, and cosine similarity was calculated for each pair. Task-level scores were the arithmetic mean of the sample-level similarities and were multiplied by 100 for reporting.

\noindent BERTScore precision, recall, and F1 were calculated using \texttt{microsoft/deberta-xlarge-mnli} with \texttt{lang=en}. Baseline rescaling was not applied. BERTScore F1 was used as the principal BERTScore summary in the manuscript.

\noindent BLEU-1 through BLEU-4 were calculated at the sentence level using the \texttt{nltk} package. Predictions and references were lowercased, periods were separated from the preceding text, and the resulting strings were split on whitespace. BLEU-$n$ used uniform weights of $1/n$ over orders 1 through $n$. No smoothing function was applied; an $n$-gram precision of zero therefore produced a score of zero for the corresponding BLEU order.

\noindent ROUGE was calculated using the PyPI \texttt{rouge} implementation
(\texttt{from rouge import Rouge}) after lowercasing predictions and
references. ROUGE-L was used as the principal ROUGE summary. Unless otherwise specified, all reference-based metrics were calculated for each example and summarized using the arithmetic mean.

\paragraph{Multiple-choice accuracy.}
For multiple-choice VQA, each generated response was mapped to one of the available option letters and compared with the reference option. Responses that could not be mapped to a single valid option were scored as incorrect.

\paragraph{Uncertainty estimation.}
For open-ended PMC-MI-Bench performance, 95\% confidence intervals were obtained from 10,000 task-stratified bootstrap resamples of questions with replacement (seed 42). For paired model comparisons, the same resampled questions were applied to both models in each replicate. Overall MI-VQA was recalculated in every replicate as the unweighted mean of the single-subimage, multi-subimage, and relative-position VQA task means.

\noindent For multiple-choice accuracy, 95\% confidence intervals for individual models were calculated using the binomial Wilson method. Between-model differences were evaluated using paired question-level bootstrap resampling.

\subsubsection{Model-based and human assessment of open-ended responses}
\label{sec:supp_response_assessment}
\noindent Model-based and human evaluators used the same three-dimensional five-point rubric, enabling response quality to be summarized consistently across completeness, clarity, and correctness while retaining separate evaluator-specific analyses.

\paragraph{Rating rubric.}
Candidate responses were rated independently from 1 to 5 for completeness, clarity, and correctness. Completeness assessed coverage of the principal conclusion, supporting information, and relevant qualifications in the reference answer. Clarity assessed readability, logical organization, and precision. Correctness assessed consistency with the reference answer, including factual contradictions, unsupported material claims, and potentially safety-critical errors. Where clinically applicable, the latter included errors concerning laterality, severity, contraindications, or other statements that could materially alter clinical interpretation or management. The dimensions were scored separately: omissions primarily affected completeness, whereas incorrect or contradictory statements primarily affected correctness. Response length was not rewarded in itself, and the lower score was assigned when a response fell between two adjacent levels. The complete scoring rubric is shown in Supplementary Figure~\ref{fig:supp_response_rating_rubric}.

\begin{figure*}[p]
    \centering
    \scriptsize
    \setlength{\tabcolsep}{5pt}
    \renewcommand{\arraystretch}{1.10}
    \begin{tabularx}{0.98\textwidth}{@{}c>{\raggedright\arraybackslash}X@{}}
    \toprule
    \multicolumn{2}{@{}l}{\textbf{Completeness}: coverage of important information in the reference answer} \\
    \midrule
    5 & \textbf{Complete.} Covers all or nearly all key points, including the main conclusion and important qualifications; any omission is trivial. \\
    4 & \textbf{Mostly complete.} Covers the main conclusion and most supporting points; one or two secondary details or qualifications may be missing. \\
    3 & \textbf{Partially complete.} Covers the main conclusion or approximately half of the key points, but omits one or more important components. \\
    2 & \textbf{Minimally complete.} Mentions only a small portion of the relevant information; the main conclusion or several essential points are missing. \\
    1 & \textbf{Not complete.} Empty, off-topic, nonresponsive, or contains almost none of the relevant information. \\
    \midrule
    \multicolumn{2}{@{}l}{\textbf{Clarity}: readability, logical organization, and precision of the response} \\
    \midrule
    5 & \textbf{Very clear.} Precise, concise, well organized, and easy to follow; relationships between ideas are explicit. \\
    4 & \textbf{Mostly clear.} Generally easy to understand, with only minor awkward wording, redundancy, or organizational issues. \\
    3 & \textbf{Moderately clear.} The main meaning is understandable, but there is noticeable vagueness, repetition, awkward phrasing, or weak organization. \\
    2 & \textbf{Unclear.} Difficult to follow because of confusing structure, ambiguous wording, fragmented reasoning, or substantial redundancy. \\
    1 & \textbf{Incoherent.} Unintelligible, internally disorganized, or impossible to interpret reliably. \\
    \midrule
    \multicolumn{2}{@{}l}{\textbf{Correctness}: agreement with the reference answer, material impact of errors, and presence of safety-critical errors} \\
    \midrule
    5 & \textbf{Fully correct.} All substantive claims are consistent with the reference answer; no factual errors, contradictions, unsupported material claims, or safety-critical errors. \\
    4 & \textbf{Mostly correct.} The main conclusion and key claims are correct; only minor non-central imprecision or overstatement is present, with no safety-critical error. \\
    3 & \textbf{Partially correct.} A meaningful mix of correct and incorrect information, or a partly correct conclusion with important errors, but no clear safety-critical error. \\
    2 & \textbf{Mostly incorrect.} The central conclusion or multiple important claims conflict with the reference answer, or the response contains a safety-critical error that could plausibly lead to inappropriate clinical interpretation or management. \\
    1 & \textbf{Fundamentally incorrect.} Primarily incorrect, directly contradicts the reference answer, contains serious fabricated claims, or includes a severe safety-critical error with substantial potential for harm. \\
    \bottomrule
    \end{tabularx}
    \caption{\textbf{Rubric for model-based and human assessment of open-ended responses.} Responses were independently rated from 1 to 5 for completeness, clarity, and correctness relative to the manually verified reference answer. Safety-critical errors included claims that could plausibly lead to harmful clinical interpretation or management, including errors in laterality, severity, contraindications, treatment, or urgency.}
    \label{fig:supp_response_rating_rubric}
\end{figure*}

\paragraph{LLM-as-judge protocol.}
Three LLM judges were used: DeepSeek-V4-Pro, GPT-5.6, and Claude Opus 4.8. Each judge evaluated responses to all 248 questions from the five open-ended PMC-MI-Bench categories for M$^3$LLM, MedGemma-27B, and Lingshu-7B, yielding 744 candidate-response items. For each item, the judge received only the question, the manually verified reference answer, and one anonymized candidate response. Images, captions, auxiliary context, and candidate-model identity were not provided.

\noindent DeepSeek-V4-Pro was accessed through the DeepSeek OpenAI-compatible Chat Completions API with thinking mode disabled. GPT-5.6 was accessed through the Azure OpenAI Chat Completions API (API version \path{2024-12-01-preview}), and Claude Opus 4.8 was accessed through the Azure Anthropic Foundry Messages API. Evaluation was performed on 30 July 2026. Temperature was not set explicitly and therefore followed the default setting of each provider. Output length was limited to 1,024 tokens, and failed requests were retried up to five times.

\noindent The released evaluation code additionally provides an open-weight judge configuration for \path{deepseek-ai/DeepSeek-V4-Flash}. The configuration uses the same rubric and input fields, with thinking mode disabled, temperature 0, \texttt{top\_p}=1.0, a maximum of 1,024 newly generated tokens, random seed 42, and up to five request attempts. The results reported here use the three judges specified above.

\noindent DeepSeek-V4-Pro returned ratings for all 744 items. GPT-5.6 returned 743 ratings; one request remained unavailable after repeated Azure content-filter responses. Claude Opus 4.8 returned 739 ratings; five requests remained unavailable after repeated empty responses. Missing ratings were not imputed. All successfully returned structured outputs were parsed. Summary values were calculated as arithmetic means over the available item-level ratings for each judge, candidate model, and evaluation dimension.

\paragraph{Human evaluation protocol.}
Seventy questions were randomly sampled from the five open-ended PMC-MI-Bench categories. Responses from M$^3$LLM, MedGemma-27B, and Lingshu-7B were evaluated for each question, yielding 210 response items. Five medical reviewers (M.L., M.K.K., S.M., A.G., and S.A.) used the same completeness, clarity, and correctness rubric as the LLM judges.

\noindent Reviewers were shown only the question, reference answer, and anonymized candidate response. Images, auxiliary context, and candidate-model identity were withheld, and response items were presented in randomized order. Thirty response items were independently rated by all five reviewers, whereas each of the remaining 180 items was rated by exactly two reviewers. Each reviewer completed 102 ratings. No adjudication was performed before the agreement analysis. For items with multiple ratings, the item-level human score was the arithmetic mean of the available reviewer ratings. Model-level scores were then calculated as arithmetic means of the item-level human scores.

\paragraph{Confidence intervals for human ratings.}
For each question--response item, ratings from the available human reviewers were first averaged to obtain an item-level score. For each model and evaluation dimension, the 95\% confidence interval for the mean human rating was estimated by resampling these item-level scores with replacement 10,000 times. The interval bounds were defined as the 2.5th and 97.5th percentiles of the bootstrap mean distribution. The bootstrap procedure used a NumPy random-number generator initialized with seed 42 (\path{np.random.default_rng(42)}).

\paragraph{Agreement analysis.}
Inter-rater agreement among the five human reviewers was evaluated on the 30 fully crossed response items using ordinal Krippendorff's $\alpha$, the prespecified primary agreement statistic. Exact agreement and agreement within one scale point were additionally reported. Two-way random-effects, absolute-agreement intraclass correlation coefficients were calculated for an individual rating, ICC(A,1), and for the mean of five ratings, ICC(A,5).

\noindent Human-LLM agreement was assessed by comparing each LLM-judge rating with the item-level human mean using Spearman correlation. Quadratic-weighted Cohen's $\kappa$ was additionally calculated after rounding the human mean to the nearest integer on the 1--5 scale. Pairwise agreement among the three LLM judges was assessed using Spearman correlation and quadratic-weighted Cohen's $\kappa$, and three-way agreement was summarized using ordinal Krippendorff's $\alpha$. Confidence intervals were obtained from 10,000 question-clustered bootstrap resamples, retaining response items derived from the same question within the same resampled cluster.

\subsubsection{Evaluation on public medical benchmarks}
\label{sec:supp_public_benchmarks}
\noindent OmniMedVQA and MMMU-Med were evaluated using MedEvalKit (git commit \texttt{9b12e3b}) with the released model checkpoints and no benchmark-specific adaptation or in-context examples. Before OmniMedVQA inference and scoring, we reviewed the source-dataset provenance of the 88,996-question publicly accessible pool. Covid CT (199 question--answer items), CoronaHack (684), Covid19 heywhale (690), and COVIDx CXR-4 (485) were excluded because their documentation indicated potential inclusion of images collected from scientific publications or preprints. The resulting evaluation set contained 86,938 question--answer items from 38 source datasets across eight imaging modalities. MMMU-Med used the \texttt{MMMU-Medical-val} task, corresponding to the 150-question validation split of the Health and Medicine component of MMMU: Basic Medical Science, Clinical Medicine, Diagnostics and Laboratory Medicine, Pharmacy, and Public Health (30 questions per subject).

\noindent Evaluation used random seed 42, one inference pass per example, and reasoning mode disabled. Generation used greedy decoding with temperature 0, \texttt{top\_p}=0.0001, a repetition penalty of 1.0, and a maximum of 8,192 newly generated tokens. At most eight input images were permitted by the evaluation configuration. Predictions were scored using the default MedEvalKit answer-extraction procedure, and accuracy was calculated from the extracted option letters.

\subsubsection{Common target-task adaptation and inference protocol for
clinical evaluations}
\label{sec:supp_clinical_adaptation}
\noindent The same parameter-efficient adaptation and constrained decoding protocol was applied to all generative MLLMs within both clinical evaluations. Dataset-specific task construction, partitions, and specialized baselines are described separately below.

\paragraph{Target-task adaptation of generative MLLMs.}
For the MS-CXR-T-derived benchmark and the YNHH dermatology cohort, generative MLLMs were adapted by supervised fine-tuning with low-rank adaptation (LoRA) using LLaMA-Factory (version \texttt{0.9.5.dev0}). LoRA was applied to eligible linear layers of the language-model backbone, including \texttt{q\_proj}, \texttt{k\_proj}, \texttt{v\_proj}, \texttt{o\_proj}, \texttt{gate\_proj}, \texttt{up\_proj}, and \texttt{down\_proj}. The remaining language-model parameters, vision tower, and multimodal projector were frozen.

\noindent We used a LoRA rank of $r=16$, a scaling parameter of $\alpha=32$, and a dropout rate of 0.05. Models were trained for three epochs using AdamW with an initial learning rate of $1\times10^{-4}$, zero weight decay, cosine learning-rate decay, and a warm-up ratio of 0.05. Training used bfloat16 precision, a per-device batch size of 1, and eight gradient-accumulation steps, yielding an effective batch size of 8 on one NVIDIA H200 GPU. The maximum sequence length was 2,048 tokens. Per-image pixel counts were constrained to $[1{,}024,\,\,589{,}824]$, corresponding to $32\times 32$ and $768\times 768$
pixels for square images, respectively.

\noindent Within each dataset and task, all generative MLLMs received the same training examples and optimization settings. The optimization settings were also held constant across the two clinical evaluations. No validation partition was used for target-task adaptation, and the final checkpoint after the third epoch was evaluated. No held-out test label was used for model selection or hyperparameter adjustment.

\paragraph{Inference and answer parsing.}
For each generative model, evaluation without and after target-task adaptation used identical task prompts, image ordering, decoding settings, and answer-parsing procedures. Inference used greedy decoding with temperature 0, \texttt{top\_p}=0.0001, a repetition penalty of 1.0, and a maximum of 128 newly generated tokens.

\noindent Generated responses were mapped to the permitted task labels using a prespecified deterministic parser. Responses that could not be mapped to a valid label were treated as incorrect for accuracy calculations.

\subsubsection{Evaluation on the MS-CXR-T-derived longitudinal
chest-radiograph benchmark}
\label{sec:supp_mscxrt_evaluation}
\noindent This section details the patient-level split, derived task prompts, specialized representation-model implementations, and patient-clustered uncertainty analysis used for the longitudinal evaluation.

\paragraph{Benchmark construction and data partitioning.}
The benchmark was constructed from temporally ordered chest-radiograph pairs and finding-specific progression annotations provided by MS-CXR-T~\cite{bannur2023mscxr}. It contained 1,326 temporal image--label examples covering consolidation, edema, pleural effusion, pneumonia, and pneumothorax. Each example comprised a prior radiograph, a subsequent radiograph from the same patient, an annotated target finding, and a progression label of \emph{improving}, \emph{stable}, or \emph{worsening}.

\noindent MS-CXR-T extends the Chest ImaGenome progression resource. In the source curation, candidate progression labels derived from radiology reports were reviewed by a board-certified radiologist using the current report and its supporting sentence. A subset was independently relabelled by a second blinded radiologist, and the released label-quality field distinguishes one-expert, multiple-expert, and disagreement annotations. When multiple frontal scans were available for a study, a radiologist selected the image used in the benchmark. We used the released annotations and image pairs without additional relabelling or adjudication.

\noindent Examples were assigned in a $1{:}1$ ratio to training and held-out test partitions by patient-level simple random sampling using seed 42. All examples from the same patient were assigned to the same partition. The resulting partitions each contained 663 examples. The test partition therefore yielded 663 progression-classification instances and 3,315 target-finding-identification queries.

\begin{table}[!htbp]
\centering
\caption{\textbf{Progression-label distribution in the MS-CXR-T-derived benchmark.}}
\label{tab:mscxrt_progression_distribution}
\small
\setlength{\tabcolsep}{10pt}
\renewcommand{\arraystretch}{1.12}
\begin{tabular}{lrrrr}
\toprule
\textbf{Partition} & \textbf{$n$} & \textbf{Improving} & \textbf{Stable} & \textbf{Worsening} \\
\midrule
Training & 663 & 124 (18.7\%) & 249 (37.6\%) & 290 (43.7\%) \\
Held-out test & 663 & 112 (16.9\%) & 303 (45.7\%) & 248 (37.4\%) \\
\bottomrule
\end{tabular}
\end{table}

\paragraph{Task construction and prompts.}
For progression classification, the input comprised the prior and subsequent radiographs and the annotated target finding. Models were instructed to return \emph{improving}, \emph{stable}, or \emph{worsening}. For target-finding identification, five queries were constructed for each image pair by separately querying consolidation, edema, pleural effusion, pneumonia, and pneumothorax. The finding associated with the source MS-CXR-T annotation was assigned \emph{Yes}, whereas the remaining four findings were assigned \emph{No}. The prompts are reproduced as searchable text in Supplementary Figure~\ref{fig:ms-cxr-prompt}.

\begin{figure*}[t]
    \centering
    \begin{minipage}{0.90\linewidth}
\begin{lstlisting}[style=suppprompt]
PROGRESSION CLASSIFICATION
You are a medical AI assistant interpreting paired longitudinal chest
radiographs.
Previous examination (Fig-0): <image>
Current examination (Fig-1): <image>
Question: Relative to the previous examination, is the patient's <finding>
improving, stable, or worsening?
Respond with exactly one label: improving, stable, or worsening.

TARGET-FINDING IDENTIFICATION
You are a medical AI assistant interpreting paired longitudinal chest
radiographs.
Previous examination (Fig-0): <image>
Current examination (Fig-1): <image>
Question: Is <finding> the annotated target finding for this image pair?
Respond with exactly one label: Yes or No.
\end{lstlisting}
    \end{minipage}
    \caption{
    \textbf{Prompts for the MS-CXR-T-derived longitudinal chest-radiograph benchmark.} The templates specify the ordered image inputs and constrained output labels for progression classification and target-finding identification.}
    \label{fig:ms-cxr-prompt}
\end{figure*}

\paragraph{Specialized chest-radiograph baselines.}
The three specialized baselines were initialized from publicly released pretrained weights. Because their original studies did not provide a common classification head directly applicable to both derived tasks, we implemented architecture-appropriate task-specific heads on top of their native
representations. Where applicable, encoder fine-tuning and optimization settings were informed by the corresponding original studies and released codebase. All head architectures and hyperparameters were specified before held-out evaluation and were not adjusted after inspection of test-set results.

\paragraph{MoCo-CXR adaptation.}
We used the released ResNet-18 query encoder of MoCo-CXR and implemented ordered two-image late fusion. The shared image encoder mapped the prior and subsequent radiographs to separate 512-dimensional embeddings. These embeddings were concatenated with a learned 64-dimensional finding embedding and passed through
$\mathrm{LayerNorm}\rightarrow\mathrm{Linear}(1,088,512)
\rightarrow\mathrm{ReLU}\rightarrow\mathrm{Dropout}(0.2)
\rightarrow\mathrm{Linear}(512,C)$,
where $C=2$ for target-finding identification and $C=3$ for progression classification. The image encoder was updated during training. Models were trained using AdamW with a learning rate of $1\times10^{-4}$, weight decay of $3\times10^{-5}$, cosine decay, and a warm-up ratio of 0.05. Training was performed for 30 epochs using a batch size of 16, two gradient accumulation steps, bfloat16 precision, cross-entropy with inverse class-frequency weights, and seed 42.

\paragraph{CXR Foundation adaptation.}
For CXR Foundation, frozen ELIXR-C image embeddings were extracted once and mean-pooled to obtain a 128-dimensional representation for each radiograph. The temporal representation was
\[
[z_{\mathrm{prior}};z_{\mathrm{current}};\Delta;|\Delta|;
e_{\mathrm{finding}}],
\qquad
\Delta=z_{\mathrm{current}}-z_{\mathrm{prior}},
\]
where $e_{\mathrm{finding}}$ was a learned 32-dimensional finding embedding.
The resulting 544-dimensional vector was passed through
$\mathrm{LayerNorm}\rightarrow\mathrm{Linear}(544,512)
\rightarrow\mathrm{GELU}\rightarrow\mathrm{Dropout}(0.1)
\rightarrow\mathrm{Linear}(512,C)$. The image encoder remained frozen. The classification head was trained for 30 epochs using AdamW with a learning rate of $1\times10^{-4}$, weight decay of $1\times10^{-4}$, cosine annealing to $0.05$ times the initial learning rate, a batch size of 16, inverse class-frequency weights, and seed 42.

\paragraph{BioViL-T adaptation.}
Following the finding-specific formulation of BioViL-T, a separate model was adapted for each finding. The native temporal multi-image encoder produced a 128-dimensional projected global representation, which was passed through
$\mathrm{LayerNorm}\rightarrow\mathrm{Linear}(128,128)
\rightarrow\mathrm{GELU}\rightarrow\mathrm{Dropout}(0.1)
\rightarrow\mathrm{Linear}(128,C)$. The image encoder was updated during training. BioViL-T was trained for 30 epochs using AdamW with a learning rate of $2\times10^{-5}$, weight decay of $1\times10^{-4}$, a linear warm-up over the first 3\% of optimization steps, and subsequent linear decay. Training used $448\times448$ image crops, a per-device batch size of 4, 32 gradient-accumulation steps, cross-entropy with inverse class-frequency weights, bfloat16 precision, and seed 42. For all three specialized baselines, the final checkpoint from the prespecified training schedule was used for evaluation.

\paragraph{Evaluation measures and uncertainty estimation.}
Accuracy was the proportion of predictions assigned the correct class. Target-finding-identification accuracy was pooled over all $n=3{,}315$ binary decisions, whereas progression-classification accuracy was calculated over all $n=663$ three-class decisions. Balanced accuracy was the unweighted mean of class-specific recall, and macro-F1 was the unweighted mean of class-specific F1 scores.

\noindent Ninety-five percent confidence intervals were estimated from 10,000 patient-clustered bootstrap resamples with seed 42. In each replicate, patients were sampled with replacement from the held-out partition, and all image-pair examples and derived finding queries belonging to a sampled patient were retained together. The 2.5th and 97.5th percentiles of the bootstrap distribution defined the confidence interval. Percentages were calculated from the underlying predictions and rounded to two decimal places. The analyses were descriptive; no formal null-hypothesis significance tests
were performed.

\subsubsection{Evaluation on the retrospective YNHH dermatology cohort}
\label{sec:supp_ynhh_validation}
\noindent This section describes the image-only cohort inputs, expert-reviewed query construction, patient-level split, paired single- versus multi-image analysis, and case-clustered uncertainty estimation.

\paragraph{Cohort eligibility and model inputs.}
The retrospective dermatology validation cohort was derived from de-identified patient cases collected at Yale New Haven Hospital (YNHH)~\cite{jiang2026multimodal}. The final analysis cohort comprised 595 patient cases spanning 11 canonical diagnostic categories: morbilliform or exanthematous drug eruption, contact dermatitis, atopic dermatitis, bullous pemphigoid, psoriasis, cellulitis, pyoderma gangrenosum, leukocytoclastic vasculitis, stasis dermatitis or lipodermatosclerosis, hidradenitis suppurativa, and urticaria. All clinical photographs associated with each eligible case were retained during cohort construction, including cases containing more than eight photographs.

\noindent Model training and inference accepted a maximum of eight photographs per case. When more than eight were available, eight photographs were selected at approximately uniform positions across the fixed within-case sequence using NumPy \texttt{linspace}, thereby sampling across the beginning, middle, and end of the sequence. Clinical notes, demographic variables, medical history, laboratory findings, and other patient-associated textual information were not provided. Each model retained its native image-preprocessing procedure.

\paragraph{Construction of differential-diagnosis queries.}
Diagnostic labels were normalized before query construction so that capitalization, punctuation, and formatting variants of the same diagnosis were mapped to one canonical category. Each retained case was associated with one expert-annotated primary diagnosis, which was used to construct one positive query. Two candidate negative diagnoses were sampled without replacement from the remaining 10 canonical categories. Sampling was reproducible: the random state for each case was derived from the global seed of 42 and the corresponding case identifier.

\noindent Each sampled negative diagnosis was reviewed by medical experts and retained only when it was confirmed not to constitute a clinically valid differential diagnosis for the corresponding case. Each case therefore contributed three binary queries: one positive query with the reference answer \emph{Yes} and two expert-confirmed negative queries with the reference answer \emph{No}. The query template was: \emph{Is $<$diagnosis$>$ a valid differential for this image? Answer with a single word: Yes or No.}

\paragraph{Patient-level data partitioning and evaluation regimes.}
Cases were assigned to training and held-out test partitions by patient-level simple random sampling using seed 42, without stratification by diagnosis. All photographs and diagnosis queries associated with a patient were retained within the same partition. The resulting approximately 80:20 split contained 473 training cases and 122 held-out test cases. The test partition contained 366 case--diagnosis queries, comprising 122 positive and 244 negative queries. It was not used for model fitting, checkpoint selection, or hyperparameter selection.

\noindent Without target-task adaptation, released checkpoints were applied directly to the held-out test partition without exposure to the YNHH training partition. With target-task adaptation, models were trained exclusively on the 473-case training partition using the common procedure described in Supplementary Methods~\ref{sec:supp_clinical_adaptation} and were evaluated on the same 122 held-out cases.

\paragraph{Single-image versus multi-image input analysis.}
The contribution of multiple photographs was examined in a paired analysis without target-task adaptation. This analysis was restricted to 116 held-out cases containing at least two photographs, yielding 348 case--diagnosis queries. In the single-image condition, the model received the first photograph according to the prespecified within-case ordering rule. In the multi-image condition, it received up to eight photographs from the same case, with approximately uniform sequence-wide subsampling when more than eight were available. Questions, reference labels, model checkpoints, inference settings, and answer parsing were otherwise identical. Comparisons were paired at the case level.

\paragraph{Analysis by the number of available photographs.}
Performance without target-task adaptation was additionally summarized for held-out cases containing 2--3, 4--5, or at least 6 available photographs. Model inputs followed the same eight-image rule, including approximately uniform sequence-wide subsampling when more than eight photographs were available. Because the strata comprised different cases rather than repeated evaluations of the same cases, between-stratum differences were interpreted descriptively and not as causal effects of increasing the number of input photographs.

\paragraph{Evaluation measures and uncertainty estimation.}
Accuracy, balanced accuracy, and macro-F1 were calculated over the 366 held-out case--diagnosis queries. Balanced accuracy was the unweighted mean of sensitivity and specificity, and macro-F1 was the unweighted mean of the F1 scores for the \emph{Yes} and \emph{No} classes. The three queries derived from the same case were treated as correlated observations.

\noindent Where reported, 95\% confidence intervals were percentile intervals from 10,000 case-clustered bootstrap resamples with seed 42. Cases were sampled with replacement, and all three queries from a sampled case were retained together within each replicate.

\paragraph{Data governance.}
All processing, target-task adaptation, and inference involving YNHH photographs were performed within the local institutional computing environment. Images were screened for visible protected health information by at least two reviewers, and flagged images were excluded. No YNHH photograph or patient identifier was transmitted to an externally hosted application programming interface. Use of the cohort was conducted under the institutional determination reported in the main Methods.

\FloatBarrier

\section{Supplementary Notes}
\noindent The Supplementary Notes report memorization analyses, additional reference-based metrics, detailed evaluator agreement, class-balanced longitudinal evaluation, complete dermatology-cohort analyses, and representative qualitative comparisons.
\subsection{Supplementary Note 1: PMC-MI-Bench memorization analysis}
\label{sec:supp_memorization_results}
\noindent In the question-only analysis, verbatim memorization was uncommon for both M$^3$LLM and the Qwen3-VL-8B backbone under the prespecified \texttt{exact@30} and \texttt{exact@50} criteria (Supplementary Table~\ref{tab:memorization_bootstrap}).

\begin{table}[H]
\centering
\caption{\textbf{Question-only verbatim memorization analysis on PMC-MI-Bench.} \texttt{exact@}$K$ estimates are equal-weight means of the four eligible task-specific match rates. Estimates and paired between-model differences are reported as percentages and percentage points, respectively, with 95\% source-article-clustered bootstrap confidence intervals.}
\label{tab:memorization_bootstrap}

\small
\setlength{\tabcolsep}{8pt}
\renewcommand{\arraystretch}{1.15}

\begin{tabularx}{0.82\textwidth}{
    @{}
    l
    >{\raggedright\arraybackslash}X
    c
    c
    @{}
}
\toprule
\textbf{Metric}
&
\textbf{Model or comparison}
&
\textbf{Estimate}
&
\textbf{95\% CI}
\\
\midrule

\multirow{3}{*}{\texttt{exact@30}}
&
M$^3$LLM
&
1.01
&
[0.00--2.58]
\\

&
Qwen3-VL-8B
&
0.50
&
[0.00--1.67]
\\

&
M$^3$LLM $-$ Qwen3-VL-8B
&
0.51
&
[0.00--1.67]
\\ \hline

\addlinespace[3pt]

\multirow{3}{*}{\texttt{exact@50}}
&
M$^3$LLM
&
0.54
&
[0.00--1.79]
\\

&
Qwen3-VL-8B
&
0.00
&
[0.00--0.00]
\\

&
M$^3$LLM $-$ Qwen3-VL-8B
&
0.54
&
[0.00--1.79]
\\

\bottomrule
\end{tabularx}

\vspace{3pt}

\begin{minipage}{0.82\textwidth}
\footnotesize
\textit{Notes:}
A response satisfied \texttt{exact@}$K$ when it shared with the reference answer an identical contiguous sequence of at least $K$ tokens at any position. Between-model differences were calculated within each source-article-clustered bootstrap replicate. The degenerate percentile interval for Qwen3-VL-8B at \texttt{exact@50} reflects the absence of an observed match in every empirical bootstrap resample and should not be interpreted as zero population-level uncertainty.
\end{minipage}
\end{table}
\FloatBarrier

\subsection{Supplementary Note 2: Representative failure modes in policy-refinement trajectories}
\noindent The secondary review confirmed material visual-grounding errors in 9 of 100 sampled trajectories, biomedical-reasoning errors in 6 of 100, and final-answer correctness errors in 8 of 100. Categories were non-mutually exclusive. Because the review was triggered by a mean score of 2 or lower in the corresponding dimension, these values are screen-confirmed counts within the sampled trajectories rather than exhaustive error rates. The principal errors involved unsupported visual findings, incorrect interpretation of scientific images, contradictions with the supplied evidence, and unsupported factual or causal claims.

\begin{figure}[H]
    \centering
    \includegraphics[width=0.95\linewidth]{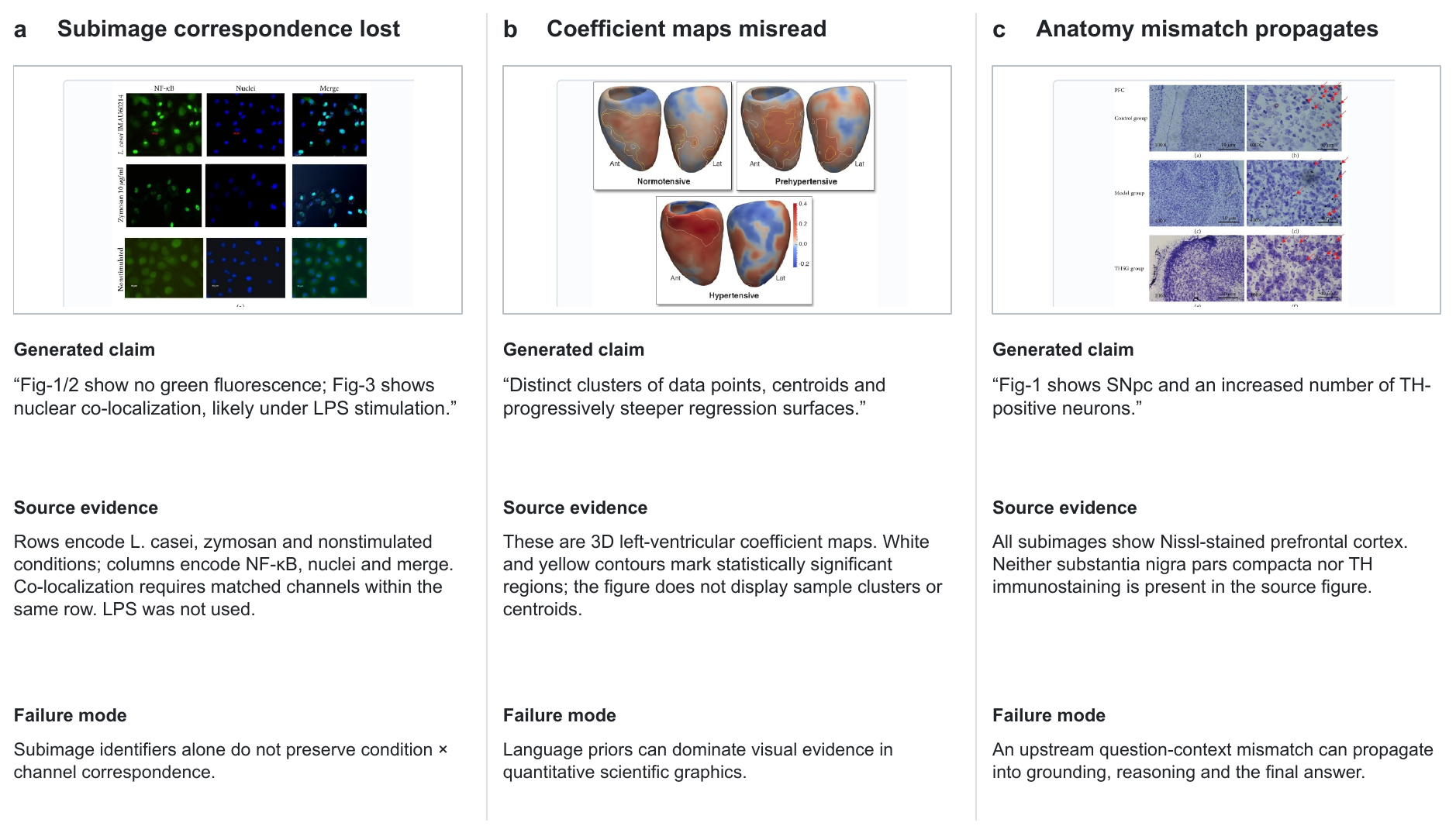}
    \caption{
    \textbf{Representative failure modes in policy-refinement trajectories.}
    Three low-scoring trajectories from the manual quality-assessment sample illustrate loss of subimage correspondence after compound-figure separation, misinterpretation of quantitative coefficient maps, and propagation of an upstream question--context mismatch. Red denotes unsupported generated content, blue denotes the interpretation supported by the source figure and associated text, and amber summarizes the corresponding methodological limitation. The examples illustrate error types identified in the secondary review; the screen-confirmed counts were calculated from the complete 100-trajectory sample. LPS, lipopolysaccharide; SNpc, substantia nigra pars compacta; TH, tyrosine hydroxylase.
    }
    \label{fig:policy_refinement_failures}
\end{figure}

\noindent Three representative cases illustrate distinct limitations involving subimage correspondence, interpretation of quantitative scientific graphics, and upstream question--context consistency (Supplementary Figure~\ref{fig:policy_refinement_failures}).

\FloatBarrier

\subsection{Supplementary Note 3: Additional reference-based metrics on PMC-MI-Bench}
\label{sec:supp_pmc_metrics}
\noindent To assess whether the comparative results were robust to metric choice, we additionally evaluated the open-ended PMC-MI-Bench responses using BLEU-4, ROUGE-L, and BERTScore F1 (Supplementary
Table~\ref{tab:pmc_mi_bench_additional_metrics}). These measures were treated as complementary reference-based diagnostics. Reference-aligned correctness was assessed separately using blinded human and LLM-based evaluations.

\begin{table*}[!htbp]
    \centering
    \caption{\textbf{Additional reference-based metrics on PMC-MI-Bench.} BLEU-4, ROUGE-L, and BERTScore F1 are reported on a 0--100 scale.}
\label{tab:pmc_mi_bench_additional_metrics}

    \scriptsize
    \setlength{\tabcolsep}{2.8pt}
    \renewcommand{\arraystretch}{1.12}

    \begin{tabularx}{\textwidth}{
        @{}
        >{\raggedright\arraybackslash}X
        c c c
        c c c
        c c c
        @{}
    }
        \toprule
        \textbf{Model}
        & \multicolumn{3}{c}{\textbf{Overall MI-VQA}}
        & \multicolumn{3}{c}{\textbf{Compound-figure VQA}}
        & \multicolumn{3}{c}{\textbf{Text-only QA}} \\
        \cmidrule(lr){2-4}\cmidrule(lr){5-7}\cmidrule(l){8-10}
        & \textbf{BLEU-4} & \textbf{ROUGE-L} & \makecell{\textbf{BERTScore}\\\textbf{F1}}
        & \textbf{BLEU-4} & \textbf{ROUGE-L} & \makecell{\textbf{BERTScore}\\\textbf{F1}}
        & \textbf{BLEU-4} & \textbf{ROUGE-L} & \makecell{\textbf{BERTScore}\\\textbf{F1}} \\
        \midrule

        \multicolumn{10}{@{}l}{\underline{\textit{General-purpose MLLMs}}} \\
        LLaVA-7B
        & 3.9 & 27.1 & 58.3
        & 2.3 & 22.0 & 55.3
        & 9.3 & 32.4 & 66.1 \\

        LLaVA-NeXT-7B
        & 4.5 & 26.5 & 59.3
        & 2.7 & 20.6 & 66.0
        & 9.8 & 35.2 & 68.3 \\

        Qwen2.5-VL-7B
        & 8.5 & 29.5 & 64.7
        & 3.4 & 23.5 & 55.5
        & 10.9 & 37.7 & 70.3 \\

        InternVL3-8B
        & 3.8 & 22.5 & 55.7
        & 6.8 & 29.5 & 58.7
        & 9.1 & 35.4 & 68.3 \\

        Qwen3-VL-8B
        & 3.8 & 24.3 & 63.7
        & 5.9 & 27.1 & 59.2
        & 10.3 & 38.1 & 71.1 \\

        \addlinespace[2pt]
        \multicolumn{10}{@{}l}{\underline{\textit{Medical MLLMs}}} \\
        LLaVA-Med-7B
        & 5.8 & 23.7 & 58.8
        & 11.6 & 34.5 & 67.2
        & 11.5 & 36.3 & 69.6 \\

        HuatuoGPT-Vision-7B
        & 9.1 & 31.4 & 66.9
        & 9.1 & 31.7 & 65.0
        & 11.3 & 37.9 & 69.1 \\

        Lingshu-7B
        & 9.8 & 30.4 & 66.8
        & 10.0 & 33.8 & 66.3
        & 11.0 & 38.8 & 68.3 \\

        HealthGPT-14B
        & 9.3 & 30.8 & 66.3
        & 9.8 & 34.0 & 67.1
        & 11.8 & 39.0 & 69.0 \\

        MedGemma-27B
        & 3.4 & 26.7 & 62.5
        & 2.3 & 19.0 & 51.9
        & 7.1 & 31.6 & 65.3 \\

        MedGemma~1.5-4B
        & 6.3 & 26.7 & 68.2
        & 2.9 & 20.3 & 69.6
        & 9.1 & 35.0 & \textbf{73.4} \\

        \addlinespace[2pt]
        \textbf{M$^3$LLM}
        & \textbf{21.2} & \textbf{38.5} & \textbf{73.7}
        & \textbf{16.8} & \textbf{38.3} & \textbf{71.4}
        & \textbf{11.5} & \textbf{39.5} & 73.1 \\
        \bottomrule
    \end{tabularx}

    \vspace{3pt}

    \begin{minipage}{0.99\textwidth}
        \footnotesize
        \textit{Notes:}
        MLLM, multimodal large language model; MI-VQA, multi-image visual question answering. BLEU-4 was calculated at the sentence level using uniform weights across one- to four-gram precisions and no smoothing. ROUGE-L denotes the F1 measure based on the longest common subsequence. BERTScore F1 was calculated using \texttt{microsoft/deberta-xlarge-mnli} without baseline rescaling. Metric implementation details are provided in the Supplementary Methods.
    \end{minipage}
\end{table*}

\FloatBarrier

\subsection{Supplementary Note 4: Human and model-based evaluation of open-ended responses}
\noindent Supplementary Table~\ref{tab:response_quality_summary} summarizes model rankings and evaluator agreement, followed by evaluator-specific reliability and agreement estimates in Supplementary Tables~\ref{tab:human_rating_reliability_detailed}--\ref{tab:llm_judge_agreement_detailed}.

\begin{table*}[!htbp]
    \centering
    \caption{\textbf{Human and model-based ratings of open-ended responses.} \textbf{(a)} Mean five-point ratings; 95\% confidence intervals are shown for the human ratings. \textbf{(b)} Agreement within and across evaluator groups. Bracketed intervals are 95\% confidence intervals; unbracketed ranges span the three LLM judges.}
\label{tab:response_quality_summary}

    \small
    \setlength{\tabcolsep}{5.5pt}
    \renewcommand{\arraystretch}{1.12}

    \textbf{(a)} Response-quality ratings

    \vspace{2pt}

    \begin{tabularx}{\textwidth}{
        @{}
        >{\raggedright\arraybackslash}X
        c c c
        @{}
    }
        \toprule
        \textbf{Candidate model}
        & \textbf{Completeness}
        & \textbf{Clarity}
        & \textbf{Correctness} \\
        \midrule

        \multicolumn{4}{@{}l}{LLM judge: \underline{\textit{DeepSeek-V4-Pro}}} \\[1pt]
        \textbf{M$^3$LLM} & \textbf{3.24} & \textbf{4.38} & \textbf{3.33} \\
        MedGemma-27B      & 2.92 & 4.16 & 2.59 \\
        Lingshu-7B        & 2.65 & 3.79 & 2.63 \\

        \addlinespace[2pt]
        \multicolumn{4}{@{}l}{LLM judge: \underline{\textit{GPT-5.6}}} \\[1pt]
        \textbf{M$^3$LLM} & \textbf{3.21} & \textbf{4.80} & \textbf{3.51} \\
        MedGemma-27B      & 2.91 & 4.54 & 2.88 \\
        Lingshu-7B        & 2.53 & 4.59 & 2.74 \\

        \addlinespace[2pt]
        \multicolumn{4}{@{}l}{LLM judge: \underline{\textit{Claude Opus 4.8}}} \\[1pt]
        \textbf{M$^3$LLM} & \textbf{3.15} & \textbf{4.21} & \textbf{3.36} \\
        MedGemma-27B      & 2.93 & 3.76 & 2.72 \\
        Lingshu-7B        & 2.43 & 3.79 & 2.52 \\

        \addlinespace[2pt]
        \toprule
        \textbf{Blinded human reviewers}
        & \textbf{Completeness}
        & \textbf{Clarity}
        & \textbf{Correctness} \\
        \midrule

        \textbf{M$^3$LLM}
        & \textbf{3.94} [3.73--4.14]
        & \textbf{4.79} [4.72--4.86]
        & \textbf{3.87} [3.62--4.12] \\
        MedGemma-27B
        & 3.80 [3.58--4.00]
        & 4.11 [3.94--4.27]
        & 3.12 [2.82--3.42] \\
        Lingshu-7B
        & 3.57 [3.32--3.81]
        & 4.45 [4.34--4.56]
        & 3.12 [2.80--3.43] \\

        \bottomrule
    \end{tabularx}

    \vspace{4mm}

    \textbf{(b)} Agreement summary

    \vspace{2pt}

    \scriptsize
    \setlength{\tabcolsep}{3.5pt}
    \begin{tabularx}{\textwidth}{
        @{}
        >{\raggedright\arraybackslash}X
        c c c c c
        @{}
    }
        \toprule
        \textbf{Dimension}
        & \makecell{\textbf{Human ordinal}\\$\boldsymbol{\alpha}$}
        & \textbf{ICC(A,5)}
        & \makecell{\textbf{Human-LLM}\\\textbf{Spearman $\rho$}}
        & \makecell{\textbf{Human-LLM}\\\textbf{weighted $\kappa$}}
        & \makecell{\textbf{Three-LLM}\\\textbf{ordinal $\alpha$}} \\
        \midrule
        Completeness
        & 0.37 [0.13--0.56]
        & 0.76 [0.46--0.87]
        & 0.55--0.72
        & 0.41--0.52
        & 0.81 [0.75--0.85] \\
        Clarity
        & 0.22 [0.07--0.35]
        & 0.62 [0.45--0.71]
        & 0.27--0.48
        & 0.20--0.40
        & 0.38 [0.31--0.44] \\
        Correctness
        & 0.61 [0.34--0.76]
        & 0.90 [0.78--0.95]
        & 0.77--0.83
        & 0.68--0.77
        & 0.85 [0.82--0.88] \\
        \bottomrule
    \end{tabularx}

    \vspace{3pt}

    \begin{minipage}{0.985\textwidth}
        \footnotesize
        \textit{Notes:}
        LLM, large language model; ICC, intraclass correlation coefficient. Human estimates are based on 70 response items per model (210 response items in total). DeepSeek-V4-Pro returned complete rating records for all 744 response items, GPT-5.6 returned records for 743, and Claude Opus 4.8 returned records for 739; missing records were not imputed. Each record contained separate completeness, clarity, and correctness scores. LLM and human ratings were analysed within their corresponding evaluation pools; direct evaluator comparisons used matched responses. ICC(A,5) denotes the absolute-agreement reliability of the five-rater mean. Human--LLM weighted $\kappa$ denotes quadratic-weighted Cohen's $\kappa$. Full evaluator-specific results are reported in the Supplementary Tables.
    \end{minipage}
\end{table*}

\begin{table*}[t]
    \centering
    \caption{\textbf{Reliability of human response-quality ratings.} Values are reported with 95\% question-clustered bootstrap confidence intervals. ICC, intraclass correlation coefficient.}
\label{tab:human_rating_reliability_detailed}

    \small
    \setlength{\tabcolsep}{5.0pt}
    \renewcommand{\arraystretch}{1.12}

    \begin{tabularx}{\textwidth}{
        @{}
        >{\raggedright\arraybackslash}X
        c c c c c
        @{}
    }
        \toprule
        \textbf{Dimension}
        & \makecell{\textbf{Ordinal}\\$\boldsymbol{\alpha}$}
        & \makecell{\textbf{Exact}\\\textbf{agreement}}
        & \makecell{\textbf{Agreement}\\\textbf{within 1}}
        & \textbf{ICC(A,1)}
        & \textbf{ICC(A,5)} \\
        \midrule
        Completeness
        & 0.37 [0.13--0.56]
        & 44.0\% [35.3--52.5]
        & 77.3\% [69.4--85.2]
        & 0.38 [0.15--0.58]
        & 0.76 [0.46--0.87] \\
        Clarity
        & 0.22 [0.07--0.35]
        & 55.3\% [45.2--65.9]
        & 85.3\% [77.8--92.2]
        & 0.24 [0.14--0.33]
        & 0.62 [0.45--0.71] \\
        Correctness
        & 0.61 [0.34--0.76]
        & 44.0\% [30.7--56.7]
        & 76.0\% [66.4--84.3]
        & 0.65 [0.42--0.79]
        & 0.90 [0.78--0.95] \\
        \bottomrule
    \end{tabularx}
\end{table*}

\begin{table*}[t]
    \centering
    \caption{\textbf{Human--LLM agreement by judge and rating dimension.} Values are reported with 95\% question-clustered bootstrap confidence intervals. LLM, large language model.}
\label{tab:human_llm_agreement_detailed}

    \small
    \setlength{\tabcolsep}{7.0pt}
    \renewcommand{\arraystretch}{1.10}

    \begin{tabularx}{\textwidth}{
        @{}
        >{\raggedright\arraybackslash}X
        l c c
        @{}
    }
        \toprule
        \textbf{LLM judge}
        & \textbf{Dimension}
        & \textbf{Spearman $\rho$}
        & \makecell{\textbf{Quadratic-weighted}\\$\boldsymbol{\kappa}$} \\
        \midrule

        DeepSeek-V4-Pro
        & Completeness & 0.55 [0.42--0.66] & 0.41 [0.31--0.51] \\
        & Clarity     & 0.27 [0.15--0.39] & 0.20 [0.10--0.30] \\
        & Correctness & 0.77 [0.70--0.82] & 0.68 [0.59--0.75] \\

        \addlinespace[2pt]
        GPT-5.6
        & Completeness & 0.72 [0.64--0.79] & 0.52 [0.43--0.60] \\
        & Clarity     & 0.48 [0.37--0.58] & 0.40 [0.32--0.48] \\
        & Correctness & 0.80 [0.72--0.85] & 0.76 [0.68--0.82] \\

        \addlinespace[2pt]
        Claude Opus 4.8
        & Completeness & 0.71 [0.62--0.78] & 0.51 [0.42--0.59] \\
        & Clarity     & 0.43 [0.31--0.53] & 0.37 [0.28--0.46] \\
        & Correctness & 0.83 [0.77--0.88] & 0.77 [0.70--0.83] \\

        \bottomrule
    \end{tabularx}
\end{table*}

\begin{table*}[t]
    \centering
    \caption{\textbf{Agreement among the three LLM judges.} \textbf{(a)} Pairwise agreement. \textbf{(b)} Three-way ordinal agreement. Values are reported with 95\% question-clustered bootstrap confidence intervals. LLM, large language model.}
\label{tab:llm_judge_agreement_detailed}

    \small
    \setlength{\tabcolsep}{7.0pt}
    \renewcommand{\arraystretch}{1.10}

    \textbf{(a)} Pairwise agreement

    \vspace{2pt}

    \begin{tabularx}{\textwidth}{
        @{}
        >{\raggedright\arraybackslash}X
        l c c
        @{}
    }
        \toprule
        \textbf{Judge pair}
        & \textbf{Dimension}
        & \textbf{Spearman $\rho$}
        & \makecell{\textbf{Quadratic-weighted}\\$\boldsymbol{\kappa}$} \\
        \midrule

        DeepSeek--GPT-5.6
        & Completeness & 0.74 [0.66--0.81] & 0.74 [0.65--0.81] \\
        & Clarity     & 0.48 [0.40--0.55] & 0.37 [0.29--0.44] \\
        & Correctness & 0.85 [0.82--0.87] & 0.82 [0.78--0.85] \\

        \addlinespace[2pt]
        DeepSeek--Claude
        & Completeness & 0.77 [0.69--0.83] & 0.77 [0.69--0.83] \\
        & Clarity     & 0.63 [0.56--0.69] & 0.62 [0.55--0.69] \\
        & Correctness & 0.86 [0.82--0.88] & 0.85 [0.81--0.88] \\

        \addlinespace[2pt]
        GPT-5.6--Claude
        & Completeness & 0.91 [0.88--0.93] & 0.91 [0.89--0.93] \\
        & Clarity     & 0.59 [0.52--0.65] & 0.37 [0.30--0.43] \\
        & Correctness & 0.91 [0.88--0.93] & 0.90 [0.87--0.92] \\

        \bottomrule
    \end{tabularx}

    \vspace{4mm}

    \textbf{(b)} Three-way agreement

    \vspace{2pt}

    \begin{tabularx}{0.70\textwidth}{
        @{}
        >{\raggedright\arraybackslash}X
        c
        @{}
    }
        \toprule
        \textbf{Dimension}
        & \makecell{\textbf{Ordinal Krippendorff's}\\$\boldsymbol{\alpha}$} \\
        \midrule
        Completeness & 0.81 [0.75--0.85] \\
        Clarity      & 0.38 [0.31--0.44] \\
        Correctness  & 0.85 [0.82--0.88] \\
        \bottomrule
    \end{tabularx}
\end{table*}

\FloatBarrier

\subsection{Supplementary Note 5: Class-balanced evaluation on the MS-CXR-T-derived benchmark}
\label{sec:supp_mscxrt_class_balanced_results}

\paragraph{Target-finding identification.}
\label{sec:supp_mscxrt_target_finding_class_balanced}
\noindent The held-out target-finding-identification set contained 3,315 binary decisions derived from 663 image pairs. Each image pair contributed one positive target-finding query and four negative non-target queries, yielding 663 positive and 2,652 negative decisions. Given this class imbalance, we report balanced accuracy and macro-F1 alongside the pooled micro-accuracy reported in the main text. Balanced accuracy was calculated as the unweighted mean of sensitivity and specificity, whereas macro-F1 was calculated as the unweighted mean of the positive- and negative-class F1 scores. For M$^3$LLM, balanced accuracy was 67.57\% without target-task adaptation and 79.03\% after adaptation; the corresponding macro-F1 values were 63.70\% and 73.42\% (Supplementary Table~\ref{tab:mscxrt_target_finding_class_balanced}).

\begin{table*}[!htbp]
\centering
\caption{
\textbf{Target-finding-identification performance on the MS-CXR-T-derived held-out test set.}
The test set comprised 3,315 binary finding decisions, including 663 positive target-finding queries and 2,652 negative non-target queries. Balanced accuracy is the unweighted mean of sensitivity and specificity, and macro-F1 is the unweighted mean of the positive- and negative-class F1 scores.
}
\label{tab:mscxrt_target_finding_class_balanced}

\scriptsize
\setlength{\tabcolsep}{6pt}
\renewcommand{\arraystretch}{1.06}

\begin{tabular}{@{}llccc@{}}
\toprule
\textbf{Model} &
\textbf{Target-task adaptation} &
\makecell{\textbf{Micro-accuracy}\\\textbf{(\%)}} &
\makecell{\textbf{Balanced accuracy}\\\textbf{(\%)}} &
\makecell{\textbf{Macro-F1}\\\textbf{(\%)}} \\
\midrule

\multicolumn{5}{@{}l}{\textit{Specialized chest-radiograph representation models}} \\

MoCo-CXR
& Yes
& 60.45
& 58.43
& 53.60 \\

CXR Foundation
& Yes
& 75.60
& 73.49
& 68.45 \\

BioViL-T
& Yes
& 73.45
& 71.42
& 66.23 \\

\addlinespace[2pt]
\multicolumn{5}{@{}l}{\textit{Generative MLLMs}} \\

LLaVA-7B
& No
& {46.70}
& {36.48}
& {37.18} \\

& Yes
& {61.30}
& {57.99}
& {53.79} \\

LLaVA-NeXT-7B
& No
& 48.14
& 37.61
& 38.20 \\

& Yes
& 63.20
& 60.14
& 55.68 \\

Qwen2.5-VL-7B
& No
& 52.31
& 42.08
& 41.92 \\

& Yes
& 65.61
& 63.12
& 58.24 \\

InternVL3-8B
& No
& 53.06
& 43.68
& 43.10 \\

& Yes
& 65.01
& 62.41
& 57.61 \\

Qwen3-VL-8B
& No
& 58.40
& 49.62
& 48.18 \\

& Yes
& 71.92
& 70.46
& 64.90 \\

LLaVA-Med-7B
& No
& {60.81}
& {53.51}
& {51.26} \\

& Yes
& {66.61}
& {63.52}
& {58.92} \\

HuatuoGPT-Vision-7B
& No
& 63.68
& 56.66
& 54.09 \\

& Yes
& 67.81
& 65.23
& 60.33 \\

Lingshu-7B
& No
& 56.38
& 46.49
& 45.69 \\

& Yes
& 59.61
& 56.77
& 52.45 \\

HealthGPT-14B
& No
& 60.03
& 51.77
& 49.98 \\

& Yes
& 71.49
& 69.80
& 64.37 \\

MedGemma-27B
& No
& 69.08
& 63.82
& 60.28 \\

& Yes
& 73.91
& 72.83
& 67.10 \\

MedGemma~1.5-4B
& No
& 62.87
& 55.02
& 52.85 \\

& Yes
& 78.67
& 77.68
& 72.20 \\

\textbf{M$^3$LLM}
& No
& \textbf{72.10}
& \textbf{67.57}
& \textbf{63.70} \\

& Yes
& \textbf{79.67}
& \textbf{79.03}
& \textbf{73.42} \\

\bottomrule
\end{tabular}
\end{table*}

\paragraph{Progression classification.}
\noindent The held-out progression-classification set contained 112 \emph{improving}, 303 \emph{stable}, and 248 \emph{worsening} examples (Supplementary Table~\ref{tab:mscxrt_progression_distribution}). In addition to the pooled micro-accuracies reported in the main text, we calculated balanced accuracy as the unweighted mean of the three class-specific recalls and macro-F1 as the unweighted mean of the three class-specific F1 scores. The resulting model-level metrics are reported in Supplementary Table~\ref{tab:mscxrt_progression_class_balanced}. The corresponding class-level confusion counts are provided in Supplementary Table~\ref{tab:mscxrt_progression_confusion}.

\begin{table*}[!htbp]
\centering
\caption{
\textbf{Progression-classification performance on the MS-CXR-T-derived
held-out test set.}
The test set comprised 663 examples. Balanced accuracy is the unweighted
mean of class-specific recall, and macro-F1 is the unweighted mean of
class-specific F1 scores.
}
\label{tab:mscxrt_progression_class_balanced}

\scriptsize
\setlength{\tabcolsep}{6pt}
\renewcommand{\arraystretch}{1.06}

\begin{tabular}{@{}llccc@{}}
\toprule
\textbf{Model} &
\textbf{Target-task adaptation} &
\makecell{\textbf{Micro-accuracy}\\\textbf{(\%)}} &
\makecell{\textbf{Balanced accuracy}\\\textbf{(\%)}} &
\makecell{\textbf{Macro-F1}\\\textbf{(\%)}} \\
\midrule

\multicolumn{5}{@{}l}{
\textit{Specialized chest-radiograph representation models}
} \\

MoCo-CXR
& Yes
& 40.72
& 38.74
& 38.78 \\

CXR Foundation
& Yes
& 48.57
& 46.43
& 46.59 \\

BioViL-T
& Yes
& 50.38
& 48.24
& 48.47 \\

\addlinespace[2pt]
\multicolumn{5}{@{}l}{\textit{Generative MLLMs}} \\

LLaVA-7B
& No
& 17.80
& 16.44
& 16.22 \\

& Yes
& 28.36
& 25.87
& 25.84 \\

LLaVA-NeXT-7B
& No
& 18.70
& 17.26
& 16.99 \\

& Yes
& 29.56
& 27.01
& 27.02 \\

Qwen2.5-VL-7B
& No
& 22.47
& 21.27
& 20.91 \\

& Yes
& 39.97
& 37.48
& 37.77 \\

InternVL3-8B
& No
& 23.53
& 22.30
& 21.93 \\

& Yes
& 43.44
& 40.80
& 41.18 \\

Qwen3-VL-8B
& No
& 20.81
& 19.58
& 19.27 \\

& Yes
& 49.32
& 46.74
& 47.24 \\

LLaVA-Med-7B
& No
& 24.28
& 23.13
& 22.75 \\

& Yes
& 42.99
& 40.28
& 40.68 \\

HuatuoGPT-Vision-7B
& No
& 25.19
& 23.98
& 23.57 \\

& Yes
& 43.74
& 41.20
& 41.57 \\

Lingshu-7B
& No
& 24.43
& 23.19
& 22.81 \\

& Yes
& 49.47
& 46.88
& 47.38 \\

HealthGPT-14B
& No
& 23.53
& 22.30
& 21.93 \\

& Yes
& 46.00
& 43.38
& 43.82 \\

MedGemma-27B
& No
& 25.79
& 24.47
& 24.08 \\

& Yes
& 55.81
& 53.18
& 53.88 \\

MedGemma~1.5-4B
& No
& 21.87
& 20.61
& 20.29 \\

& Yes
& 57.47
& 54.86
& 55.61 \\

\textbf{M$^3$LLM}
& No
& \textbf{27.15}
& \textbf{25.91}
& \textbf{25.47} \\

& Yes
& \textbf{60.78}
& \textbf{58.25}
& \textbf{59.10} \\

\bottomrule
\end{tabular}

\end{table*}

\begin{table*}[!htbp]
\centering
\caption{
\textbf{Class-level confusion counts for progression classification.}
Rows correspond to reference labels and columns to predicted labels.
For each model, row totals are 112 improving, 303 stable, and 248
worsening examples.
}
\label{tab:mscxrt_progression_confusion}

\scriptsize
\setlength{\tabcolsep}{3.3pt}
\renewcommand{\arraystretch}{1.06}

\begin{tabular}{@{}llccc ccc ccc@{}}
\toprule
& &
\multicolumn{3}{c}{\textbf{True improving} ($n=112$)} &
\multicolumn{3}{c}{\textbf{True stable} ($n=303$)} &
\multicolumn{3}{c}{\textbf{True worsening} ($n=248$)} \\
\cmidrule(lr){3-5}
\cmidrule(lr){6-8}
\cmidrule(l){9-11}

\textbf{Model} & \textbf{Adapt.} &
\textbf{Pred. I} & \textbf{Pred. S} & \textbf{Pred. W} &
\textbf{Pred. I} & \textbf{Pred. S} & \textbf{Pred. W} &
\textbf{Pred. I} & \textbf{Pred. S} & \textbf{Pred. W} \\
\midrule

MoCo-CXR
& Yes
& 37 & 58 & 17
& 44 & 147 & 112
& 29 & 133 & 86 \\

CXR Foundation
& Yes
& 45 & 52 & 15
& 37 & 172 & 94
& 26 & 117 & 105 \\

BioViL-T
& Yes
& 47 & 51 & 14
& 35 & 177 & 91
& 25 & 113 & 110 \\

\addlinespace[2pt]

LLaVA-7B
& No
& 14 & 69 & 29
& 80 & 70 & 153
& 64 & 150 & 34 \\

LLaVA-7B
& Yes
& 21 & 73 & 18
& 47 & 116 & 140
& 30 & 167 & 51 \\

LLaVA-NeXT-7B
& No
& 15 & 68 & 29
& 77 & 76 & 150
& 63 & 152 & 33 \\

LLaVA-NeXT-7B
& Yes
& 22 & 72 & 18
& 46 & 120 & 137
& 29 & 165 & 54 \\

Qwen2.5-VL-7B
& No
& 20 & 64 & 28
& 77 & 83 & 143
& 57 & 145 & 46 \\

Qwen2.5-VL-7B
& Yes
& 34 & 62 & 16
& 38 & 151 & 114
& 25 & 143 & 80 \\

InternVL3-8B
& No
& 21 & 64 & 27
& 76 & 86 & 141
& 56 & 143 & 49 \\

InternVL3-8B
& Yes
& 37 & 60 & 15
& 35 & 162 & 106
& 24 & 135 & 89 \\

Qwen3-VL-8B
& No
& 18 & 66 & 28
& 79 & 78 & 146
& 58 & 148 & 42 \\

Qwen3-VL-8B
& Yes
& 44 & 54 & 14
& 31 & 180 & 92
& 22 & 123 & 103 \\

LLaVA-Med-7B
& No
& 22 & 63 & 27
& 77 & 86 & 140
& 55 & 140 & 53 \\

LLaVA-Med-7B
& Yes
& 36 & 61 & 15
& 35 & 160 & 108
& 24 & 135 & 89 \\

HuatuoGPT-Vision-7B
& No
& 23 & 62 & 27
& 74 & 91 & 138
& 55 & 140 & 53 \\

HuatuoGPT-Vision-7B
& Yes
& 38 & 59 & 15
& 35 & 163 & 105
& 24 & 135 & 89 \\

Lingshu-7B
& No
& 22 & 63 & 27
& 75 & 89 & 139
& 55 & 142 & 51 \\

Lingshu-7B
& Yes
& 44 & 54 & 14
& 31 & 180 & 92
& 22 & 122 & 104 \\

HealthGPT-14B
& No
& 21 & 64 & 27
& 76 & 86 & 141
& 56 & 143 & 49 \\

HealthGPT-14B
& Yes
& 40 & 58 & 14
& 33 & 170 & 100
& 23 & 130 & 95 \\

MedGemma-27B
& No
& 23 & 62 & 27
& 74 & 93 & 136
& 54 & 139 & 55 \\

MedGemma-27B
& Yes
& 51 & 49 & 12
& 26 & 200 & 77
& 19 & 110 & 119 \\

MedGemma~1.5-4B
& No
& 19 & 65 & 28
& 78 & 81 & 144
& 57 & 146 & 45 \\

MedGemma~1.5-4B
& Yes
& 53 & 47 & 12
& 24 & 205 & 74
& 19 & 106 & 123 \\

\textbf{M$^3$LLM}
& No
& 25 & 61 & 26
& 72 & 97 & 134
& 53 & 137 & 58 \\

\textbf{M$^3$LLM}
& Yes
& 57 & 44 & 11
& 22 & 214 & 67
& 17 & 99 & 132 \\

\bottomrule
\end{tabular}

\vspace{3pt}
\begin{minipage}{0.99\textwidth}
\footnotesize
\textit{Notes:} Adapt., target-task adaptation; Pred., predicted; I, improving; S, stable; W, worsening.
\end{minipage}
\end{table*}

\FloatBarrier

\subsection{Supplementary Note 6: Additional evaluation on the YNHH dermatology cohort}
\label{sec:supp_ynhh_results}
\noindent Detailed results from the primary held-out evaluation, the paired comparison of single-image and multi-image inputs, and the image-count-stratified analysis are reported in Supplementary
Tables~\ref{tab:derm_clinical_validation}--\ref{tab:ynhh_image_count_stratified}.

\providecommand{\scoreci}[3]{\makecell{#1\\[-1pt]{\tiny [#2--#3]}}}
\providecommand{\bestscoreci}[3]{\makecell{\textbf{#1}\\[-1pt]{\tiny [#2--#3]}}}

\makeatletter
\setlength{\@fptop}{0pt}
\setlength{\@fpsep}{14pt}
\setlength{\@fpbot}{0pt plus 1fil}
\makeatother
\begin{table*}[p]
    \centering
    \caption{
    \textbf{Performance on the held-out Yale New Haven Hospital (YNHH) dermatology cohort.}
    Accuracy, balanced accuracy, and macro-F1 are reported as percentages with
    95\% case-clustered bootstrap confidence intervals.
    }
    \label{tab:derm_clinical_validation}

    \scriptsize
    \setlength{\tabcolsep}{5.5pt}
    \renewcommand{\arraystretch}{1.03}

    \begin{tabularx}{0.88\textwidth}{
        @{}
        >{\raggedright\arraybackslash}X
        c c c
        @{}
    }
        \toprule
        \textbf{Model}
        & \textbf{Accuracy}
        & \makecell{\textbf{Balanced}\\\textbf{accuracy}}
        & \textbf{Macro-F1} \\
        \midrule

        \multicolumn{4}{@{}l}{\textit{No target-task adaptation}} \\[2pt]

        LLaVA-7B
        & \scoreci{56.56}{51.91}{61.20}
        & \scoreci{56.76}{51.64}{61.89}
        & \scoreci{55.05}{50.43}{59.73} \\

        LLaVA-NeXT-7B
        & \scoreci{58.20}{53.28}{63.11}
        & \scoreci{58.81}{53.69}{63.93}
        & \scoreci{56.84}{52.02}{61.65} \\

        Qwen2.5-VL-7B
        & \scoreci{61.48}{57.38}{65.57}
        & \scoreci{62.50}{58.20}{66.60}
        & \scoreci{60.22}{56.26}{65.19} \\

        InternVL3-8B
        & \scoreci{52.73}{48.36}{57.38}
        & \scoreci{60.45}{56.15}{64.75}
        & \scoreci{52.69}{48.05}{57.36} \\

        Qwen3-VL-8B
        & \scoreci{62.13}{59.84}{66.94}
        & \scoreci{63.88}{58.81}{68.85}
        & \scoreci{61.13}{56.25}{65.81} \\

        LLaVA-Med-7B
        & \scoreci{56.56}{51.37}{61.75}
        & \scoreci{59.43}{53.89}{64.75}
        & \scoreci{56.01}{50.83}{61.03} \\

        HuatuoGPT-Vision-7B
        & \scoreci{46.45}{42.90}{50.27}
        & \scoreci{59.02}{56.15}{62.09}
        & \scoreci{44.65}{40.18}{49.09} \\

        Lingshu-7B
        & \scoreci{57.65}{52.73}{62.57}
        & \scoreci{61.68}{56.56}{66.60}
        & \scoreci{57.34}{52.49}{62.00} \\

        HealthGPT-14B
        & \scoreci{60.98}{55.46}{66.12}
        & \scoreci{61.24}{55.74}{66.60}
        & \scoreci{59.42}{54.16}{64.52} \\

        MedGemma-27B
        & \scoreci{62.02}{56.56}{67.21}
        & \scoreci{61.48}{55.74}{67.01}
        & \scoreci{60.07}{54.62}{65.34} \\

        MedGemma~1.5-4B
        & \scoreci{64.11}{59.89}{68.22}
        & \scoreci{61.82}{57.45}{66.01}
        & \scoreci{61.18}{56.98}{65.23} \\

        \textbf{M$^3$LLM}
        & \bestscoreci{69.13}{64.75}{73.50}
        & \bestscoreci{69.88}{65.37}{74.39}
        & \bestscoreci{67.70}{63.29}{71.95} \\

        \addlinespace[3pt]
        \midrule
        \multicolumn{4}{@{}l}{\textit{Target-task adaptation using LoRA}} \\[2pt]

        LLaVA-7B
        & \scoreci{63.93}{59.02}{68.85}
        & \scoreci{62.30}{56.97}{67.62}
        & \scoreci{61.39}{56.28}{66.41} \\

        LLaVA-NeXT-7B
        & \scoreci{68.58}{64.21}{72.95}
        & \scoreci{67.01}{62.09}{71.93}
        & \scoreci{66.10}{61.38}{70.81} \\

        Qwen2.5-VL-7B
        & \scoreci{73.22}{68.58}{77.60}
        & \scoreci{71.31}{66.19}{76.43}
        & \scoreci{70.67}{65.75}{75.54} \\

        InternVL3-8B
        & \scoreci{72.40}{67.76}{76.78}
        & \scoreci{70.90}{65.78}{75.82}
        & \scoreci{70.03}{65.05}{74.84} \\

        Qwen3-VL-8B
        & \scoreci{75.41}{71.04}{79.51}
        & \scoreci{73.57}{68.85}{78.07}
        & \scoreci{72.97}{68.41}{77.37} \\

        LLaVA-Med-7B
        & \scoreci{70.22}{65.57}{74.59}
        & \scoreci{69.06}{64.13}{73.77}
        & \scoreci{67.97}{63.14}{72.58} \\

        HuatuoGPT-Vision-7B
        & \scoreci{70.49}{66.12}{74.86}
        & \scoreci{69.06}{64.34}{73.77}
        & \scoreci{68.11}{63.50}{72.58} \\

        Lingshu-7B
        & \scoreci{72.13}{67.21}{76.78}
        & \scoreci{70.49}{65.37}{75.41}
        & \scoreci{69.68}{64.66}{74.61} \\

        HealthGPT-14B
        & \scoreci{72.95}{68.03}{77.60}
        & \scoreci{70.90}{65.78}{75.82}
        & \scoreci{70.32}{65.25}{75.23} \\

        MedGemma-27B
        & \scoreci{76.23}{72.13}{80.33}
        & \scoreci{73.98}{69.26}{78.48}
        & \scoreci{73.63}{69.01}{78.01} \\

        MedGemma~1.5-4B
        & \scoreci{78.14}{74.04}{82.24}
        & \scoreci{73.98}{69.06}{78.69}
        & \scoreci{74.64}{69.74}{79.27} \\

        \textbf{M$^3$LLM}
        & \bestscoreci{80.05}{75.96}{83.88}
        & \bestscoreci{77.05}{72.34}{81.56}
        & \bestscoreci{77.32}{72.70}{81.70} \\

        \bottomrule
    \end{tabularx}

    \vspace{3pt}
    \begin{minipage}{0.88\textwidth}
        \footnotesize
        \textit{Notes:}
        The held-out test set comprised 122 cases and 366 case--diagnosis queries (122 positive and 244 negative). Confidence intervals were obtained by resampling cases, with the three queries from each case retained as one cluster. LoRA denotes low-rank adaptation. Bold values denote the highest point estimate within each evaluation condition.
    \end{minipage}
\end{table*}

\begin{table*}[t]
    \centering
    \caption{
    \textbf{Single-image and multi-image performance without target-task adaptation on the held-out Yale New Haven Hospital (YNHH) dermatology cohort.}
    Values are accuracy (\%); $\Delta$ denotes the difference in accuracy (in percentage points) between multi-image and single-image inputs, calculated using unrounded values.
    }
    \label{tab:ynhh_zero_shot_multi_image}

    \small
    \setlength{\tabcolsep}{8.0pt}
    \renewcommand{\arraystretch}{1.12}

    \begin{tabularx}{0.78\textwidth}{
        @{}
        >{\raggedright\arraybackslash}X
        c c c
        @{}
    }
        \toprule
        \textbf{Model}
        & \makecell{\textbf{Single-image}\\\textbf{input}}
        & \makecell{\textbf{Multi-image}\\\textbf{input}}
        & \boldmath$\Delta$\unboldmath \\
        \midrule

        LLaVA-7B
        & 56.61
        & 55.46
        & -1.15 \\

        LLaVA-NeXT-7B
        & 55.46
        & 56.32
        & +0.86 \\

        Qwen2.5-VL-7B
        & 61.49
        & 63.22
        & +1.72 \\

        InternVL3-8B
        & 53.16
        & 55.17
        & +2.01 \\

        Qwen3-VL-8B
        & 66.67
        & 66.38
        & -0.29 \\

        LLaVA-Med-7B
        & 55.46
        & 56.90
        & +1.44 \\

        HuatuoGPT-Vision-7B
        & 44.25
        & 46.84
        & +2.59 \\

        Lingshu-7B
        & 58.33
        & 58.91
        & +0.57 \\

        HealthGPT-14B
        & 60.63
        & 62.36
        & +1.73 \\

        MedGemma-27B
        & \textbf{66.67}
        & 62.07
        & -4.60 \\

        MedGemma~1.5-4B
        & 51.72
        & 65.42
        & \textbf{+13.70} \\

        \textbf{M$^3$LLM}
        & 64.94
        & \textbf{69.25}
        & +4.31 \\

        \bottomrule
    \end{tabularx}

    \vspace{3pt}
    \begin{minipage}{0.78\textwidth}
        \footnotesize
        \textit{Notes:}
        The analysis included 116 cases containing at least two photographs,
        corresponding to 348 case--diagnosis queries. Single-image
        and multi-image conditions were evaluated on the identical set of
        queries. Multi-image input was capped at eight photographs; cases with
        more than eight available photographs were represented by approximately
        uniform sequence-wide subsampling. Accuracy was calculated as the number of correct predictions
        divided by 348, and $\Delta$ was calculated from the corresponding
        unrounded accuracies. Bold values denote the highest point estimate
        in each column.
    \end{minipage}
\end{table*}

\begin{table}[t]
    \centering
    \caption{
    \textbf{Performance without target-task adaptation stratified by the number of available photographs.}
    Accuracy (\%) is reported for held-out Yale New Haven Hospital (YNHH) cases containing at least two photographs. Six representative models are shown for readability.
    }
    \label{tab:ynhh_image_count_stratified}

    \small
    \setlength{\tabcolsep}{7.0pt}
    \renewcommand{\arraystretch}{1.12}

    \begin{tabular}{@{}lccc@{}}
        \toprule
        \textbf{Model}
        & \makecell{\textbf{2--3 available}\\\textbf{photographs}\\$n=21$ cases\\63 queries}
        & \makecell{\textbf{4--5 available}\\\textbf{photographs}\\$n=20$ cases\\60 queries}
        & \makecell{\boldmath$\geq 6$\unboldmath\textbf{ available}\\\textbf{photographs}\\$n=75$ cases\\225 queries} \\
        \midrule

        Qwen3-VL-8B          & 60.32 & 73.33 & 63.84 \\
        HuatuoGPT-Vision-7B  & 44.44 & 53.33 & 45.78 \\
        Lingshu-7B           & 53.97 & 66.67 & 58.67 \\
        MedGemma-27B         & \textbf{73.02} & 73.33 & 56.00 \\
        MedGemma~1.5-4B      & 66.67 & 70.00 & 66.22 \\
        \textbf{M$^3$LLM}   & 69.84 & \textbf{75.00} & \textbf{67.56} \\

        \bottomrule
    \end{tabular}

    \vspace{3pt}
    \begin{minipage}{0.94\linewidth}
        \footnotesize
        \textit{Notes:}
        Each case contributed three case--diagnosis queries. Model input was capped at eight photographs; cases with more than eight available photographs were represented by approximately uniform sequence-wide subsampling. The stratified results are descriptive because the number of available photographs may covary with case complexity and clinical acquisition practice. Bold values denote the highest point estimate in each column.
    \end{minipage}
\end{table}

\FloatBarrier
\clearpage

\subsection{Supplementary Note 7: Representative qualitative comparisons}
\begin{figure}[H]
\centering
   \includegraphics [width=0.85\textwidth]{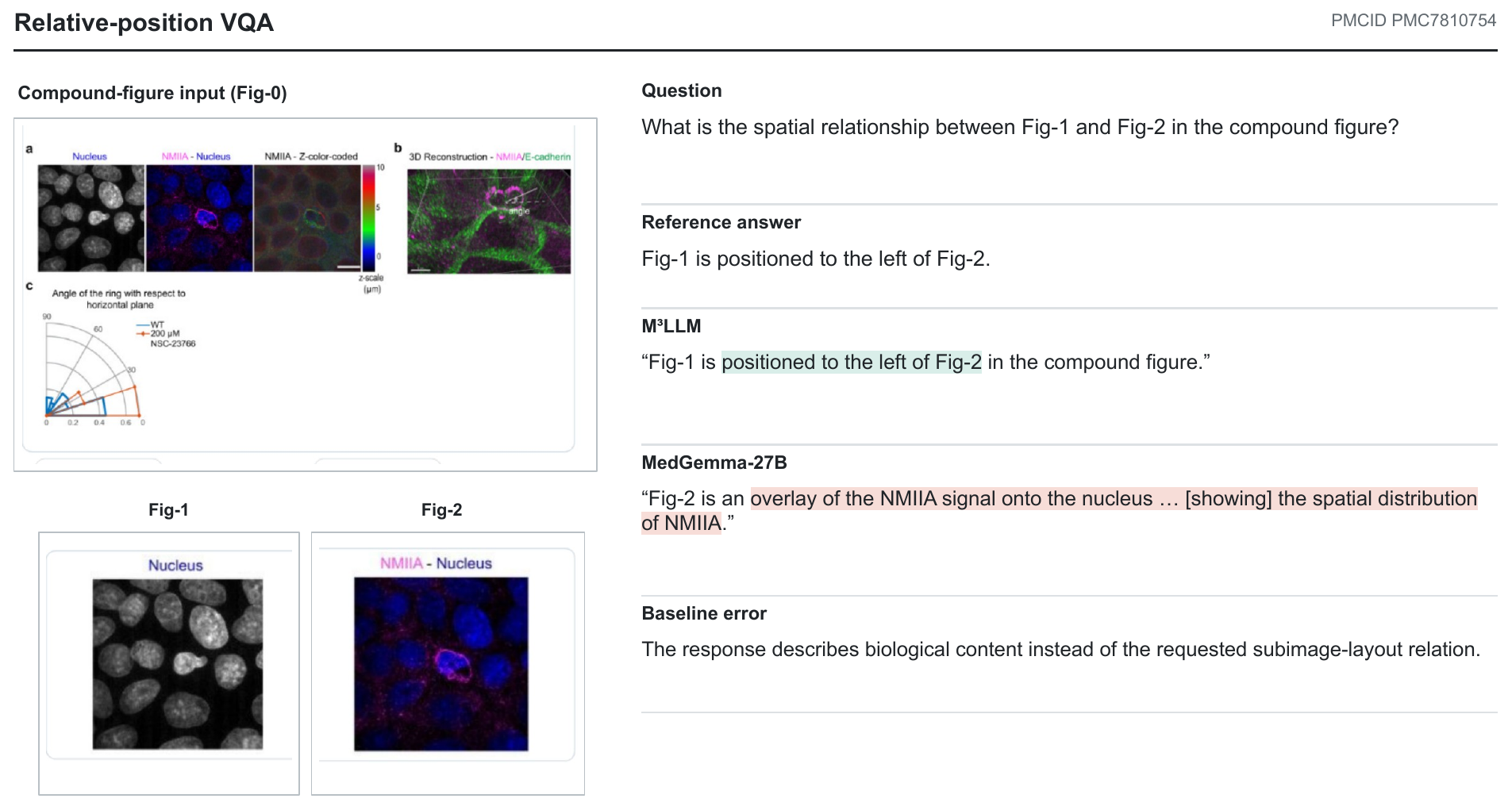}
\caption{
\textbf{Illustrative relative-position visual question answering (VQA) example.}
M$^3$LLM correctly identified Fig-1 as being positioned to the left of Fig-2, whereas MedGemma-27B interpreted the biological content of the subimages rather than their relative layout. NMIIA, non-muscle myosin IIA.
}
\label{fig:supp_relative_position}
\end{figure}

\begin{figure}[H]
\centering
   \includegraphics [width=0.85\textwidth]{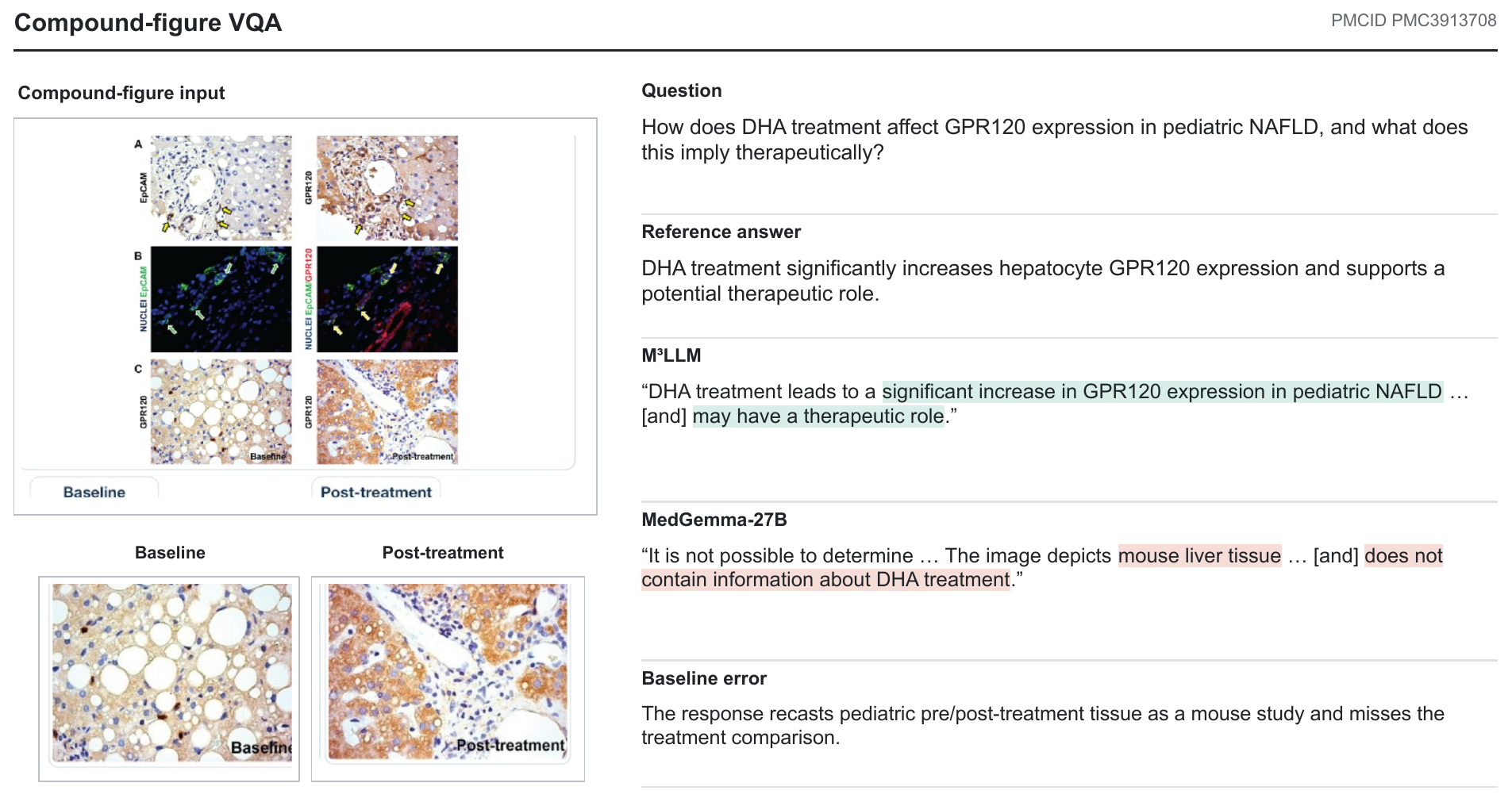}
\caption{
\textbf{Illustrative compound-figure visual question answering (VQA) example.}
M$^3$LLM preserved the paediatric pre- and post-treatment context and
identified increased GPR120 expression after DHA treatment. MedGemma-27B
misidentified the source cohort and did not recover the treatment comparison. GPR120, G-protein-coupled receptor 120; DHA, docosahexaenoic acid; NAFLD, non-alcoholic fatty liver disease.
}
\label{fig:supp_compound_figure}
\end{figure}

\begin{figure}[H]
\centering
   \includegraphics [width=0.85\textwidth]{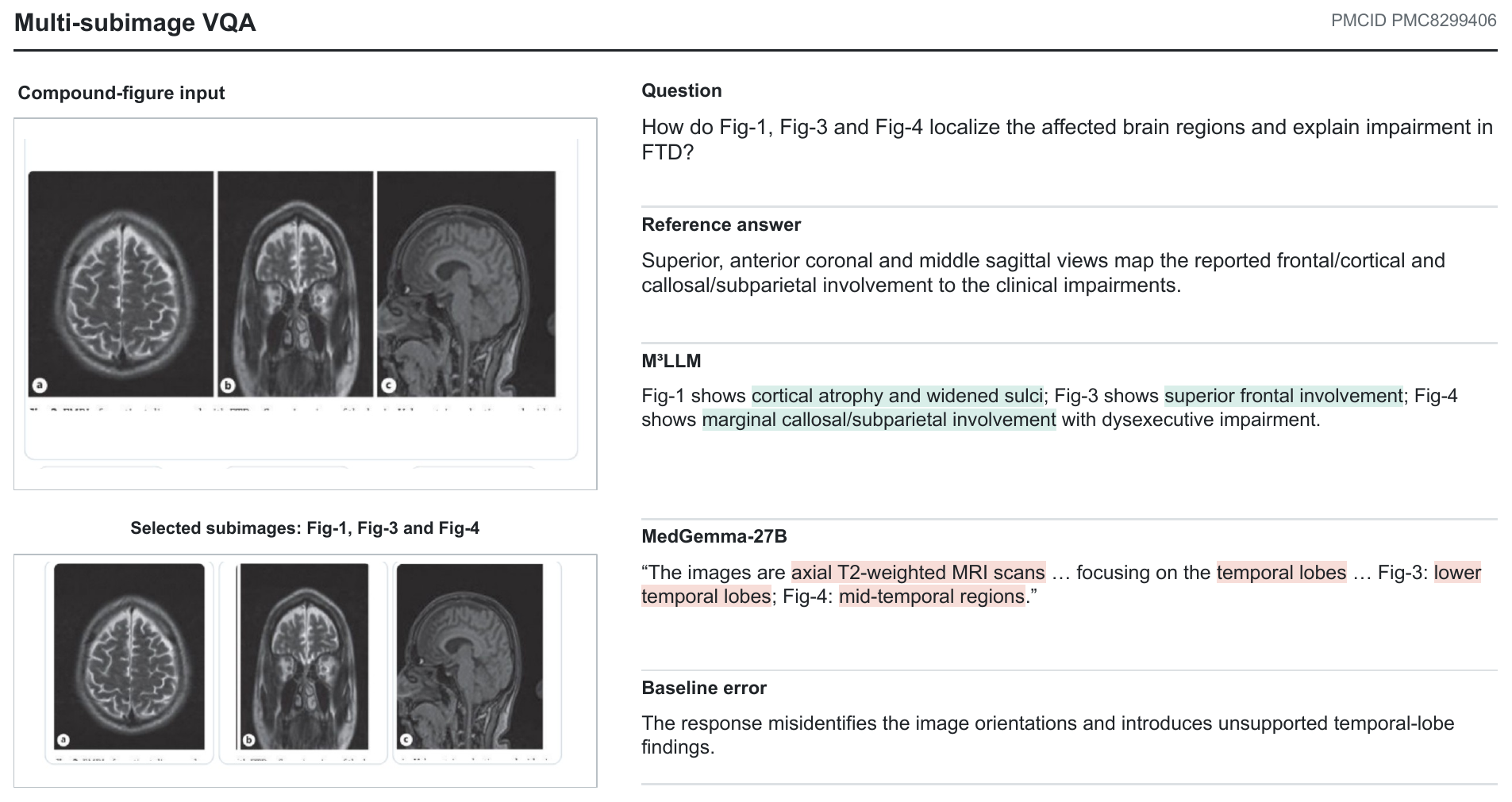}
\caption{
\textbf{Illustrative multi-subimage visual question answering (VQA) example.}
M$^3$LLM integrated the specified MRI views to localize the reported frontal and callosal/subparietal involvement, whereas MedGemma-27B misidentified the image orientations and introduced unsupported temporal-lobe findings. MRI, magnetic resonance imaging; FTD, frontotemporal dementia.
}
\label{fig:supp_multi_subimage}
\end{figure}

\begin{figure}[H]
\centering
   \includegraphics [width=0.85\textwidth]{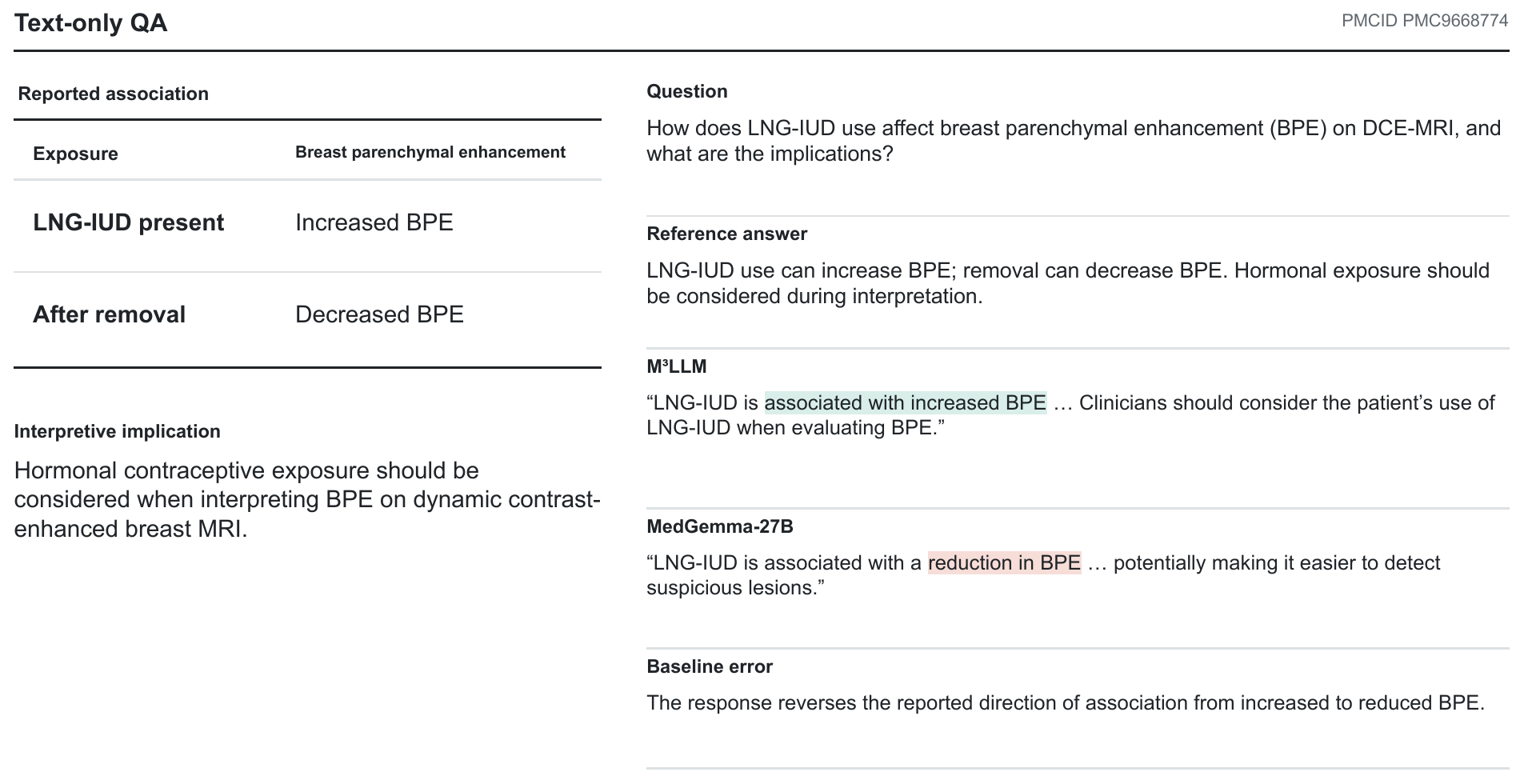}
\captionsetup{width=0.95\textwidth}
\caption{
\textbf{Illustrative text-only question answering (QA) example.}
M$^3$LLM retained the reported association between LNG-IUD use and increased breast parenchymal enhancement, whereas MedGemma-27B reversed the direction of this association. LNG-IUD, levonorgestrel-releasing intrauterine device.
}
\label{fig:supp_text_only}
\end{figure}

\begin{figure}[H]
\centering
   \includegraphics [width=0.85\textwidth]{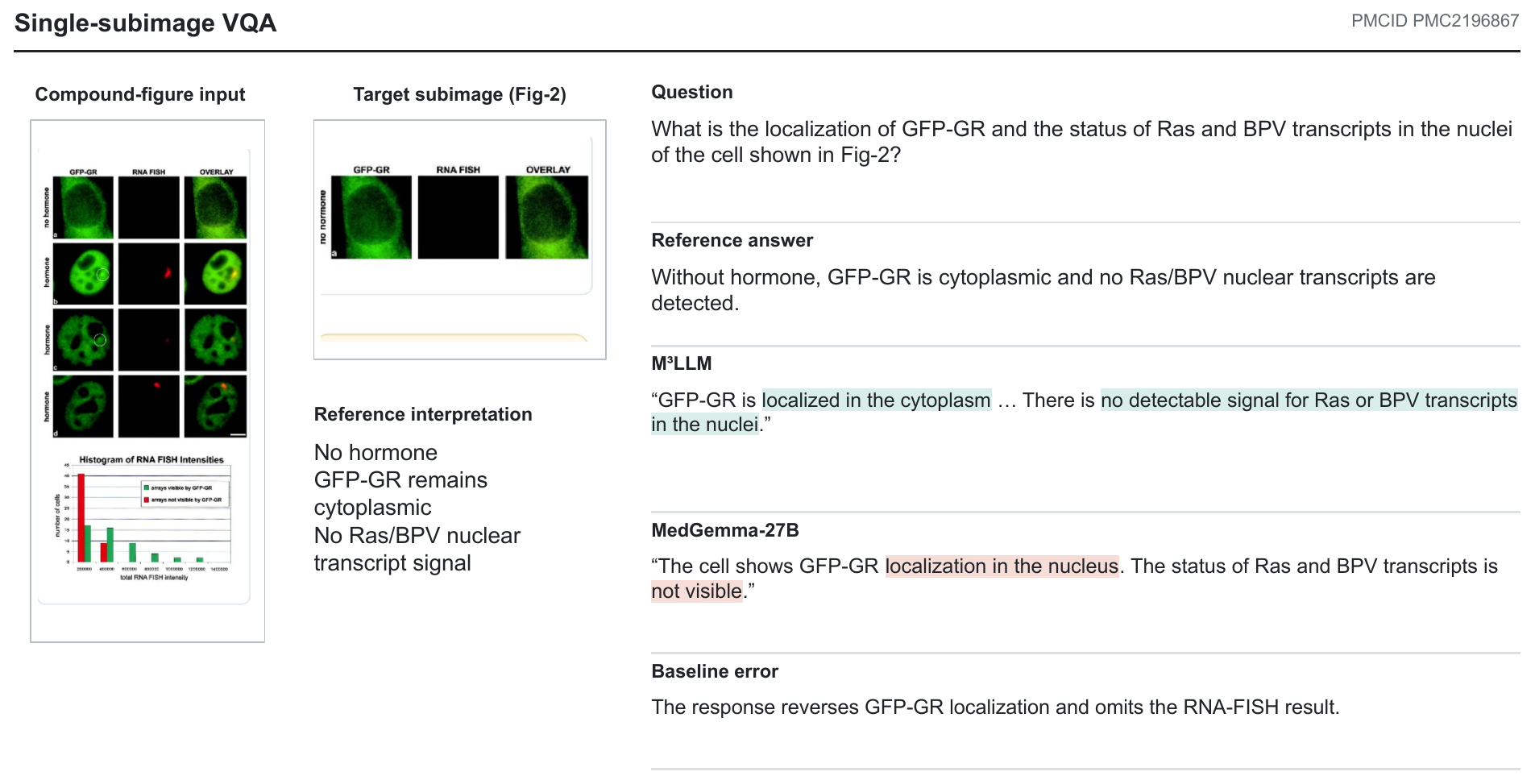}
\caption{
\textbf{Illustrative single-subimage visual question answering (VQA) example.}
M$^3$LLM correctly localized GFP--GR to the cytoplasm and identified the absence of nuclear Ras and BPV transcript signals. MedGemma-27B reversed the localization and did not recover the RNA-FISH finding. GFP--GR, green fluorescent protein--glucocorticoid receptor; BPV, bovine papillomavirus; RNA-FISH, RNA fluorescence in situ hybridization.
}
\label{fig:supp_single_subimage}
\end{figure}

\noindent To complement the aggregate evaluation results, we present five illustrative cases comparing M$^3$LLM with MedGemma-27B across different PMC-MI-Bench task formats (Supplementary Figs.~\ref{fig:supp_relative_position}--\ref{fig:supp_single_subimage}).
These selected examples illustrate distinct response patterns and were not used to estimate the prevalence of specific error types.

\FloatBarrier

\clearpage
\renewcommand{\refname}{Supplementary References}
\begin{spacing}{0.95}
\bibliographystyle{naturemag}
\bibliography{ref}
\end{spacing}